\documentclass[10pt,twocolumn,letterpaper]{article}

\usepackage{cvpr}
\usepackage{times}
\usepackage{epsfig}
\usepackage{graphicx}
\usepackage{amsmath}
\usepackage{amssymb}
\usepackage{graphicx}
\usepackage{subcaption}
\usepackage{caption}

\usepackage{amsthm}
\usepackage{algorithm}
\usepackage{algpseudocode}
\usepackage{mathtools}
\usepackage{overpic}
\usepackage{float}
\usepackage{siunitx}

\newcommand{\R}{\mathbb{R}}
\let\bs=\boldsymbol

\def \diag {\mathrm{diag}}
\def \det {\mathrm{det}}
\def \gt {\textup{gt}}

\def \saliency {\textup{\saliency}}

\def \gt {\mathit{gt}}

\def \init {\mathit{in}}
\def \path {\mathit{path}}

\makeatother

\theoremstyle{plain}
\newtheorem{lem}{\textbf{Lemma}}
\newtheorem{proposition}{\textbf{Proposition}}

\newtheorem{definition}{\textbf{Definition}}

 \theoremstyle{definition}
\numberwithin{theorem}{section}
\numberwithin{lem}{section}

\theoremstyle{remark}\newtheorem{remark}{\textbf{Remark}}

  \let\set=\mathcal 

 \global\long\def\diag{\mathrm{diag}}

\newcommand{\dd}{\mathrm{d}}

 \global\long\def\R{\mathbb{R}}

\usepackage[pagebackref=true,breaklinks=true,letterpaper=true,colorlinks,bookmarks=false]{hyperref}

\cvprfinalcopy 

\def\cvprPaperID{3497} 

\ifcvprfinal\fi
\begin{document}

\title{Learning Transformation Synchronization}

\author{
Xiangru Huang\\
UT Austin\\
\and
Zhenxiao Liang\\
UT Austin\\
\and
Xiaowei Zhou\\
Zhejiang University\thanks{Xiaowei Zhou is affiliated with the StateKey Lab of CAD\&CG and the ZJU-SenseTime Joint Lab of 3D Vision.}\\
\and
Yao Xie\\
Georgia Tech\\
\and
Leonidas Guibas\\
Facebook AI Research, Stanford University\\
\and
Qixing Huang\thanks{huangqx@cs.utexas.edu}\\
UT Austin\\
}

\maketitle

\begin{abstract}
Reconstructing the 3D model of a physical object typically requires us to align the depth scans obtained from different camera poses into the same coordinate system. Solutions to this global alignment problem usually proceed in two steps. The first step estimates relative transformations between pairs of scans using an off-the-shelf technique. Due to limited information presented between pairs of scans, the resulting relative transformations are generally noisy. The second step then jointly optimizes the relative transformations among all input depth scans. A natural constraint used in this step is the cycle-consistency constraint, which allows us to prune incorrect relative transformations by detecting inconsistent cycles. The performance of such approaches, however, heavily relies on the quality of the input relative transformations. Instead of merely using the relative transformations as the input to perform transformation synchronization, we propose to use a neural network to learn the weights associated with each relative transformation. Our approach alternates between transformation synchronization using weighted relative transformations and predicting new weights of the input relative transformations using a neural network. We demonstrate the usefulness of this approach across a wide range of datasets. 
\end{abstract}

\section{Introduction}
\label{Section:Introduction}

Transformation synchronization, i.e., estimating consistent rigid transformations across a collection of images or depth scans, is a fundamental problem in various computer vision applications, including multi-view structure from motion~\cite{DBLP:conf/iccv/ChatterjeeG13,ozyesil2015robust,conf/eccv/WilsonS14,DBLP:conf/iccv/SweeneySHTP15}, geometry reconstruction from depth scans~\cite{Huber-2002-8601,conf/cvpr/ChoiZK15}, image editing via solving jigsaw puzzles~\cite{DBLP:journals/pami/ChoAF10}, simultaneous localization and mapping~\cite{DBLP:conf/icra/CarloneTDD15}, and reassembling fractured surfaces~\cite{Huang:2006:RFO}, to name just a few. A common approach to transformation synchronization proceeds in two phases. The first phase establishes the relative rigid transformations between pairs of objects in isolation. Due to incomplete information presented in isolated pairs, the estimated relative transformations are usually quite noisy. The second phase improves the relative transformations by jointly optimizing them across all input objects. This is usually made possible by utilizing the so-called \textsl{cycle-consistency} constraint, which states that the composite transformation along every cycle should be the identity transformation, or equivalently, the data matrix that stores pair-wise transformations in blocks is low-rank (c.f.~\cite{Huang:2013:CSM}). This cycle-consistency constraint allows us to jointly improve relative transformations by either detecting inconsistent cycles~\cite{DBLP:journals/pami/ChoAF10,DBLP:journals/cgf/NguyenBWYG11} or performing low-rank matrix recovery~\cite{Huang:2013:CSM,Wang:2013:IMA,DBLP:conf/nips/PachauriKS13,arrigoni2016spectral,bernard2015solution}.

\begin{figure}
\centering
\setlength\tabcolsep{1.75pt}
\begin{tabular}{cc}
\includegraphics[width=0.48\columnwidth]{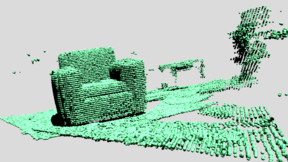} &
\includegraphics[width=0.48\columnwidth]{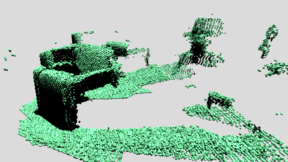} \\
(a) & (b) \\
\includegraphics[width=0.48\columnwidth]{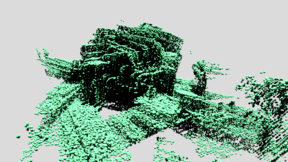} &
\includegraphics[width=0.48\columnwidth]{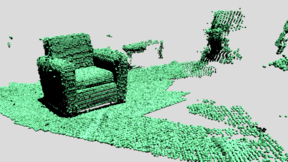} \\
(c) & (d)
\end{tabular}
\caption{\small{Reconstruction results from 30 RGBD images of an indoor environment using different transformation synchronization methods. (a) Our approach. (b) Rotation Averaging \cite{chatterjee2018robust}. (c) Geometric Registration\cite{conf/cvpr/ChoiZK15}. (d) Ground Truth.}}
\label{Figure:Best:Result}
\end{figure}

However, the success of existing transformation synchronization~\cite{Wang:2013:IMA,DBLP:conf/iccv/ChatterjeeG13,DBLP:journals/corr/ArrigoniFRF15,DBLP:conf/nips/HuangLBH17} and more general map synchronization~\cite{Huang:2013:CSM,DBLP:conf/nips/PachauriKS13,DBLP:conf/nips/PachauriKSS14,DBLP:conf/icml/ChenGH14,NIPS2016_6128,DBLP:conf/nips/HuangLBH17} techniques heavily depends on the compatibility between the loss function and the noise pattern of the input data. For example, approaches based on robust norms (e.g., L1~\cite{Huang:2013:CSM,DBLP:conf/icml/ChenGH14}) can tolerate either a constant fraction of adversarial noise (c.f.\cite{Huang:2013:CSM,DBLP:conf/nips/HuangLBH17}) or a sub-linear outlier ratio when the noise is independent (c.f.\cite{DBLP:conf/icml/ChenGH14,NIPS2016_6128}). Such assumptions, unfortunately, deviate from many practical settings, where the majority of the input relative transformations may be incorrect (e.g., when the input scans are noisy), and/or the noise pattern in relative transformations is highly correlated (there are a quadratic number of measurements from a linear number of sources). This motivates us to consider the problem of \textsl{learning transformation synchronization}, which seeks to learn a suitable loss function that is compatible with the noise pattern of specific datasets.

In this paper, we introduce an approach that formulates transformation synchronization as an end-to-end neural network. Our approach is motivated by reweighted least squares and their application in  transformation synchronization (c.f.~\cite{DBLP:conf/iccv/ChatterjeeG13,DBLP:journals/corr/ArrigoniFRF15,conf/cvpr/ChoiZK15,DBLP:conf/nips/HuangLBH17}), where the loss function dictates how we update the weight associated with each input relative transformation during the synchronization process. Specifically, we design a recurrent neural network that reflects this reweighted scheme. By learning the weights from data directly, our approach implicitly captures a suitable loss function for performing transformation synchronization.

\begin{figure*}
\centering
\includegraphics[width=0.8\textwidth]{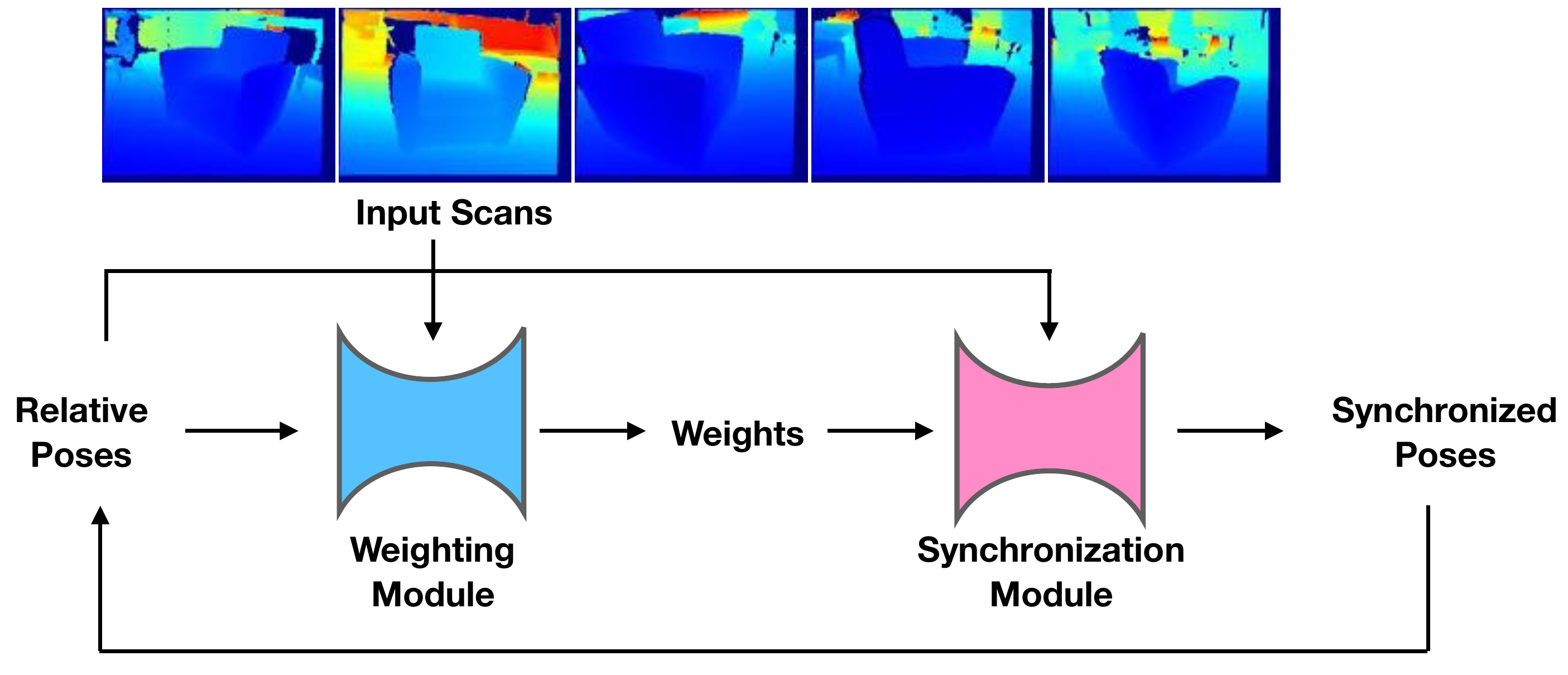}
\caption{Illustration of our network design.}
\label{Figure:Overview}
\end{figure*}

We have evaluated the proposed technique on two real datasets: Redwood~\cite{Choi2016} and ScanNet~\cite{dai2017scannet}. Experimental results show that our approach leads to considerable improvements compared to the state-of-the-art transformation synchronization techniques. For example, on Redwood and Scannet, the best combination of existing pairwise matching and transformation synchronization techniques lead to mean angular rotation errors $22.4^{\circ}$ and $64.4^{\circ}$, respectively. In contrast, the corresponding statistics of our approach are $6.9^{\circ}$ and $42.9^{\circ}$, respectively. We also perform an ablation study to evaluate the effectiveness of our approach.

Code is publicly available at \url{https://github.com/xiangruhuang/Learning2Sync}.

\section{Related Works}
\label{Section:Related:Works}

Existing techniques on transformation synchronization fall into two categories. The first category of methods~\cite{Huber-2002-8601,Huang:2006:RFO,conf/cvpr/ZachKP10,DBLP:journals/cgf/NguyenBWYG11,DBLP:conf/cvpr/ZhouLYE15} uses combinatorial optimization to select a subgraph that only contains consistent cycles. The second category of methods~\cite{Wang:2013:IMA,Kim:2012:ECM,DBLP:journals/tog/HuangZGHBG12,Huang:2013:CSM,DBLP:journals/tog/HuangWG14,DBLP:conf/icml/ChenGH14,zhou2015multi,NIPS2016_6128,DBLP:conf/icra/LeonardosZD17,DBLP:conf/nips/HuangLBH17, arrigoni2016spectral, DBLP:conf/nips/PachauriKS13,DBLP:conf/nips/PachauriKSS14, arrigoni2016camera, bernard2015solution, arrigoni2014robust, ArrFusAl18b, sharp2004multiview, fusiello2002model, torsello2011multiview, arrigoni2016global, govindu2014averaging} can be viewed from the perspective that there is an equivalence between cycle-consistent transformations and the fact that the map collection matrix that stores relative transformations in blocks is semidefinite and/or low-rank (c.f.\cite{Huang:2013:CSM}). These methods formulate transformation synchronization as low-rank matrix recovery, where the input relative transformations are considered noisy measurements of this low-rank matrix. In the literature, people have proposed convex optimization~\cite{Wang:2013:IMA,Huang:2013:CSM,DBLP:journals/tog/HuangWG14,DBLP:conf/icml/ChenGH14}, non-convex optimization~\cite{DBLP:conf/iccv/ChatterjeeG13,zhou2015multi,DBLP:conf/icra/LeonardosZD17,DBLP:conf/nips/HuangLBH17}, and spectral techniques~\cite{Kim:2012:ECM,DBLP:journals/tog/HuangZGHBG12,DBLP:conf/nips/PachauriKS13,DBLP:conf/nips/PachauriKSS14,NIPS2016_6128,DBLP:conf/eccv/SunLHH18, arrigoni2016spectral, arrigoni2016camera, bernard2015solution} for solving various low-rank matrix recovery formulations. Compared with the first category of methods, the second category of methods is computationally more efficient. Moreover, tight exact recovery conditions of many methods have been established. 

A message from these exact recovery conditions is that existing methods only work if the fraction of noise in the input relative transformations is below a threshold. The magnitude of this threshold depends on the noise pattern. Existing results either assume adversarial noise~\cite{Huang:2013:CSM,DBLP:conf/nips/HuangLBH17} or independent random noise~\cite{Wang:2013:IMA,DBLP:conf/icml/ChenGH14,NIPS2016_6128,DBLP:conf/icml/BajajGHHL18}. However, as relative transformations are computed between pairs of objects, it follows that these relative transformations are dependent (i.e., between the same source object to different target objects). This means there are a lot of structures in the noise pattern of relative transformations. Our approach addresses this issue by optimizing transformation synchronization techniques to fit the data distribution of a particular dataset. To best of our knowledge, this work is the first to apply supervised learning to the problem of transformation synchronization.

Our approach is also relevant to utilizing recurrent neural networks for solving the pairwise matching problem. Recent examples include learning correspondences between pairs of images~\cite{Yi_2018_CVPR}, predicting the fundamental matrix between two different images of the same underlying environment~\cite{DBLP:conf/eccv/RanftlK18}, and computing a dense image flow between an image pair~\cite{DBLP:conf/nips/KimLJMS18}. In contrast, we study a different problem of transformation synchronization in this paper. In particular, our weighting module leverages problem specific features (e.g., eigen-gap) for determining the weights associated with relative transformations. Learning transformation synchronization also poses great challenges in making the network trainable end-to-end. 

\section{Problem Statement and Approach Overview}

In this section, we describe the problem statement of transformation synchronization (Section~\ref{Section:Problem:Statement}) and present an overview of our approach (Section~\ref{Section:Approach:Overview}). 

\subsection{Problem Statement}
\label{Section:Problem:Statement}

Consider $n$ input scans $\set{S} = \{S_i, 1\leq i \leq n\}$ capturing the same underlying object/scene from different camera poses. Let $\Sigma_i$ denote the local coordinate system associated with $S_i$. The input to transformation synchronization can be described as a model graph $\set{G} = (\set{S}, \set{E})$~\cite{Huber01fullyautomatic}. Each edge $(i,j)\in \set{E}$ of the model graph is associated with a relative transformation $T_{ij}^{\init}=(R_{ij}^{\init}, \bs{t}_{ij}^{\init})\in \R^{3\times 4}$, where $R_{ij}^{\init}\in \R^{3\times 3}$ and $\bs{t}_{ij}^{\init}\in \R^{3}$ are rotational and translational components of $T_{ij}^{\init}$, respectively. $T_{ij}^{\init}$ is usually pre-computed using an off-the-shelf algorithm (e.g.,~\cite{Mellado:2014:SFG,DBLP:conf/eccv/ZhouPK16}). For simplicity, we impose the assumption that $(i,j)\in \set{E}$ if and only if (i) $(j,i)\in \set{E}$, and (ii) their associated transformations are compatible, i.e., 
$$
R_{ji}^{\init} = {R_{ij}^{\init}}^{T}, \quad \bs{t}_{ji}^{\init} = -{R_{ij}^{\init}}^{T}\bs{t}_{ij}^{\init}. 
$$
It is expected that many of these relative transformations are incorrect, due to limited information presented between pairs of scans and limitations of the off-the-shelf method being used. The goal of transformation synchronization is to recover the absolute pose $T_i = (R_i, \bs{t}_i) \in \R^{3\times 4}$ of each scan $S_i$ in a world coordinate system $\Sigma$. Without losing generality, we assume the world coordinate system is given by $\Sigma:=\Sigma_1$. Note that unlike traditional transformation synchronization approaches that merely use $T_{ij}^{\init}$ (e.g.,\cite{DBLP:conf/iccv/ChatterjeeG13,Wang:2013:IMA,DBLP:journals/corr/ArrigoniFRF15}), our approach also incorporates additional information extracted from the input scans $S_i, 1\leq i \leq n$. 

\subsection{Approach Overview}
\label{Section:Approach:Overview}

Our approach is motivated from iteratively reweighted least squares (or IRLS)\cite{Daubechies:2008:IRWa}, which has been applied to transformation synchronization (e.g.~\cite{DBLP:conf/iccv/ChatterjeeG13,DBLP:journals/corr/ArrigoniFRF15,conf/cvpr/ChoiZK15,DBLP:conf/nips/HuangLBH17}). The key idea of IRLS is to maintain an edge weight $w_{ij},(i,j)\in \set{E}$ for each input transformation $T_{ij}^{\init}$ so that the objective function becomes quadratic in the variables, and transformation synchronization admits a closed-form solution. One can then use the closed-form solution to update the edge weights. 
One way to understand reweighting schemes is that when the weights converged, the reweighted square loss becomes the actual robust loss function that is used to solve the corresponding transformation synchronization problem. In contrast to using a generic weighting scheme, we propose to learn the weighting scheme from data by designing a recurrent network that replicates the reweighted transformation synchronization procedure. By doing so, we implicitly learn a suitable loss function for transformation synchronization. 

As illustrated in Figure~\ref{Figure:Overview}, the proposed recurrent module combines a synchronization layer and a weighting module. At the $k$th iteration, the synchronization layer takes as input the initial relative transformations $T_{ij}^{\init}\in \R^{3\times 4},\forall (i,j)\in \set{E}$ and their associated weights $w_{ij}^{(k)}\in (0,1)$ and outputs synchronized poses $T_i^{(k)}: \Sigma_i \rightarrow \Sigma$ for the input objects $S_i, 1\leq i \leq n$. Initially, we set $w_{ij}^{(1)}=1, \forall (i,j)\in \set{E}$. The technical details of the synchronization layer are described in Section~\ref{Section:Synchronization:Operation}.

The weighting module operates on each object pair in isolation. For each edge $(i,j)\in \set{E}$, the input to the proposed weighting module consists of (1) the input relative transformation $T_{ij}^{\init}$, 
(2) features extracted from the initial alignment of the two input scans, and (3) a status vector $\bs{v}^{(k)}$ that collects global signals from the synchronization layer at the $k$th iteration (e.g., spectral gap). The output is the associated weight $w_{ij}^{(k+1)}$ at the $k+1$th iteration. 


The network is trained end-to-end by penalizing the differences between the ground-truth poses and the output of the last synchronization layer. The technical details of this end-to-end training procedure are described in Section~\ref{Section:End:2:End:Training}.
\section{Approach}
\label{Section:Overview}

In this section, we introduce the technical details of our learning transformation synchronization approach. In Section~\ref{Section:Synchronization:Operation}, we introduce details of the synchronization layer. In Section~\ref{Section:Weighting:Operation}, we describe the weighting module. Finally, we show how to train the proposed network end-to-end in Section~\ref{Section:End:2:End:Training}.
Note that the proofs of the propositions introduced in this section are deferred to the supplementary material. 

\subsection{Synchronization Layer}
\label{Section:Synchronization:Operation}

For simplicity, we ignore the superscripts $^k$ and $^{\init}$ when introducing the synchronization layer. Let $T_{ij} = (R_{ij}, \bs{t}_{ij})$ and $w_{ij}$ be the input relative transformation and its weights associated with the edge $(i,j)\in \set{E}$. We assume that this weighted graph is connected. The goal of the synchronization layer is to compute the synchronized pose $T_i^{\star}=(R_i^{\star},\bs{t}_i^{\star})$ associated with each scan $S_i$. Note that a correct relative transformation $T_{ij} = (R_{ij}, \bs{t}_{ij})$ induces two separate constraints on the rotations $R_i^{\star}$ and translations $\bs{t}_i^{\star}$, respectively:
$$
R_{ij}R_i^{\star} = R_j^{\star}, \quad R_{ij}\bs{t}_i^{\star}+\bs{t}_{ij} = \bs{t}_j^{\star}.
$$
We thus perform rotation synchronization and translation synchronization separately.

\noindent\textbf{Rotation synchronization.} Our rotation synchronization approach adapts a Laplacian rotation synchronization formulation proposed in the literature~\cite{Arie-Nachimson:2012:GME,arrigoni2016camera, bernard2015solution, arrigoni2014robust}. 
More precisely, we introduce a connection Laplacian $L\in \R^{3n\times 3n}$~\cite{Singer:2012:VDM},  whose blocks are given by
\begin{equation}
L_{ij} := \left\{
\begin{array}{cc}
\sum\limits_{j\in \set{N}(i)} w_{ij}I_3 & i = j \\
-w_{ij}R_{ij}^T & (i,j)\in \set{E} \\
0 & \textup{otherwise}
\end{array}
\right.\
\label{Eq:Laplacian:Matrix}
\end{equation}
where $\set{N}(i)$ collects all neighbor vertices of $i$ in $\set{G}$.

\begin{algorithm}[t]
\begin{algorithmic}
\Function{SYNC}{$(w_{ij},T_{ij}),\forall (i,j)\in \set{E}$}
\State Form the connection Laplacian $L$ and vector $\bs{b}$;
\State Compute first $3$ eigenvectors $U$ of $L$;
\State Perform SVD on blocks of $U$ to obtain $\{R_i^{\star},1\leq i \leq n\}$ via (\ref{Eq:Rotation:Sync:Solution});
\State Solve (\ref{Eq:Translation:Sync:Solution}) to obtain $\{\bs{t}_i^{\star},1\leq i \leq n\}$;
\State \Return $T_i^{\star} = (R_i^{\star},\bs{t}_i^{\star}),1\leq i \leq n$;
\EndFunction
\end{algorithmic}
\caption{Translation Synchronization Layer.}
\label{alg:basis_for_dag}
\end{algorithm}

Let $U = (U_1^{T},\cdots, U_n^{T})^T\in \R^{3n\times 3}$ collect the eigenvectors of $L$ that correspond to the three smallest eigenvalues. We choose the sign of each eigenvector such that $\sum_{i=1}^n \textup{det}(U_i) > 0$. To compute the absolute rotations, we first perform singular value decomposition (SVD) on each
$$
U_i=V_i\Sigma_i W_i^T. 
$$
We then output the corresponding absolute rotation estimate as
\begin{equation}
R_i^{*}=V_i W_i^T
\label{Eq:Rotation:Sync:Solution}
\end{equation}

It can be shown that when the observation graph is connected and $R_{ij}, (i,j)\in \set{E}$ are exact, then $R_i^{*}, 1\leq i \leq n$ recover the underlying ground-truth solution (c.f.\cite{Arie-Nachimson:2012:GME,arrigoni2016camera, bernard2015solution, arrigoni2014robust}). In Section~\ref{Section:Prop:4:1} of the supplementary material, we present a robust recovery result that $R_i^{\star}$ approximately recover the underlying ground-truth even when $R_{ij}$ are inexact. 

\noindent\textbf{Translation synchronization} solves the following least square problem to obtain $\bs{t}_i$:
\begin{equation}
\underset{\bs{t}_i, 1\leq i \leq n}{\textup{minimize}}\ \sum\limits_{(i,j)\in \set{E}} w_{ij} \|R_{ij}\bs{t}_i + \bs{t}_{ij}-\bs{t}_j\|^2
\label{Eq:Translation:Sync}
\end{equation}
Let $\bs{t} = (\bs{t}_1^{T},\cdots, \bs{t}_n^{T})^{T}\in \R^{3n}$ collect the translation components of the synchronized poses in a column vector. Introduce a column vector $\bs{b} = (\bs{b}_1^{T},\cdots, \bs{b}_n^{T})^{T}\in \R^{3n}$ where
$$
\bs{b}_{i}:= -\sum\limits_{j \in \set{N}(i)}w_{ij} R_{ij}^{T}\bs{t}_{ij}.
\label{Eq:Matrix:Vector:Block}
$$
Then an\footnote{When $L$ is positive semidefinite, then the solution is unique, and (\ref{Eq:Translation:Sync:Solution}) gives one optimal solution.} optimal solution $\bs{t}^{\star}$ to (\ref{Eq:Translation:Sync}) is given by
\begin{equation}
\bs{t}^{\star} = L^{+}\bs{b}.
\label{Eq:Translation:Sync:Solution}    
\end{equation}
Similar to the case of rotation synchronization, we can show that when the observation graph is connected, and $R_{ij}, \bs{t}_{ij}, (i,j)\in \set{E}$ are exact, then $\bs{t}^{\star}$ recovers the underlying ground-truth rotations. Section~\ref{Section:Prop:4:2} of the supplementary material presents a robust recovery result for translations. 

\begin{figure*}
\centering
\includegraphics[width=1.0\textwidth]{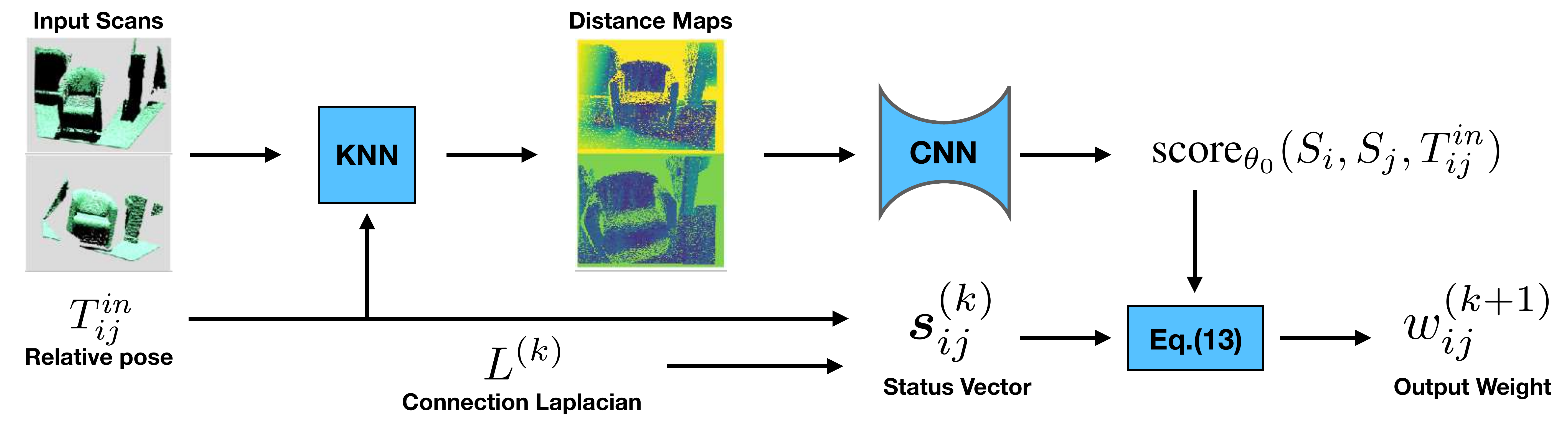}
\caption{\small{Illustration of network design of the weighting module. 
We first compute the nearest neighbor distance between a pair of depth images, which form the images (shown as heat maps) in the middle. In this paper, we use $k=1$. We then apply a classical convolutional neural network to output a score between $(0, 1)$, which is then combined with the status vector to produce the weight of this relative pose according to \eqref{Eq:weighting:module:definition}}. 
}
\label{Figure:Network:Design}
\vspace{-0.1in}
\end{figure*}

\subsection{Weighting Module}
\label{Section:Weighting:Operation}

We define the weighting module as the following function:
\begin{equation}
w_{ij}^{(k+1)} \leftarrow \textup{Weight}_{\theta}(S_i, S_j, T_{ij}^{\init}, \bs{s}_{ij}^{(k)})
\label{Eq:Matching:Module}
\end{equation}
\normalsize
where the input consists of (i) a pair of scans $S_i$ and $S_j$, (ii) the input relative transformation $T_{ij}^{\init}$ between them, and (iii) a status vector $\bs{s}_{ij}^{(k)}\in \R^4$. The output of this weighting module is given by the new weight $w_{ij}^{(k+1)}$ at the $k+1$th iteration. With $\theta$ we denote the trainable weights of the weighting module. In the following, we first introduce the definition of the status vector $\bs{s}_{ij}^{(k)}$. 

\noindent\textbf{Status vector.} The purpose of the status vector $\bs{s}_{ij}^{(k)}$ is to collect additional signals that are useful for determining the output of the weighting module. Define
\begin{align}
s_{ij1}^{(k)} &:= \|R_{ij}^{\init} -R_j^{(k)}{R_i^{(k)}}^{T}\|_{\set{F}},\label{Eq:1} \\
s_{ij2}^{(k)} &:= \|R_{ij}^{\init}\bs{t}_i^{(k)} +\bs{t}_{ij}^{\init}-\bs{t}_j^{(k)}\|.\label{Eq:2} \\
s_{ij3}^{(k)} &:= \lambda_4(L^{(k)}) - \lambda_3(L^{(k)}),\label{Eq:3} \\
s_{ij4}^{(k)} &:= \sum\limits_{(i,j)\in \set{E}}w_{ij}^{(k)}\|\bs{t}_{ij}^{(k)}\|^2 -{\bs{b}^{(k)}}^{T} {L^{(k)}}^+\bs{b}^{(k)}, \label{Eq:4}
\end{align}
Essentially, $s_{ij1}^{(k)}$ and $s_{ij2}^{(k)}$ characterize the difference between current synchronized transformations and the input relative transformations. The motivation for using them comes from the fact that for a standard reweighted scheme  for transformation synchronization (c.f.~\cite{DBLP:conf/nips/HuangLBH17}), one simply sets $w_{ij}^{(k+1)}=\rho(s_{ij1}^{(k)},s_{ij2}^{(k)})$ for a weighting function $\rho$ (c.f.~\cite{Daubechies:2008:IRWa}). This scheme can already recover the underlying ground-truth in the presence of a constant fraction of adversarial incorrect relative transformations (Please refer to Section~\ref{Section:Exact:Recovery:Rot:Sync} of the supplementary material for a formal analysis). In contrast, our approach seeks to go beyond this limit by leveraging additional information. The definition of $s_{ij3}^{(k)}$ captures the spectral gap of the connection Laplacian. $s_{ij4}^{(k)}$ equals to the residual of (\ref{Eq:Translation:Sync}). Intuitively, when $s_{ij3}^{(k)}$ is large and $s_{ij4}^{(k)}$ is small, the weighted relative transformations $w_{ij}^{(k)}\cdot T_{ij}^{\init}$ will be consistent, from which we can recover accurate synchronized transformations $T_i^{(k)}$. We now describe the network design.

\noindent\textbf{Network design.} As shown in Figure~\ref{Figure:Network:Design}, the key component of our network design is a sub-network $\textup{score}_{\theta_0}(S_i,S_j,T_{ij}^{\init})$ that takes two scans $S_i$ and $S_j$ and a relative transformation $T_{ij}^{\init}$ between them and output a score in $[0,1]$ that indicates whether this is a good scan alignment or not, i.e., $1$ means a good alignment, and $0$ means an incorrect alignment.

We design $\textup{score}_{\theta_0}$ as a feed-forward network. Its input consists of two color maps that characterize the alignment patterns between the two input scans. The value of each pixel represents the distance of the corresponding 3D point to the closest points on the other scan under $T_{ij}^{\init}$ (See the second column of images in Figure~\ref{Figure:Network:Design}). We then concatenate these two color images and feed them into a neural network (we used a modified AlexNet architecture\cite{Krizhevsky:2012:ICD}), which outputs the final score. 

With this setup, we define the output weight $w_{ij}^{(k+1)}$ as
\begin{align}
w_{ij}^{(k+1)} & := \frac{e^{\theta_1 \theta_2}}{e^{\theta_1 \theta_2} + (\textup{score}_{\theta_0}(S_i, S_j, T_{ij}^{\init}) {\bs{s}_{ij}^{(k)}}^T\theta_3)^{\theta_2}}
\label{Eq:weighting:module:definition}
\end{align}
Note that (\ref{Eq:weighting:module:definition}) is conceptually similar to the reweighting scheme $\rho_{\sigma}(x) = x^2/(\sigma^2 + x^2)$ that is widely used in $L^0$ minimization (c.f\cite{Daubechies:2008:IRWa}). However, we make elements of the factors and denominators parametric, so as to incorporate status vectors and to capture dataset specific distributions. 
Moreover, we use exponential functions in (\ref{Eq:weighting:module:definition}), since they lead to a loss function that is easier to optimize. 
With $\theta = (\theta_0, {\theta}_1,\theta_2, \theta_3)$ we collect all trainable parameters of (\ref{Eq:weighting:module:definition}).  

\subsection{End-to-End Training}
\label{Section:End:2:End:Training}

Let $\set{D}$ denote a dataset of scan collections with annotated ground-truth poses. Let $k_{\max}$ be the number of recurrent steps (we used four recurrent steps in our experiments) . We define the following loss function for training the weighting module $\textup{Weight}_{\theta}$:
\begin{align}
\min\limits_{\theta} \sum\limits_{\set{S}\in \set{D}} \sum\limits_{1\leq i < j \leq |\set{S}|} & \Big(\|{R_j^{k_{\max}}}{R_i^{k_{\max}}}^{T}-{R_j^{\gt}}{R_i^{\gt}}^{T}\|_{\set{F}}^2 \nonumber \\
&+ \lambda \|\bs{t}_i^{k_{\max}}-\bs{t}_i^{\gt}\|^2\Big)
\label{Eq:Total:Loss:Func}
\end{align}
where we set $\lambda = 10$ in all of our experiments. Note that we compare relative rotations in (\ref{Eq:Total:Loss:Func}) to factor out the global orientation among the poses. The global shift in translation is already handled by (\ref{Eq:Translation:Sync:Solution}).

We perform back-propagation to optimize (\ref{Eq:Total:Loss:Func}). The technical challenges are to compute the derivatives that pass through the synchronization layer, including 1) the derivatives of $R_j^{\star}{R_i^{\star}}^{T}$ with respect to the elements of $L$, 2) the derivatives of $\bs{t}_i^{\star}$ with respect to the elements of $L$ and $\bs{b}$, and 3) the derivatives of each status vector with respect to the elements of $L$ and $\bs{b}$. In the following, we provide explicit expressions for computing these derivatives.  

We first present the derivative between the output of rotation synchronization and its input. To make the notation uncluterred, we compute the derivative by treating $L$ as a matrix function. The derivative with respect to $w_{ij}$ can be easily obtained via chain-rule. 
\begin{proposition} Let $\bs{u}_i$ and $\lambda_i$ be the $i$-th eigenvector and eigenvalue of $L$, respectively. Expand the SVD of $U_i = V_i\Sigma_i W_i^{T}$ as follows:
\begin{align*}
V_i & = (\bs{v}_{i,1},\bs{v}_{i,2},\bs{v}_{i,3}),\ \Sigma_i = \diag(\sigma_{i,1},\sigma_{i,2},\sigma_{i,3}),\\
W_i & = (\bs{w}_{i,1},\bs{w}_{i,2},\bs{w}_{i,3}).
\end{align*}
Let $\bs{e}_{j}^{t}\in \R^t$ be the $j$th canonical basis of $\R^t$. We then have
$$
\dd (R_j^{\star}{R_i^{\star}}^T) = \dd R_j \cdot {R_i^{\star}}^T +R_j^{\star}\cdot {\dd R_i}^T,  
$$
where
$$
\dd R_i:= \sum_{1\leq s,t\leq 3}\frac{{\bs{v}_{i,s}}^T\dd U_i\bs{w}_{i,t}-{\bs{v}_{i,t}}^T\dd U_i \bs{w}_{i,s}}{\sigma_{i,s}+\sigma_{i,t}}\bs{v}_{i,s}{\bs{w}_{i,t}}^T,
$$
where $\dd U_i$ is defined by $\forall 1\leq j \leq 3$,
$$
\dd U_i \bs{e}_j^{(3)} = ({\bs{e}_i^{(n)}}^T\otimes I_3)\sum_{l=4}^{3n}\frac{\bs{u}_l\bs{u}_l^T}{\lambda_j - \lambda_l}\dd L\bs{u}_j.
$$
\label{Prop:3}
\end{proposition}
The following proposition specifies the derivative of $\bs{t}^{\star}$ with respect to the elements of $L$ and $\bs{b}$:
\begin{proposition}
The derivatives of $\bs{t}^{\star}$ are given by
$$
\dd \bs{t}^{\star} = L^{+} \dd L L^{+} + L^{+} \dd \bs{b}.
$$
\label{Prop:4}
\end{proposition}
\vspace{-0.15in}

Regarding the status vectors, the derivatives of $s_{ij,1}$ with respect to the elements of $L$ are given by Prop.~\ref{Prop:3}; The derivatives of $s_{ij,2}$ and $s_{ij,4}$ with respect to the elements of $L$ are given by Prop.~\ref{Prop:4}. It remains to compute the derivatives of $s_{ij.3}$ with respect to the elements of $L$, which can be easily obtained via the derivatives of the eigenvalues of $L$~\cite{Kadalbajoo:2011:OEC}, i.e., 
$$
\dd \lambda_i = \bs{u}_i^{T}\dd L \bs{u}_i.
$$

\section{Experimental Results}
\label{Section:Experimental:Results}

This section presents an experimental evaluation of the proposed learning transformation synchronization approach. We begin with describing the experimental setup in Section~\ref{Section:Experimental:Setup}. 
In Section~\ref{Section:Analysis:Of:Results}, we analyze the results of our approach and compare it against baseline approaches. Finally, we present an ablation study in Section~\ref{Section:Ablation:Study}. 

\begin{figure*}[t]
\centering
\vspace{+0.2cm}
\label{table_heldout}
\begin{footnotesize}
{
\setlength\tabcolsep{1.25pt}
\begin{tabular}{|ccc|cccccc|cccccc|cccccc|cccccc|} \hline 
\multicolumn{3}{|c|}{Methods} &\multicolumn{12}{c|}{Redwood} & \multicolumn{12}{c|}{ScanNet} \\
\hline 
&&& \multicolumn{6}{c|}{Rotation Error} & \multicolumn{6}{c|}{Translation Error (m)} & \multicolumn{6}{c|}{Rotation Error} & \multicolumn{6}{c|}{Translation Error (m)}
\\ \hline
 & & & \ang{3} & \ang{5} & \ang{10} & \ang{30} & \ang{45} & Mean & 0.05 & 0.1 & 0.25 & 0.5 & 0.75 & Mean & \ang{3} & \ang{5} &\ang{10} & \ang{30} & \ang{45} & Mean & 0.05 & 0.1 & 0.25 & 0.5 & 0.75 & Mean \\ \hline
 \multicolumn{3}{|c|}{FastGR (all)} & 29.4 & 40.2 & 52.0 & 63.8 & 70.4 & \ang{37.4} & 22.0 & 39.6 & 53.0 & 60.3 & 67.0 & 0.68 & 9.9 & 16.8 & 23.5 & 31.9 & 38.4 & \ang{76.3} & 5.5 & 13.3 & 22.0 & 29.0 & 36.3 & 1.67  \\ 
 \multicolumn{3}{|c|}{FastGR (good)} & 33.9 & 45.2 & 57.2 & 67.4 & 73.2& \ang{34.1}& 26.7 & 45.7 & 58.8 & 65.9 & 71.4 & 0.59 & 12.4 & 21.4 & 29.5 & 38.6 & 45.1 & \ang{68.8} & 7.7 & 17.6 & 28.2  & 36.2 & 43.4 & 1.43 \\ 
 \multicolumn{3}{|c|}{Super4PCS (all)}& 6.9& 10.1 & 16.7& 39.6& 52.3& \ang{55.8} & 4.2 & 8.9 & 18.2 & 31.0 & 43.5 & 1.14 & 0.5 & 1.3 & 4.0 & 17.4 & 25.2 & \ang{98.5} & 0.3 & 1.2 & 5.3 & 13.3& 21.6& 2.11 \\
 \multicolumn{3}{|c|}{Super4PCS (good)}& 10.3& 14.9 & 23.9& 48.0& 60.0& \ang{49.2} & 6.4 & 13.3 & 26.2 & 41.2 & 53.2 & 0.93 & 0.8 & 2.3 & 6.4 & 23.0  & 31.7  & \ang{90.8} & 0.6 & 2.2 & 8.9 & 19.5 & 29.5 & 1.80 \\ \hline
 \multicolumn{3}{|c|}{RotAvg (FastGR)}& 30.4 & 42.6 & 59.4 & 74.4 & 82.1 & \ang{22.4} & 23.3 & 43.2 & 61.8 & 72.4 & 80.7 & 0.42 & 6.0 & 10.4 & 17.3 & 36.1 & 46.1 & \ang{64.4} & 3.7 & 9.2 & 19.5 & 34.0& 45.6 & 1.26 \\ 
 \multicolumn{3}{|c|}{GeoReg (FastGR)}& 17.8 & 28.7 & 47.5 & 74.2& 83.2& \ang{27.7} & 4.9 & 18.4 & 50.2 & 72.6 & 81.4 & 0.93 & 0.2 & 0.6 & 2.8 & 16.4 & 27.1 & \ang{87.2} & 0.1 & 0.7 & 4.8 & 16.4& 28.4& 1.80\\ 
 \multicolumn{3}{|c|}{RotAvg (Super4PCS)} & 5.4 & 8.7 & 17.4& 45.1& 59.2& \ang{49.6} & 3.2 & 7.4 & 17.0 & 32.3 & 46.3 & 0.95 & 0.3 & 0.8 & 3.0 & 15.4 & 23.3 & \ang{96.8} & 0.2 & 1.0 & 5.8 & 16.5 & 27.6 & 1.70 \\ 
 \multicolumn{3}{|c|}{GeoReg (Super4PCS)}& 2.1& 4.1& 10.2& 33.1& 48.3& \ang{60.6} & 1.1 & 3.1 & 10.3 & 21.5 & 31.8 & 1.25 & 1.9  & 5.1  & 13.9 & 36.6 & 47.1  & \ang{72.9} & 0.4 & 2.1 & 9.8 & 23.2 & 34.5 & 1.82 \\ 
 \multicolumn{3}{|c|}{TranSyncV2 (FastGR)}& 9.5& 17.9& 35.8& 69.7& 80.1& \ang{27.5} & 1.5 & 6.2 & 24.0 & 48.8 & 67.5 & 0.62 & 0.4  & 1.5  & 6.1 & 29.0 & 42.2  & \ang{68.1} & 0.2 & 1.5 & 11.3 & 32.0 & 46.3 & 1.44 \\ 
 \multicolumn{3}{|c|}{EIGSE3 (FastGR)}& 36.6& 47.2& 60.4& 74.8& 83.3& \ang{21.3} & 21.5 & 36.7 & 57.2 & 70.4 & 79.2 & 0.43 & 1.5  & 4.3  & 12.1 & 34.5 & 47.7 & \ang{68.1} & 1.2 & 4.1 & 14.7 & 32.6 & 46.0 & 1.29 \\ 
 \hline
\multicolumn{3}{|c|}{Our Approach (FastGR)} & \textbf{67.5} & \textbf{77.5} & \textbf{85.6}& \textbf{91.7}& \textbf{94.4} & $\textbf{6.9}^{\circ}$ & 20.7 & 40.0 & 70.9 & \textbf{88.6} & \textbf{94.0} & 0.26 & \textbf{34.4} & \textbf{41.1} & \textbf{49.0} & \textbf{58.9} & \textbf{62.3} & \textbf{42.9}$^{\circ}$ & 2.0 & 7.3 & 22.3 & 36.9& 48.1& 1.16 \\
 \multicolumn{3}{|c|}{Our Approach (Super4PCS)} & 2.3 & 5.1 & 13.2& 42.5& 60.9& \ang{46.7} & 1.1 & 4.0 & 13.8 & 29.0 & 42.3 & 1.02 & 0.4 & 1.7 & 6.8 & 29.6 & 43.5 &\ang{66.9} & 0.1 & 0.8 &5.6  &16.6 & 27.0 & 1.90 \\
 \hline
\multicolumn{3}{|c|}{Transf. Sync. (FastGR)}& 27.1 & 37.7 & 56.9 & 74.4 & 82.4 & \ang{22.1} & 17.4 & 34.4 & 55.9 & 70.4 & 81.3 & 0.43 & 3.2 & 6.5 & 14.6 & 35.8 & 47.4 & \ang{63.5} & 1.6 & 5.6 & 15.5 & 30.9  & 43.4 & 1.31\\ 
\multicolumn{3}{|c|}{Input Only (FastGR)} & 36.7 & 51.4 & 68.1 & 87.7 & 91.7 & \ang{13.7} & 25.1 & \textbf{49.3} & 73.2 & 86.4 & 91.6 & 0.26 & 11.7 & 19.4 & 30.5 & 50.7   & 57.7 & \ang{51.7} & \textbf{5.9} & \textbf{15.4} & \textbf{30.5} & 43.7 & 52.2 & 1.03 \\ 
\multicolumn{3}{|c|}{No Recurrent (FastGR)} & 37.8 & 52.8 & 71.1 & 87.7 & 91.7 & \ang{12.9} & \textbf{26.3} & 51.1 & \textbf{77.3} & 87.1 & 92.0 & \textbf{0.24} & 8.6 & 15.3 & 26.9 & 51.4 & 58.2 & \ang{49.8} & 3.9 & 11.1 & 27.3 & \textbf{43.7} & \textbf{53.9} & \textbf{1.01} \\
\hline

\end{tabular}
}
\end{footnotesize}

\caption{Benchmark evaluations on Redwood~\protect\cite{Choi2016} and ScanNet~\protect\cite{dai2017scannet}. Quality of absolute poses are evaluated by computing errors to pairwise ground truth poses. Angular distances between rotation matrices are computed via angular $(R_{ij}, R_{ij}^{\star}) = \arccos (\frac{tr(R_{ij}^T R_{ij}^{\star}) - 1}{2})$. Translation distances are computed by $\|t_{ij} - t_{ij}^{\star}\|$. We collect statistics on percentages of rotation and translation errors that are below a varying threshold.
I) The 4th to 7th rows contain evaluations for upstream algorithms. (all) refers to statistics among all pairs where (good) refers to the statistics computed among relative poses with good quality overlap regions. II) For the second part, we report results of all baselines computed from this good set of relative poses, which is consistently better than the results from all relative poses. Since there are two input methods, we report the results of each transformation synchronization approach on both inputs. III) The third parts contain results for ablation study performed only on FastGR\protect\cite{DBLP:conf/eccv/ZhouPK16} inputs. The first row reports state-of-the-art rotation and translation synchronization results, followed by variants of our approach.
}
\vspace{-0.1in}
\label{Figure:Benchmark:Evaluation}
\end{figure*}

\subsection{Experimental Setup}
\label{Section:Experimental:Setup}

\noindent\textbf{Datasets.} We consider two datasets in this paper, Redwood~\cite{Choi2016} and ScanNet~\cite{dai2017scannet}:
\begin{itemize}
\item \textbf{Redwood} contains RGBD sequences of individual objects. We uniformly sample 60 sequences. For each sequence, we sample 30 RGBD images that are 20 frames away from the next one, which cover 600 frames of the original sequence. For experimental evaluation, we use the poses associated with the reconstruction as the ground-truth. We use 35 sequences for training and 25 sequences for testing. Note that the temporal order among the frames in each sequence is discarded in our experiments. 

\item \textbf{ScanNet} contains RGBD sequences, as well as reconstruction, camera pose, for 706 indoor scenes. Each scene contains 2-3 sequences of different trajectories. We randomly sample 100 sequences from ScanNet. We use 70 sequences for training and 30 sequences for testing. Again the temporal order among the frames in each sequence is discarded in our experiments. 

\end{itemize}
More details about the sampled sequences are given in the supplementary material. 

\noindent\textbf{Pairwise methods.} We consider two state-of-the-art pairwise methods for generating the input to our approach:
\begin{itemize}
\item \textbf{Super4PCS~\cite{Mellado:2014:SFG}} applies sampling to find consistent matches of four point pairs. 

\item \textbf{Fast Global Registration (FastGR)~\cite{DBLP:conf/eccv/ZhouPK16}} utilizes feature correspondences and applies reweighted non-linear least squares to extract a set of consistent feature correspondences and fit a rigid pose. We used the Open3D implementation~\cite{DBLP:journals/corr/abs-1801-09847}.
\end{itemize}

\noindent\textbf{Baseline approaches.} We consider the following baseline approaches that are introduced in the literature for transformation synchronization:
\begin{itemize}
\item \textbf{Robust Relative Rotation Averaging  (RotAvg)~\cite{chatterjee2018robust}} is a scalable algorithm that performs robust rotation averaging of relative rotations. To recover translations, we additionally apply a state-of-the-art translation synchronization approach~\cite{DBLP:conf/nips/HuangLBH17}. We use default setting of its publicly accessible code. \cite{DBLP:conf/nips/HuangLBH17} is based on our own Python implementation.
\item \textbf{Geometric Registration (GeoReg)~\cite{conf/cvpr/ChoiZK15}} solve multi-way registration via pose graph optimization. We modify the Open3D implementation
to take inputs from Super4PCS or FastGR.

\item \textbf{Transformation Synchronization (TranSyncV2)~\cite{bernard2015solution}} is a spectral approach that aims to find a low rank approximation of the null space of the Laplacian matrix. We used the authors' code. 

\item \textbf{Spectral Synchronization in SE(3) (EIGSE3)~\cite{arrigoni2016spectral}} is another spectral approach that considers translation and rotation together by working in SE(3). We used the authors' code.
\end{itemize}

Note that our approach utilizes a weighting module to score the input relative transformations. To make fair comparisons, we use the median nearest-neighbor distances between the overlapping regions (defined as points within distance $0.2m$ from the other point cloud) to filter all input transformations, and select those with median distance below $0.1m$. Note that with smaller threshold the pose graph will be disconnected. We then feed these filtered input transformations to each baseline approach for experimental evaluation. 

\noindent\textbf{Evaluation protocol.} We employ the evaluation protocols of \cite{DBLP:conf/iccv/ChatterjeeG13} and \cite{DBLP:conf/nips/HuangLBH17} for evaluating rotation synchronization and translation synchronization, respectively. Specifically, for rotations, we first solve the best matching global rotation between the ground-truth and the prediction, we then report the statistics and the cumulative distribution function (CDF) of angular deviation $\arccos(\frac{\|\log(R^{T}R^{\gt}\|_{\set{F}}}{\sqrt{2}})$ between a prediction $R$ and its corresponding ground-truth $R^{\gt}$. For translations, we report the statistics and CDF of $\|\bs{t}-\bs{t}^{\gt}\|$ between each pair of prediction $\bs{t}$ and its corresponding ground-truth $\bs{t}^{\gt}$. The unit of translation errors are meters (m). The statistics are shown in Figure \ref{Figure:Benchmark:Evaluation} and the CDF plots are shown in Section~\ref{Section:More:Results} of the supplementary material.

\begin{figure*}
\setlength\tabcolsep{1.75pt}
\begin{tabular}{cccccc}
\includegraphics[width=0.16\textwidth]{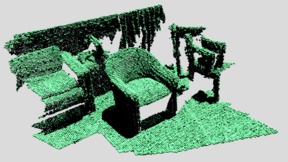}
&
\includegraphics[width=0.16\textwidth]{Figures/gt_01606.png}
&
\includegraphics[width=0.16\textwidth]{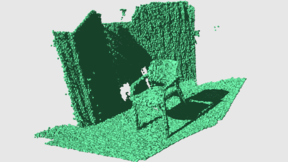}
&
\includegraphics[width=0.16\textwidth]{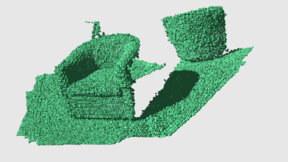}
&
\includegraphics[width=0.16\textwidth]{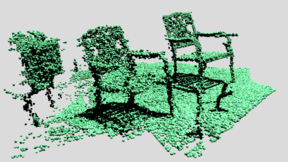} 
&
\includegraphics[width=0.16\textwidth]{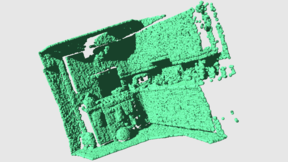} \\
\includegraphics[width=0.16\textwidth]{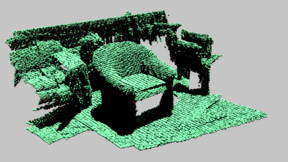}
&
\includegraphics[width=0.16\textwidth]{Figures/SO3_01606.png}
&
\includegraphics[width=0.16\textwidth]{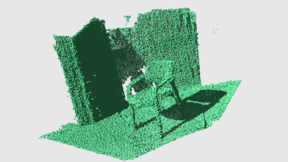}
&
\includegraphics[width=0.16\textwidth]{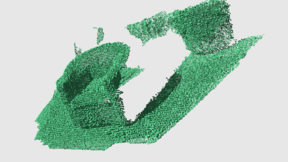}
&
\includegraphics[width=0.16\textwidth]{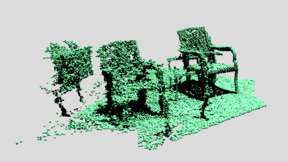}
&
\includegraphics[width=0.16\textwidth]{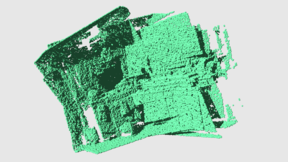} \\
\includegraphics[width=0.16\textwidth]{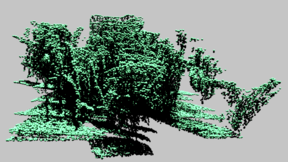}
&
\includegraphics[width=0.16\textwidth]{Figures/multiway_01606.png}
&
\includegraphics[width=0.16\textwidth]{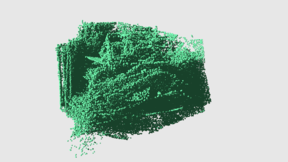}
&
\includegraphics[width=0.16\textwidth]{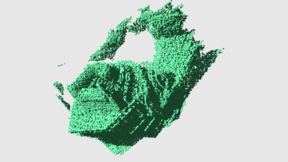}
&
\includegraphics[width=0.16\textwidth]{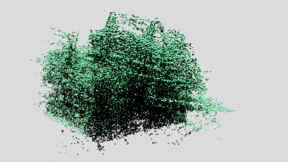}
&
\includegraphics[width=0.16\textwidth]{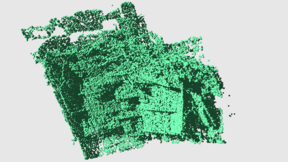} \\
\includegraphics[width=0.16\textwidth]{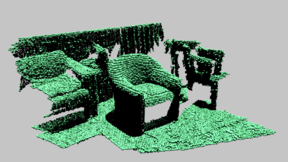}
&
\includegraphics[width=0.16\textwidth]{Figures/network_01606.png}
&
\includegraphics[width=0.16\textwidth]{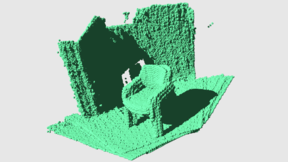}
&
\includegraphics[width=0.16\textwidth]{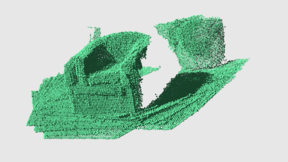}
&
\includegraphics[width=0.16\textwidth]{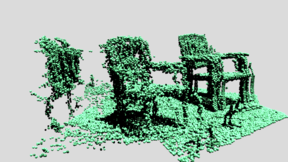}
&
\includegraphics[width=0.16\textwidth]{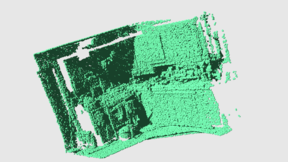} 
\end{tabular}
\caption{\small{Each column represents the results of one scene. From bottom to top, we show the results of our approach , Rotation Averaging~\protect\cite{chatterjee2018robust}+Translation Sync.~\protect\cite{DBLP:conf/nips/HuangLBH17}~ (row II), Geometric  Registration~\protect\cite{conf/cvpr/ChoiZK15} (row III), and Ground Truth (row IV) (Top). The left four scenes are from Redwood~\protect\cite{Choi2016} and the right two scenes are from ScanNet~\protect\cite{dai2017scannet}}}
\label{Figure:Visual:Results}
\vspace{-0.15in}
\end{figure*}

\subsection{Analysis of Results}
\label{Section:Analysis:Of:Results}

Figure~\ref{Figure:Benchmark:Evaluation} and Figure~\ref{Figure:Visual:Results} present quantitative and qualitative results, respectively. Overall, our approach yielded fairly accurate results. On Redwood, the mean errors in rotations/translations of FastGR and our result from FastGR are $34.1^{\circ}/0.58m$ and $6.9^{\circ}/0.26m$, respectively.  On ScanNet, the mean errors in rotations/translations of FastGR and our result from FastGR are $68.8^{\circ}/1.43m$ and $42.9^{\circ}/1.16m$, respectively. Note that in both cases, our approach leads to salient improvements from the input. The final results of our approach on ScanNet are less accurate than those on Redwood. Besides the fact that the quality of the initial relative  transformations is lower on ScanNet than that on Redwood, another factor is that depth scans from ScanNet are quite noisy, leading to noisy input (and thus less signals) for the weighting module. Still, the improvements of our approach on ScanNet are salient. 

Our approach still requires reasonable initial transformations to begin with. This can be understood from the fact that our approach seeks to perform synchronization by selecting a subset of input relative transformations. Although our approach utilizes learning, its performance shall decrease when the quality of the initial relative transformations drops. An evidence is that our approach only leads to modest performance gains when taking the output of Super4PCS as input.

\noindent\textbf{Comparison with state-of-the-art approaches.} Although all the two baseline approaches improve from the input relative transformations, our approach exhibits significant further improvements from all baseline approaches. On Redwood, the mean rotation and translation errors of the top performing method RotAvg from FastGR are $22.4^{\circ}$ and $0.418m$, respectively. The reductions in mean error of our approach are $69.2\%$ and $39.0\%$ for rotations and translations, respectively, which are significant. The reductions in mean errors of our approach on ScanNet are also noticeable, i.e., $33.3\%$ and $7.4\%$ in rotations and translations, respectively. 

Our approach also achieved relative performance gains from baseline approaches when taking the output of Super4PCS as input. In particular, for mean rotation errors, our approach leads to  reductions of $5\%$ and $9\%$ on Redwood and ScanNet, respectively. 

When comparing rotations and translations, the improvements on mean rotation errors are bigger than those on mean translation errors. One explanation is that there are a lot of planar structures in Redwood and ScanNet. When aligning such planar structures, rotation errors easily lead to a large change in nearest neighbor distances and thus can be detected by our weighting module. In contrast, translation errors suffer from the gliding effects on planar structures (c.f.\cite{DBLP:conf/3dim/GelfandRIL03}). For example, there are rich planar structures that consist of a pair of perpendicular planes, and aligning such planar structures may glide along the common line of these plane pairs. As a result, our weighting module becomes less effective for improving the translation error.  

\subsection{Ablation Study}
\label{Section:Ablation:Study}

In this section, we present two variants of our learning transformation synchronization approach to justify the usefulness of each component of our system. Due to space constraint, we perform ablation study only using FastGR.

\noindent\textbf{Input only.} In the first experiment, we simply learn to classify the input maps, and then apply transformation synchronization techniques on the filtered input transformations. In this setting, state-of-the-art transformation synchronization techniques achieves mean rotation/translation errors of $22.1^{\circ}/0.43m$ and $63.5^{\circ}/1.25m$ on Redwood and ScanNet, respectively. By applying our learning approach to fixed initial map weights, e.g., we fix $\theta_0$ of the weighting module in (\ref{Eq:weighting:module:definition}), our approach reduced the mean errors to $13.7^{\circ}/0.255m$ and $51.7^{\circ}/1.031m$ on Redwood and ScanNet, respectively. Although such improvements are noticeable, there are still gaps between this reduced approach and our full approach. This justifies the importance of learning the weighting module together.

\noindent\textbf{No recurrent module.} Another reduced approach is to directly combine the weighting module and one synchronization layer. Although this approach can improve from the input transformations. There is still a big gap between this approach and our full approach (See the last row in Figure~\ref{Figure:Benchmark:Evaluation}). This shows the importance of using weighting modules to gradually reduce the error while simultaneously make the entire procedure trainable end-to-end.

\section{Conclusions}
\label{Section:Conclusions}

In this paper, we have introduced a supervised transformation synchronization approach. It modifies a reweighted nonlinear least square approach and applies a neural network to automatically determine the input pairwise transformations and the associated weights. We have shown how to train the resulting recurrent neural network end-to-end. Experimental results show that our approach is superior to state-of-the-art transformation synchronization techniques on ScanNet and Redwood for two state-of-the-art pairwise scan matching methods.

There are ample opportunities for future research. So far we have only considered classifying pairwise transformations, it would be interesting to study how to classify high-order matches. Another interesting direction is to install ICP alignment into our recurrent procedure, i.e., we start from the current synchronized poses and perform ICP between pairs of scans to obtain more signals for transformation synchronization. Moreover, instead of maintaining one synchronized pose per scan, we can maintain multiple synchronized poses, which offer more pairwise matches between pairs of scans for evaluation. Finally, we would like to apply our approach to synchronize dense correspondences across multiple images/shapes.

\noindent\textbf{Acknowledgement:} The authors wish to thank the support of NSF grants DMS-1546206, DMS-1700234, CHS-1528025, a DoD Vannevar Bush Faculty Fellowship, a Google focused research award, a gift from adobe research, a gift from snap research, a hardware donation from NVIDIA, an Amazon AWS AI Research gift, NSFC (No. 61806176), and Fundamental Research Funds for the Central Universities.

{\small
\bibliographystyle{ieee}
\bibliography{sync}

\begin{thebibliography}{10}\itemsep=-1pt

\bibitem{Arie-Nachimson:2012:GME}
Mica Arie-Nachimson, Shahar~Z. Kovalsky, Ira Kemelmacher-Shlizerman, Amit
  Singer, and Ronen Basri.
\newblock Global motion estimation from point matches.
\newblock In {\em Proceedings of the 2012 Second International Conference on 3D
  Imaging, Modeling, Processing, Visualization \& Transmission}, 3DIMPVT '12,
  pages 81--88, Washington, DC, USA, 2012. IEEE Computer Society.

\bibitem{arrigoni2016camera}
Federica Arrigoni, Andrea Fusiello, and Beatrice Rossi.
\newblock Camera motion from group synchronization.
\newblock In {\em 3D Vision (3DV), 2016 Fourth International Conference on},
  pages 546--555. IEEE, 2016.

\bibitem{DBLP:journals/corr/ArrigoniFRF15}
Federica Arrigoni, Andrea Fusiello, Beatrice Rossi, and Pasqualina Fragneto.
\newblock Robust rotation synchronization via low-rank and sparse matrix
  decomposition.
\newblock {\em CoRR}, abs/1505.06079, 2015.

\bibitem{arrigoni2014robust}
Federica Arrigoni, Luca Magri, Beatrice Rossi, Pasqualina Fragneto, and Andrea
  Fusiello.
\newblock Robust absolute rotation estimation via low-rank and sparse matrix
  decomposition.
\newblock In {\em 3D Vision (3DV), 2014 2nd International Conference on},
  volume~1, pages 491--498. IEEE, 2014.

\bibitem{ArrFusAl18b}
Federica Arrigoni, Beatrice Rossi, Pasqualina Fragneto, and Andrea Fusiello.
\newblock Robust synchronization in {SO(3)} and {SE(3)} via low-rank and sparse
  matrix decomposition.
\newblock {\em Computer Vision and Image Understanding}, 174:95--113, 2018.

\bibitem{arrigoni2016global}
Federica Arrigoni, Beatrice Rossi, and Andrea Fusiello.
\newblock Global registration of 3d point sets via lrs decomposition.
\newblock In {\em European Conference on Computer Vision}, pages 489--504.
  Springer, 2016.

\bibitem{arrigoni2016spectral}
Federica Arrigoni, Beatrice Rossi, and Andrea Fusiello.
\newblock Spectral synchronization of multiple views in se (3).
\newblock {\em SIAM Journal on Imaging Sciences}, 9(4):1963--1990, 2016.

\bibitem{DBLP:conf/icml/BajajGHHL18}
Chandrajit Bajaj, Tingran Gao, Zihang He, Qixing Huang, and Zhenxiao Liang.
\newblock {SMAC:} simultaneous mapping and clustering using spectral
  decompositions.
\newblock In {\em Proceedings of the 35th International Conference on Machine
  Learning, {ICML} 2018, Stockholmsm{\"{a}}ssan, Stockholm, Sweden, July 10-15,
  2018}, pages 334--343, 2018.

\bibitem{bernard2015solution}
Florian Bernard, Johan Thunberg, Peter Gemmar, Frank Hertel, Andreas Husch, and
  Jorge Goncalves.
\newblock A solution for multi-alignment by transformation synchronisation.
\newblock In {\em Proceedings of the IEEE Conference on Computer Vision and
  Pattern Recognition}, pages 2161--2169, 2015.

\bibitem{DBLP:conf/icra/CarloneTDD15}
Luca Carlone, Roberto Tron, Kostas Daniilidis, and Frank Dellaert.
\newblock Initialization techniques for 3d {SLAM:} {A} survey on rotation
  estimation and its use in pose graph optimization.
\newblock In {\em {ICRA}}, pages 4597--4604. {IEEE}, 2015.

\bibitem{DBLP:conf/iccv/ChatterjeeG13}
Avishek Chatterjee and Venu~Madhav Govindu.
\newblock Efficient and robust large-scale rotation averaging.
\newblock In {\em {ICCV}}, pages 521--528. {IEEE} Computer Society, 2013.

\bibitem{chatterjee2018robust}
Avishek Chatterjee and Venu~Madhav Govindu.
\newblock Robust relative rotation averaging.
\newblock {\em IEEE transactions on pattern analysis and machine intelligence},
  40(4):958--972, 2018.

\bibitem{DBLP:conf/icml/ChenGH14}
Yuxin Chen, Leonidas~J. Guibas, and Qi{-}Xing Huang.
\newblock Near-optimal joint object matching via convex relaxation.
\newblock In {\em ICML}, pages 100--108, 2014.

\bibitem{DBLP:journals/pami/ChoAF10}
Taeg~Sang Cho, Shai Avidan, and William~T. Freeman.
\newblock The patch transform.
\newblock {\em {IEEE} Trans. Pattern Anal. Mach. Intell.}, 32(8):1489--1501,
  2010.

\bibitem{conf/cvpr/ChoiZK15}
Sungjoon Choi, Qian-Yi Zhou, and Vladlen Koltun.
\newblock Robust reconstruction of indoor scenes.
\newblock In {\em CVPR}, pages 5556--5565. IEEE Computer Society, 2015.

\bibitem{Choi2016}
Sungjoon Choi, Qian-Yi Zhou, Stephen Miller, and Vladlen Koltun.
\newblock A large dataset of object scans.
\newblock {\em arXiv:1602.02481}, 2016.

\bibitem{dai2017scannet}
Angela Dai, Angel~X Chang, Manolis Savva, Maciej Halber, Thomas Funkhouser, and
  Matthias Nie{\ss}ner.
\newblock Scannet: Richly-annotated 3d reconstructions of indoor scenes.
\newblock In {\em Proc. IEEE Conf. on Computer Vision and Pattern Recognition
  (CVPR)}, volume~1, page~1, 2017.

\bibitem{Daubechies:2008:IRWa}
Ingrid Daubechies, Ronald DeVore, Massimo Fornasier, and C.~Sinan
  G{\"u}nt{\"u}rk.
\newblock Iteratively re-weighted least squares minimization for sparse
  recovery.
\newblock Report, Program in Applied and Computational Mathematics, Princeton
  University, Princeton, NJ, USA, June 2008.

\bibitem{fusiello2002model}
Andrea Fusiello, Umberto Castellani, Luca Ronchetti, and Vittorio Murino.
\newblock Model acquisition by registration of multiple acoustic range views.
\newblock In {\em European Conference on Computer Vision}, pages 805--819.
  Springer, 2002.

\bibitem{DBLP:conf/3dim/GelfandRIL03}
Natasha Gelfand, Szymon Rusinkiewicz, Leslie Ikemoto, and Marc Levoy.
\newblock Geometrically stable sampling for the {ICP} algorithm.
\newblock In {\em 3DIM}, pages 260--267. {IEEE} Computer Society, 2003.

\bibitem{govindu2014averaging}
Venu~Madhav Govindu and A Pooja.
\newblock On averaging multiview relations for 3d scan registration.
\newblock {\em IEEE Transactions on Image Processing}, 23(3):1289--1302, 2014.

\bibitem{Huang:2006:RFO}
Qixing Huang, Simon Fl\"{o}ry, Natasha Gelfand, Michael Hofer, and Helmut
  Pottmann.
\newblock Reassembling fractured objects by geometric matching.
\newblock {\em ACM Trans. Graph.}, 25(3):569--578, July 2006.

\bibitem{Huang:2013:CSM}
Qixing Huang and Leonidas Guibas.
\newblock Consistent shape maps via semidefinite programming.
\newblock In {\em Proceedings of the Eleventh Eurographics/ACMSIGGRAPH
  Symposium on Geometry Processing}, SGP '13, pages 177--186, Aire-la-Ville,
  Switzerland, Switzerland, 2013. Eurographics Association.

\bibitem{DBLP:journals/tog/HuangWG14}
Qixing Huang, Fan Wang, and Leonidas~J. Guibas.
\newblock Functional map networks for analyzing and exploring large shape
  collections.
\newblock {\em {ACM} Trans. Graph.}, 33(4):36:1--36:11, 2014.

\bibitem{DBLP:journals/tog/HuangZGHBG12}
Qi{-}Xing Huang, Guo{-}Xin Zhang, Lin Gao, Shi{-}Min Hu, Adrian Butscher, and
  Leonidas~J. Guibas.
\newblock An optimization approach for extracting and encoding consistent maps
  in a shape collection.
\newblock {\em {ACM} Trans. Graph.}, 31(6):167:1--167:11, 2012.

\bibitem{DBLP:conf/nips/HuangLBH17}
Xiangru Huang, Zhenxiao Liang, Chandrajit Bajaj, and Qixing Huang.
\newblock Translation synchronization via truncated least squares.
\newblock In {\em {NIPS}}, 2017.

\bibitem{Huber-2002-8601}
Daniel Huber.
\newblock {\em Automatic Three-dimensional Modeling from Reality}.
\newblock PhD thesis, Carnegie Mellon University, Pittsburgh, PA, December
  2002.

\bibitem{Huber01fullyautomatic}
Daniel~F. Huber and Martial Hebert.
\newblock Fully automatic registration of multiple 3d data sets.
\newblock {\em Image and Vision Computing}, 21:637--650, 2001.

\bibitem{Kadalbajoo:2011:OEC}
Mohan~K. Kadalbajoo and Ankit Gupta.
\newblock An overview on the eigenvalue computation for matrices.
\newblock {\em Neural, Parallel Sci. Comput.}, 19(1-2):129--164, Mar. 2011.

\bibitem{DBLP:conf/nips/KimLJMS18}
Seungryong Kim, Stephen Lin, SANG~RYUL JEON, Dongbo Min, and Kwanghoon Sohn.
\newblock Recurrent transformer networks for semantic correspondence.
\newblock In {\em {NIPS}}, page to appear, 2018.

\bibitem{Kim:2012:ECM}
Vladimir Kim, Wilmot Li, Niloy Mitra, Stephen DiVerdi, and Thomas Funkhouser.
\newblock Exploring collections of 3d models using fuzzy correspondences.
\newblock {\em ACM Trans. Graph.}, 31(4):54:1--54:11, July 2012.

\bibitem{Krizhevsky:2012:ICD}
Alex Krizhevsky, Ilya Sutskever, and Geoffrey~E. Hinton.
\newblock Imagenet classification with deep convolutional neural networks.
\newblock In {\em Proceedings of the 25th International Conference on Neural
  Information Processing Systems - Volume 1}, NIPS'12, pages 1097--1105, USA,
  2012. Curran Associates Inc.

\bibitem{DBLP:conf/icra/LeonardosZD17}
Spyridon Leonardos, Xiaowei Zhou, and Kostas Daniilidis.
\newblock Distributed consistent data association via permutation
  synchronization.
\newblock In {\em {ICRA}}, pages 2645--2652. {IEEE}, 2017.

\bibitem{Mellado:2014:SFG}
Nicolas Mellado, Dror Aiger, and Niloy~J. Mitra.
\newblock Super 4pcs fast global pointcloud registration via smart indexing.
\newblock {\em Comput. Graph. Forum}, 33(5):205--215, Aug. 2014.

\bibitem{Yi_2018_CVPR}
Kwang Moo~Yi, Eduard Trulls, Yuki Ono, Vincent Lepetit, Mathieu Salzmann, and
  Pascal Fua.
\newblock Learning to find good correspondences.
\newblock In {\em The IEEE Conference on Computer Vision and Pattern
  Recognition (CVPR)}, June 2018.

\bibitem{DBLP:journals/cgf/NguyenBWYG11}
Andy Nguyen, Mirela Ben{-}Chen, Katarzyna Welnicka, Yinyu Ye, and Leonidas~J.
  Guibas.
\newblock An optimization approach to improving collections of shape maps.
\newblock {\em Comput. Graph. Forum}, 30(5):1481--1491, 2011.

\bibitem{ozyesil2015robust}
Onur Ozyesil and Amit Singer.
\newblock Robust camera location estimation by convex programming.
\newblock In {\em Proceedings of the IEEE Conference on Computer Vision and
  Pattern Recognition}, pages 2674--2683, 2015.

\bibitem{DBLP:conf/nips/PachauriKSS14}
Deepti Pachauri, Risi Kondor, Gautam Sargur, and Vikas Singh.
\newblock Permutation diffusion maps {(PDM)} with application to the image
  association problem in computer vision.
\newblock In {\em {NIPS}}, pages 541--549, 2014.

\bibitem{DBLP:conf/nips/PachauriKS13}
Deepti Pachauri, Risi Kondor, and Vikas Singh.
\newblock Solving the multi-way matching problem by permutation
  synchronization.
\newblock In {\em {NIPS}}, pages 1860--1868, 2013.

\bibitem{DBLP:conf/eccv/RanftlK18}
Ren{\'{e}} Ranftl and Vladlen Koltun.
\newblock Deep fundamental matrix estimation.
\newblock In {\em Computer Vision - {ECCV} 2018 - 15th European Conference,
  Munich, Germany, September 8-14, 2018, Proceedings, Part {I}}, pages
  292--309, 2018.

\bibitem{sharp2004multiview}
Gregory~C Sharp, Sang~W Lee, and David~K Wehe.
\newblock Multiview registration of 3d scenes by minimizing error between
  coordinate frames.
\newblock {\em IEEE Transactions on Pattern Analysis and Machine Intelligence},
  26(8):1037--1050, 2004.

\bibitem{NIPS2016_6128}
Yanyao Shen, Qixing Huang, Nati Srebro, and Sujay Sanghavi.
\newblock Normalized spectral map synchronization.
\newblock In {\em {NIPS}}, pages 4925--4933, 2016.

\bibitem{Singer:2012:VDM}
Amit Singer and Hau tieng Wu.
\newblock Vector diffusion maps and the connection laplacian.
\newblock {\em Communications in Pure and Applied Mathematics}, 65(8), Aug.
  2012.

\bibitem{DBLP:conf/eccv/SunLHH18}
Yifan Sun, Zhenxiao Liang, Xiangru Huang, and Qixing Huang.
\newblock Joint map and symmetry synchronization.
\newblock In {\em Computer Vision - {ECCV} 2018 - 15th European Conference,
  Munich, Germany, September 8-14, 2018, Proceedings, Part {V}}, pages
  257--275, 2018.

\bibitem{DBLP:conf/iccv/SweeneySHTP15}
Chris Sweeney, Torsten Sattler, Tobias H{\"{o}}llerer, Matthew Turk, and Marc
  Pollefeys.
\newblock Optimizing the viewing graph for structure-from-motion.
\newblock In {\em {ICCV}}, pages 801--809. {IEEE} Computer Society, 2015.

\bibitem{torsello2011multiview}
Andrea Torsello, Emanuele Rodola, and Andrea Albarelli.
\newblock Multiview registration via graph diffusion of dual quaternions.
\newblock In {\em CVPR 2011}, pages 2441--2448. IEEE, 2011.

\bibitem{Wang:2013:IMA}
Lanhui Wang and Amit Singer.
\newblock Exact and stable recovery of rotations for robust synchronization.
\newblock {\em Information and Inference: A Journal of the IMA}, 2:145–193,
  December 2013.

\bibitem{conf/eccv/WilsonS14}
Kyle Wilson and Noah Snavely.
\newblock Robust global translations with 1dsfm.
\newblock In David~J. Fleet, Tomás Pajdla, Bernt Schiele, and Tinne
  Tuytelaars, editors, {\em ECCV (3)}, volume 8691 of {\em Lecture Notes in
  Computer Science}, pages 61--75. Springer, 2014.

\bibitem{conf/cvpr/ZachKP10}
Christopher Zach, Manfred Klopschitz, and Marc Pollefeys.
\newblock Disambiguating visual relations using loop constraints.
\newblock In {\em CVPR}, pages 1426--1433. IEEE Computer Society, 2010.

\bibitem{DBLP:conf/eccv/ZhouPK16}
Qian{-}Yi Zhou, Jaesik Park, and Vladlen Koltun.
\newblock Fast global registration.
\newblock In {\em Computer Vision - {ECCV} 2016 - 14th European Conference,
  Amsterdam, The Netherlands, October 11-14, 2016, Proceedings, Part {II}},
  pages 766--782, 2016.

\bibitem{DBLP:journals/corr/abs-1801-09847}
Qian{-}Yi Zhou, Jaesik Park, and Vladlen Koltun.
\newblock Open3d: {A} modern library for 3d data processing.
\newblock {\em CoRR}, abs/1801.09847, 2018.

\bibitem{DBLP:conf/cvpr/ZhouLYE15}
Tinghui Zhou, Yong~Jae Lee, Stella~X. Yu, and Alexei~A. Efros.
\newblock Flowweb: Joint image set alignment by weaving consistent, pixel-wise
  correspondences.
\newblock In {\em {CVPR}}, pages 1191--1200. {IEEE} Computer Society, 2015.

\bibitem{zhou2015multi}
Xiaowei Zhou, Menglong Zhu, and Kostas Daniilidis.
\newblock Multi-image matching via fast alternating minimization.
\newblock In {\em Proceedings of the IEEE International Conference on Computer
  Vision}, pages 4032--4040, 2015.

\end{thebibliography}
}

\appendix
\newpage

\section{Overview}

We organize this supplemental material as follows. In Section \ref{Section:More:Results}, we provide more detailed experimental results. 
In Section~\ref{Section:Proofs:Propositions}, we describe the technical proofs for all the propositions in the main paper. In Section~\ref{Section:datasets}, we show the scenes we used in this paper.
\section{More Experimental Results}
\label{Section:More:Results}

\subsection{More Visual Comparison Results}

\begin{figure*}
\setlength\tabcolsep{1.75pt}
\begin{tabular}{cccc}
Ground Truth & RotAvg & Geometric Registration & Our Approach \\
\includegraphics[width=0.24\textwidth]{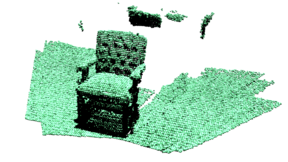}
&
\includegraphics[width=0.24\textwidth]{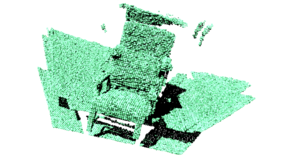}
&
\includegraphics[width=0.24\textwidth]{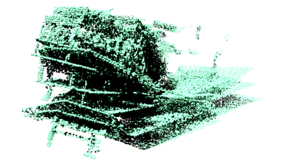}
&
\includegraphics[width=0.24\textwidth]{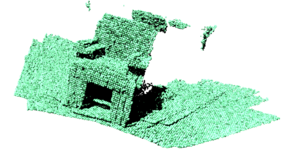}
\\
\includegraphics[width=0.24\textwidth]{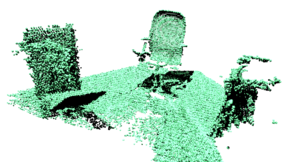}
&
\includegraphics[width=0.24\textwidth]{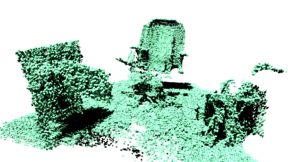}
&
\includegraphics[width=0.24\textwidth]{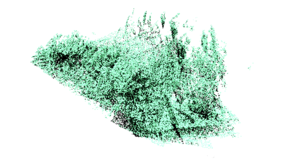}
&
\includegraphics[width=0.24\textwidth]{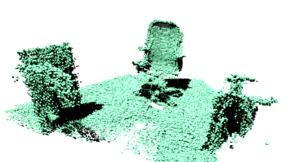}
\\
\includegraphics[width=0.24\textwidth]{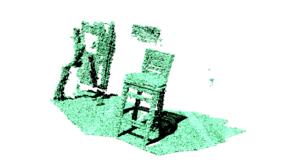}
&
\includegraphics[width=0.24\textwidth]{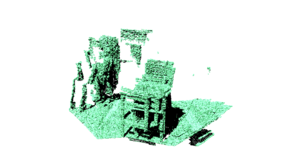}
&
\includegraphics[width=0.24\textwidth]{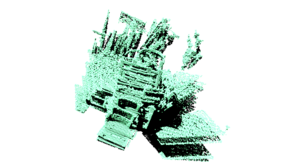}
&
\includegraphics[width=0.24\textwidth]{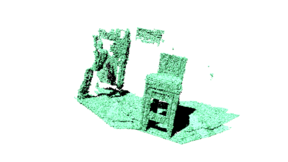}
\\
\includegraphics[width=0.24\textwidth]{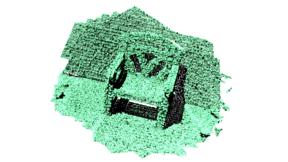}
&
\includegraphics[width=0.24\textwidth]{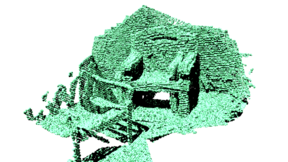}
&
\includegraphics[width=0.24\textwidth]{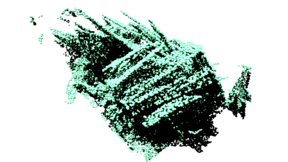}
&
\includegraphics[width=0.24\textwidth]{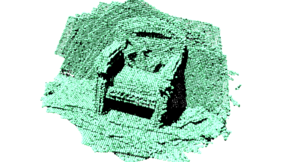}
\\
\includegraphics[width=0.24\textwidth]{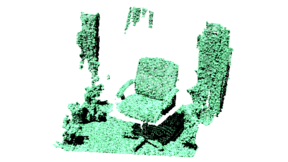}
&
\includegraphics[width=0.24\textwidth]{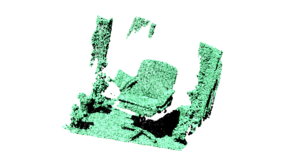}
&
\includegraphics[width=0.24\textwidth]{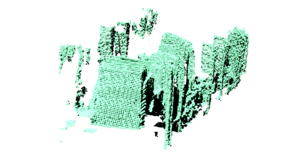}
&
\includegraphics[width=0.24\textwidth]{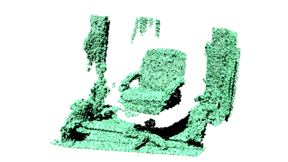}
\\
\includegraphics[width=0.24\textwidth]{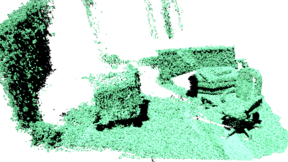}
&
\includegraphics[width=0.24\textwidth]{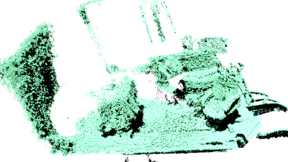}
&
\includegraphics[width=0.24\textwidth]{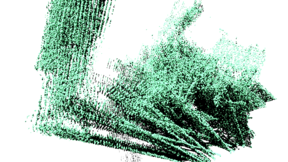}
&
\includegraphics[width=0.24\textwidth]{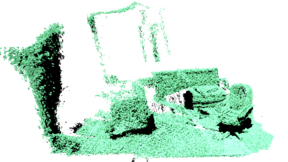}
\\
\includegraphics[width=0.24\textwidth]{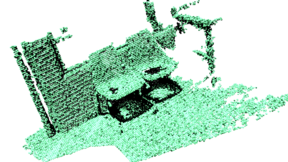}
&
\includegraphics[width=0.24\textwidth]{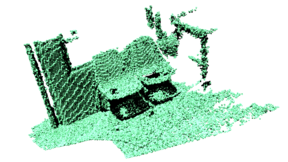}
&
\includegraphics[width=0.24\textwidth]{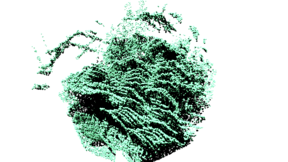}
&
\includegraphics[width=0.24\textwidth]{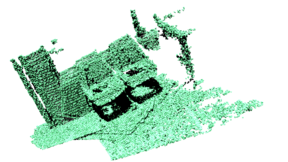}
\\
\hline 
\end{tabular}
\caption{We show the results of ground truth result (column I), Rotation Averaging~\protect\cite{chatterjee2018robust}+Translation Sync.~\protect\cite{DBLP:conf/nips/HuangLBH17}~ (column II), Geometric  Registration~\protect\cite{conf/cvpr/ChoiZK15} (column III), and Our Approach (column IV). These scenes are from Redwood Chair dataset.}
\label{Figure:More:Results}
\vspace{-0.15in}
\end{figure*}

Figure~\ref{Figure:More:Results} shows more visual comparisons between our approach and baseline approaches. Again, our approach produces alignments that are close to the underlying ground-truth. The overall quality of our alignments is superior to that of the baseline approaches. 

\subsection{Cumulative Density Function}

\begin{figure*}[h!]
\setlength\tabcolsep{1.75pt}
\begin{tabular}{cc}
Redwood & Scannet \\
\includegraphics[width=0.48\textwidth]{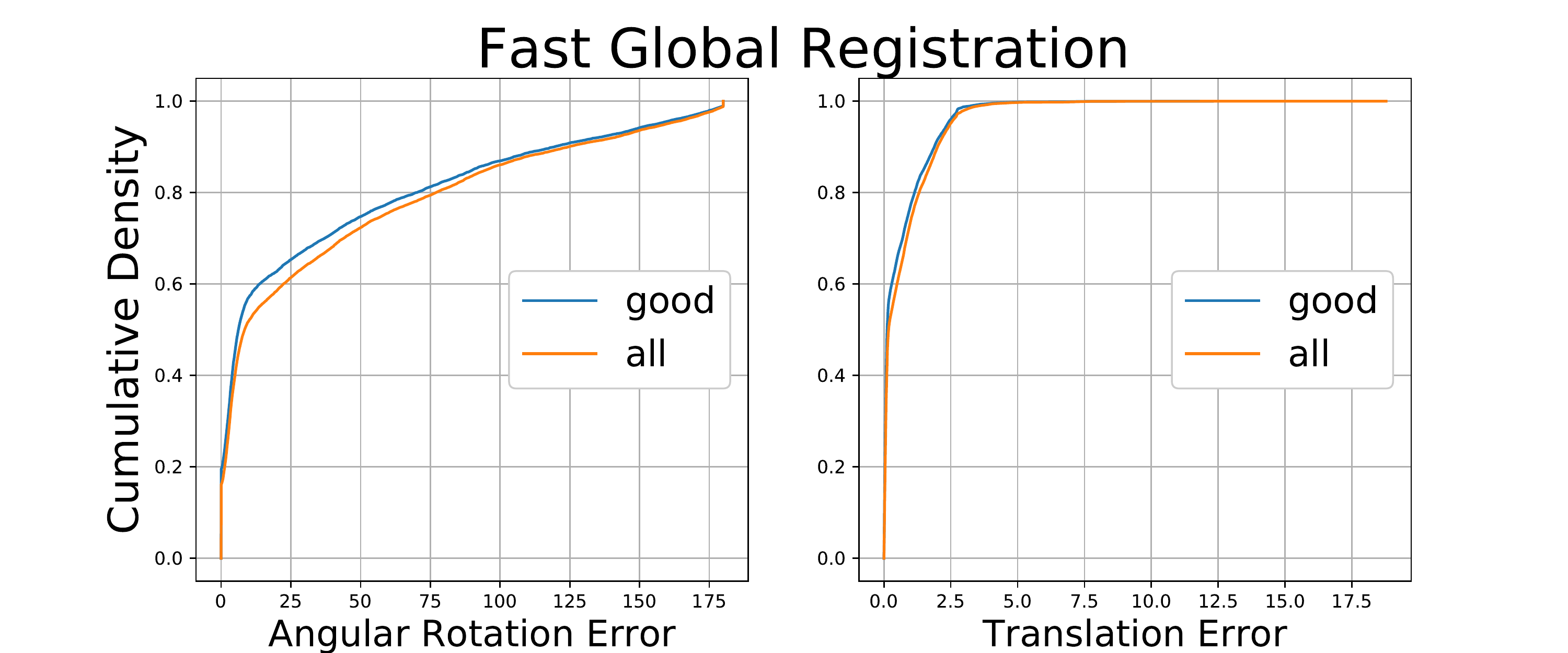}
&
\includegraphics[width=0.48\textwidth]{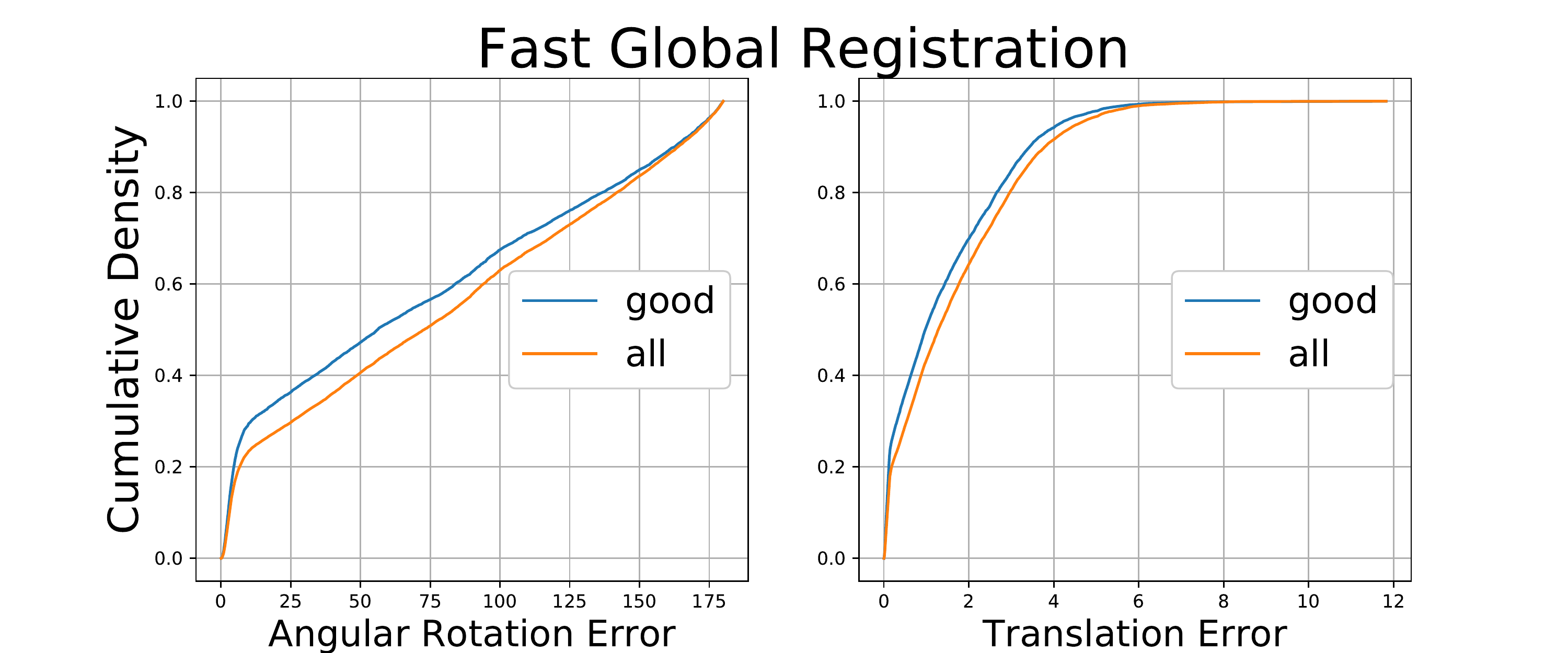}
\\
\includegraphics[width=0.48\textwidth]{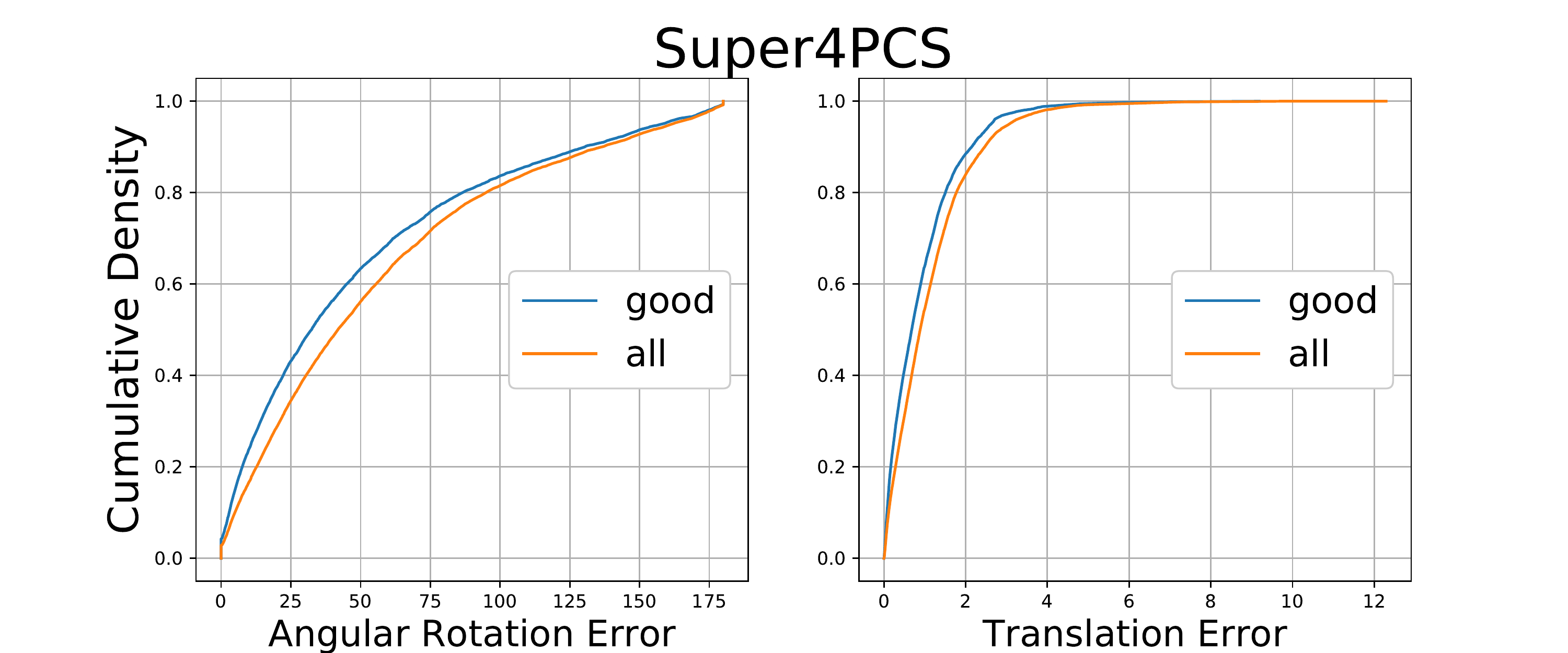}
&
\includegraphics[width=0.48\textwidth]{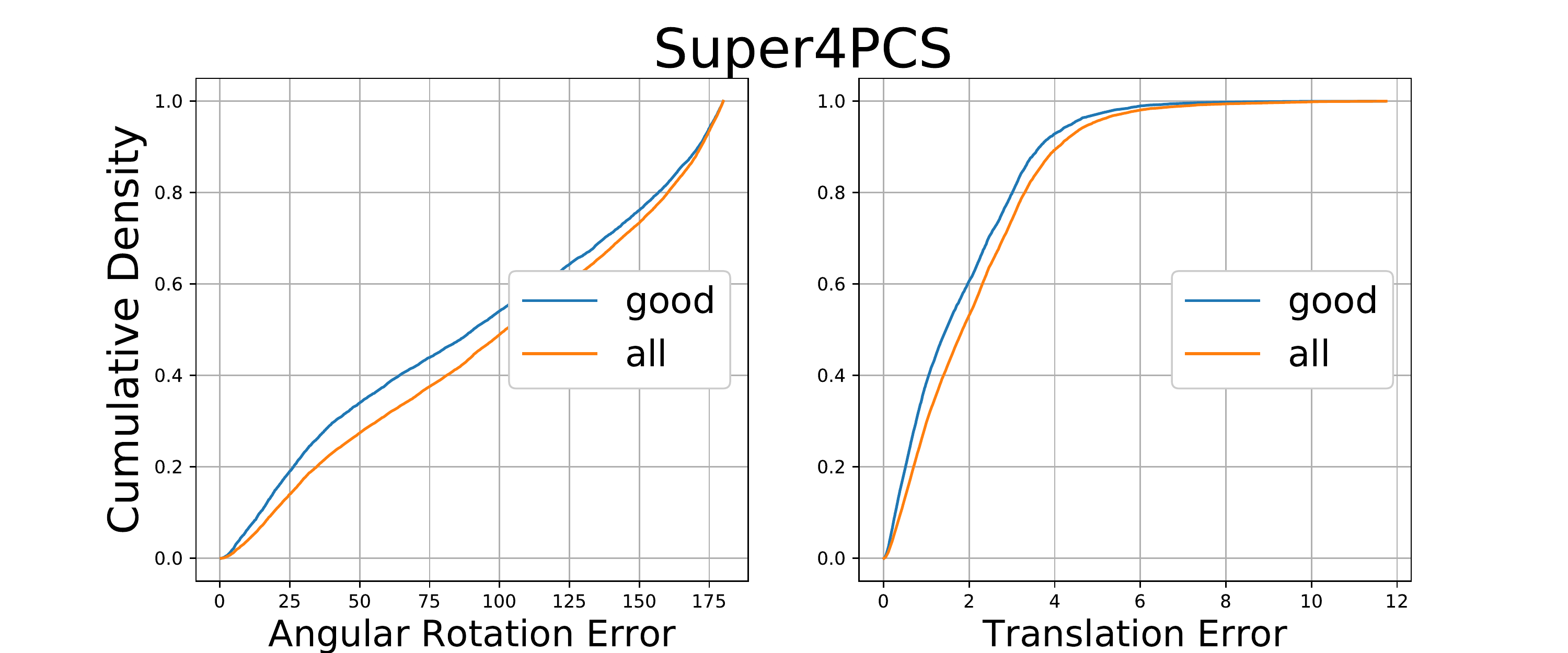}
\\
\hline 
Redwood & Scannet \\
\includegraphics[width=0.48\textwidth]{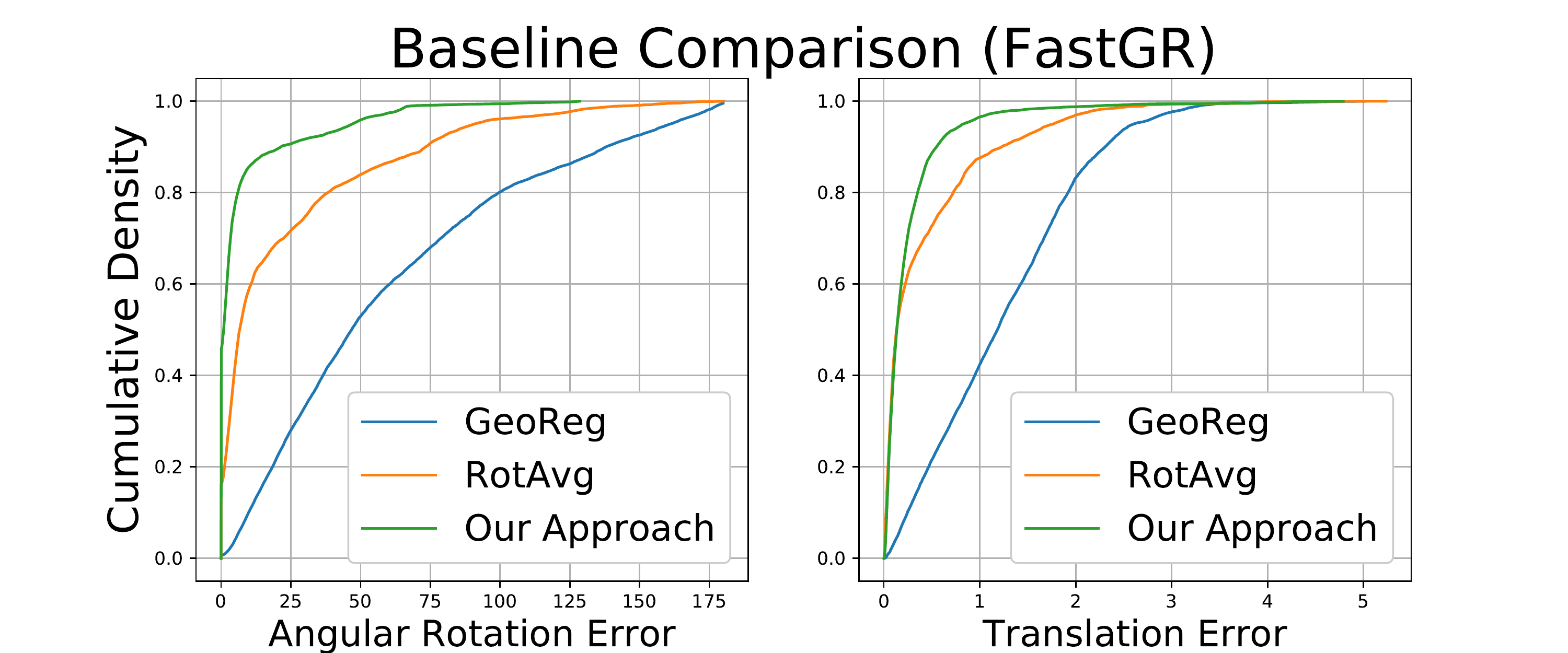}
&
\includegraphics[width=0.48\textwidth]{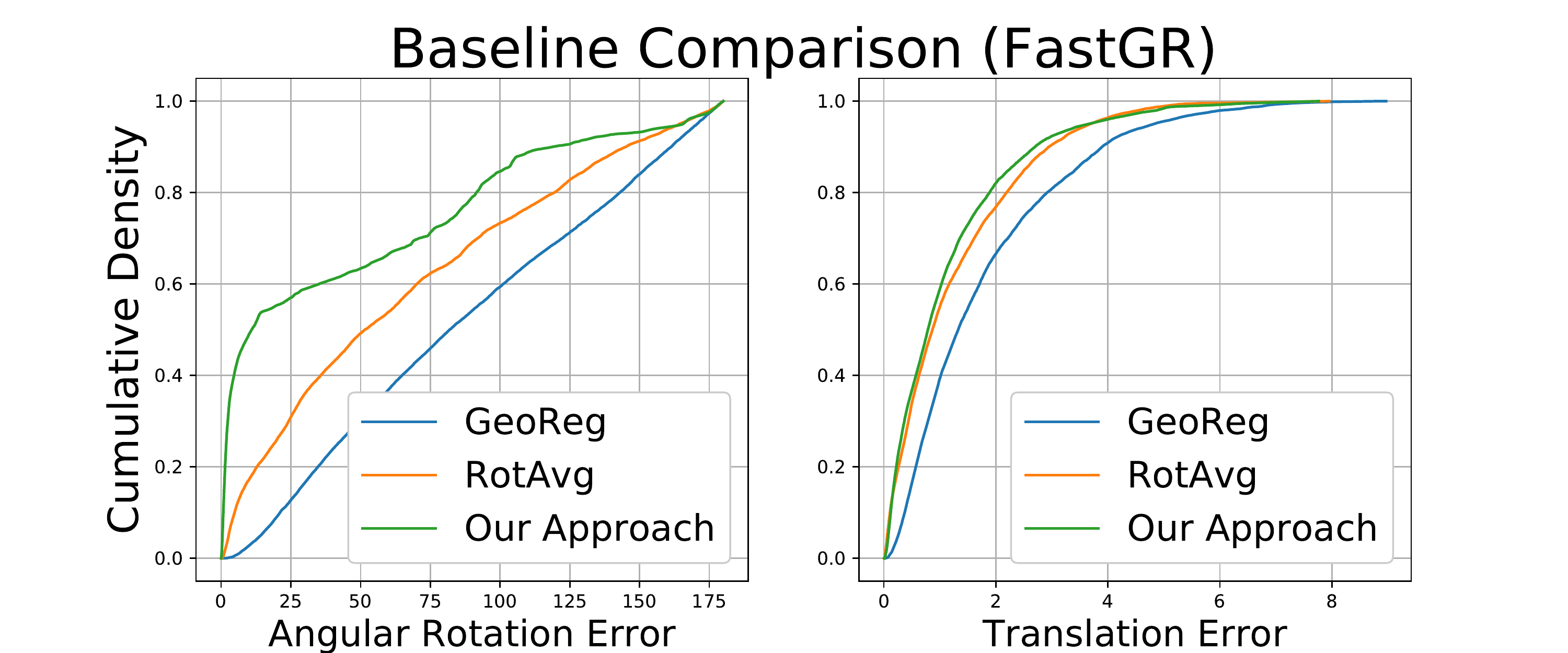}
\\
\includegraphics[width=0.48\textwidth]{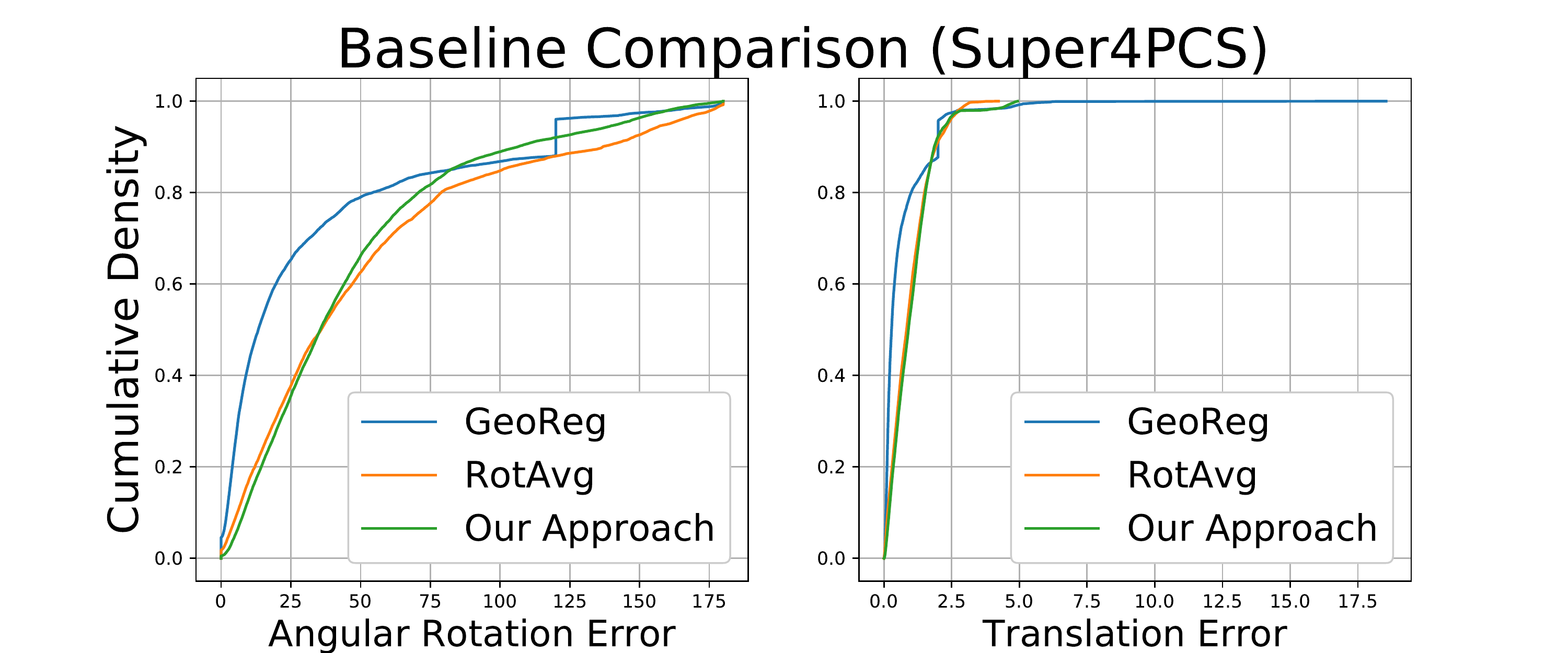}
&
\includegraphics[width=0.48\textwidth]{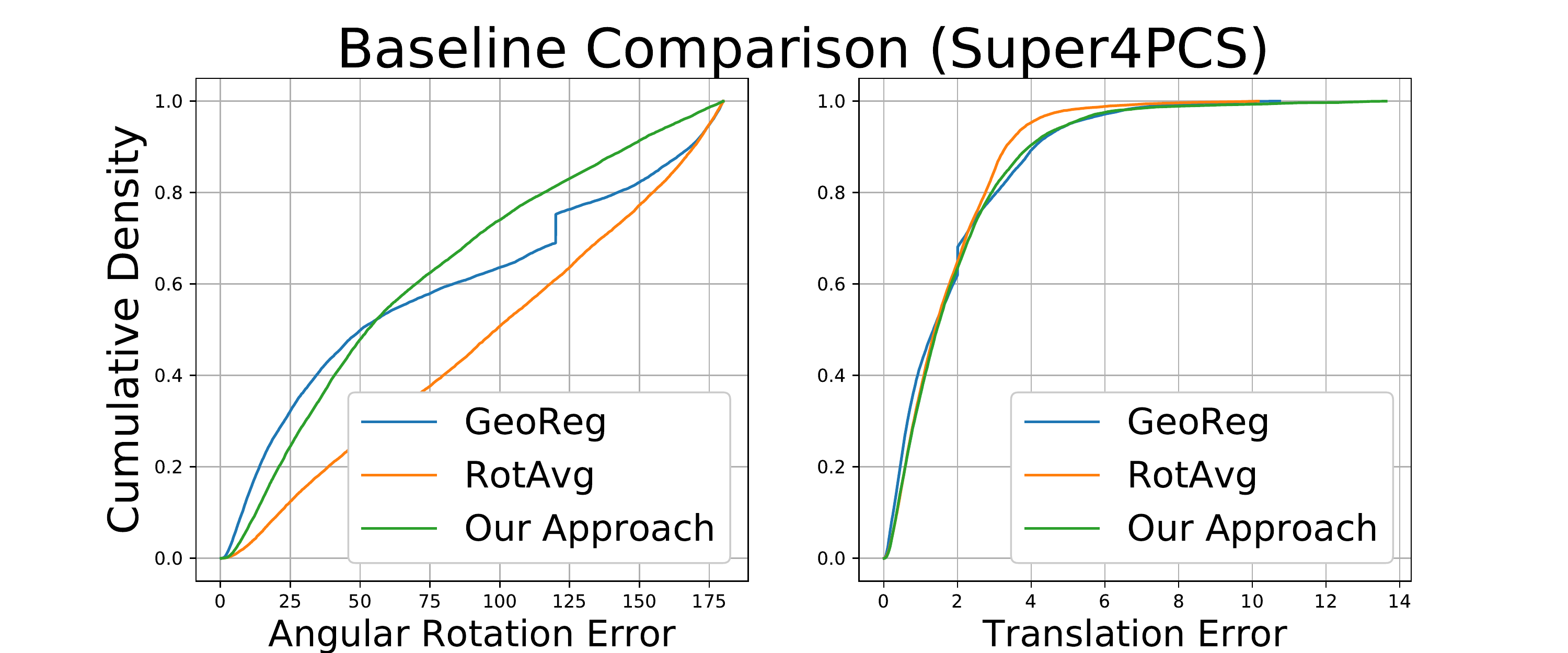}
\\
\hline
Redwood & Scannet \\
\includegraphics[width=0.48\textwidth]{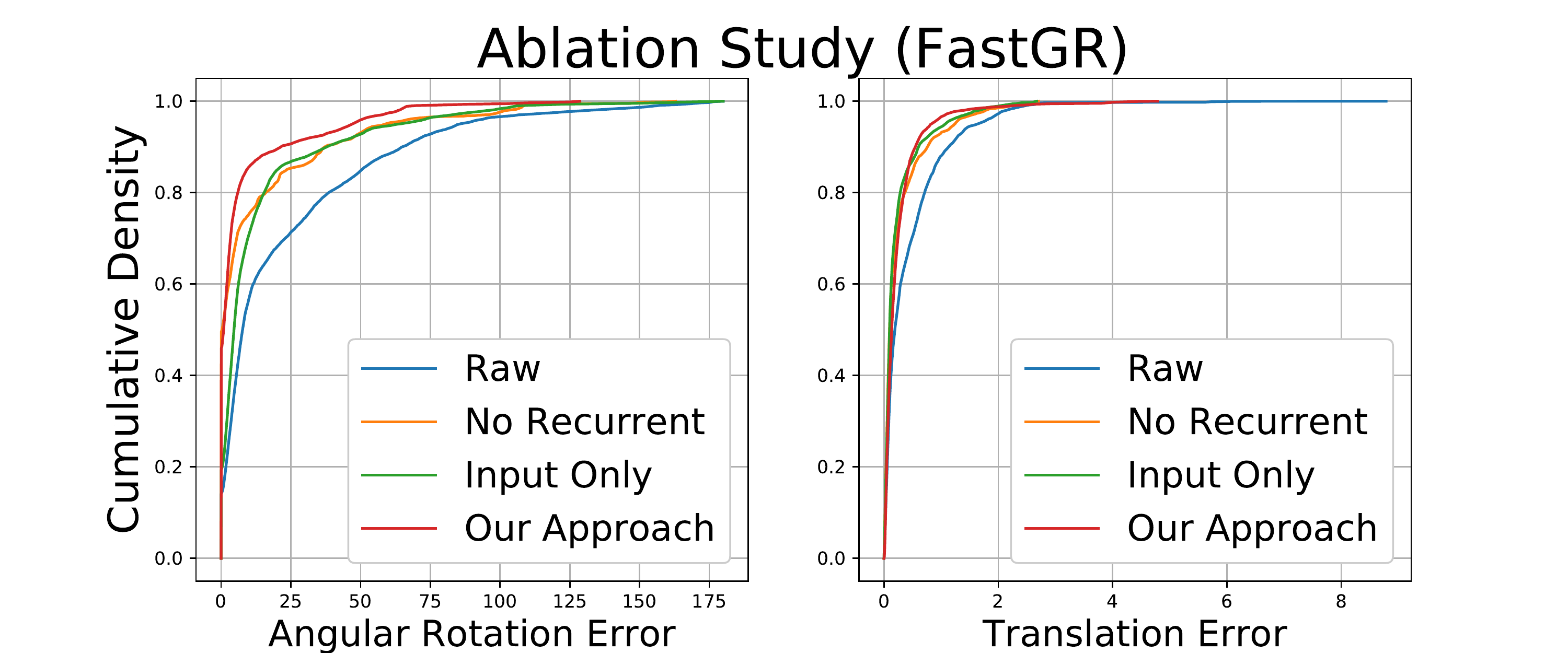}
&
\includegraphics[width=0.48\textwidth]{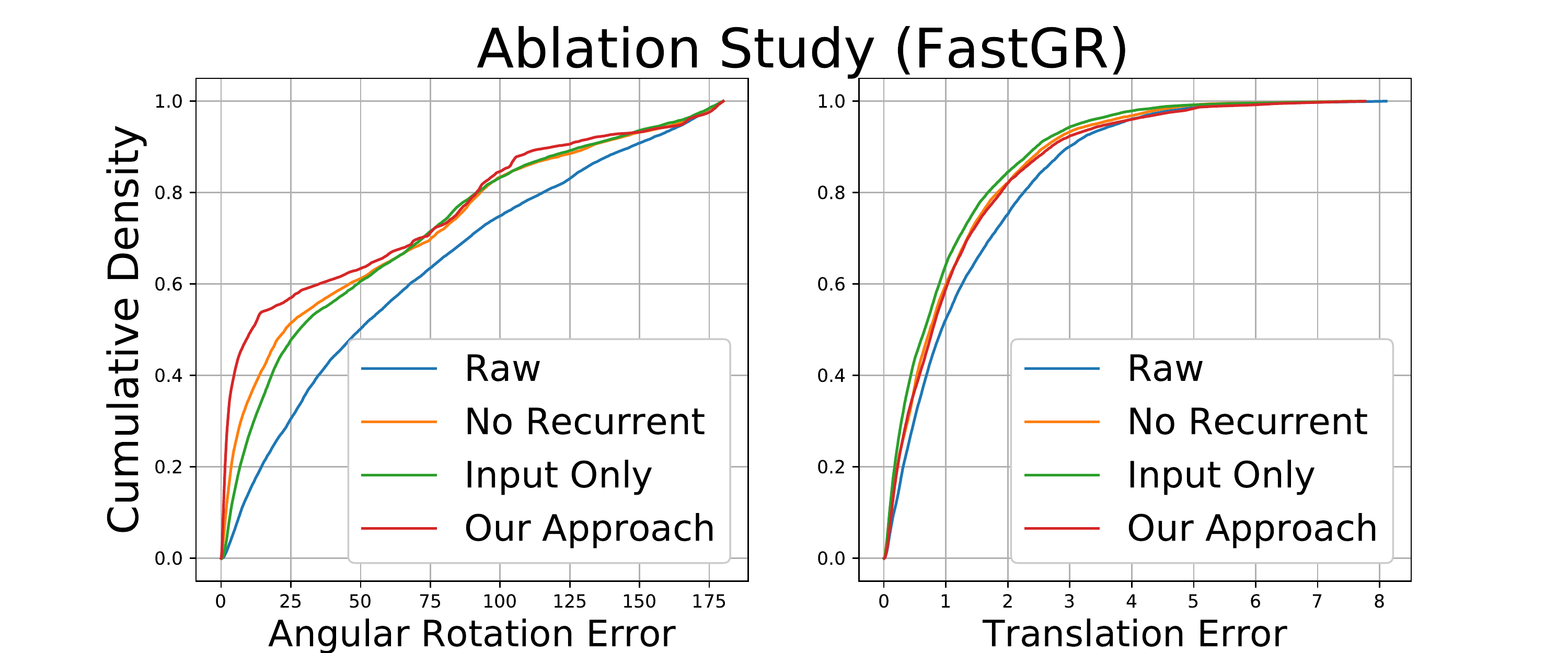}
\\
\end{tabular}
\caption{Corresponding cumulative density function (CDF) curves. For the top block, we plot CDF from different input sources. 
Here "all" corresponds to errors between all pairs and
"good" corresponds to errors between selected pairs. 
The pairs were selected by 1) computing ICP refinement, 2) computing overlapping region by finding points in source point clouds that are close to target point clouds (i.e. by setting a threshold), 
3) for these points, we compute their median distance to the target point clouds. 
For the middle block, we report the comparison of baselines and our approach. Results from different input sources are reported separately. 
For the bottom block, we report the comparison between variants of our approach using Fast Global Registration as the input pairwise alignments.}
\label{Figure:More:Results2}
\vspace{-0.15in}
\end{figure*}

Figure~\ref{Figure:More:Results2} plots the cumulative density functions of errors in rotations and translations with respect to a varying threshold. 

\subsection{Illustration of Dataset}

\begin{figure*}[h!]

\begin{tabular}{ccccc}
    \includegraphics[width=0.32\textwidth]{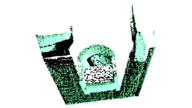}&
    \includegraphics[width=0.32\textwidth]{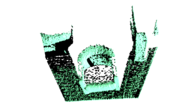}&
    \includegraphics[width=0.32\textwidth]{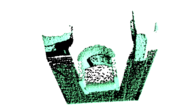}\\
    \includegraphics[width=0.32\textwidth]{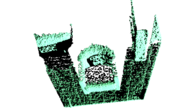}&
    \includegraphics[width=0.32\textwidth]{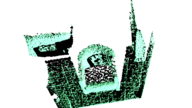}&
    \includegraphics[width=0.32\textwidth]{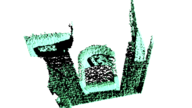}\\
    \includegraphics[width=0.32\textwidth]{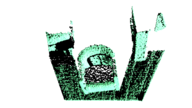}&
    \includegraphics[width=0.32\textwidth]{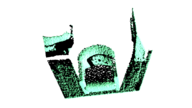}&
    \includegraphics[width=0.32\textwidth]{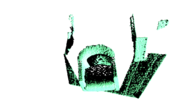}\\
    \includegraphics[width=0.32\textwidth]{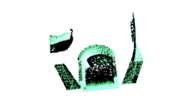}&
    \includegraphics[width=0.32\textwidth]{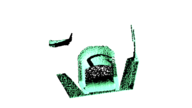}&
    \includegraphics[width=0.32\textwidth]{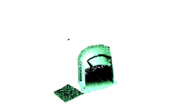}\\
    \includegraphics[width=0.32\textwidth]{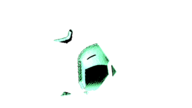}&
    \includegraphics[width=0.32\textwidth]{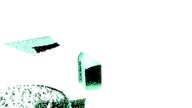}&
    \includegraphics[width=0.32\textwidth]{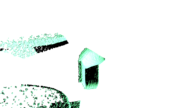}\\
\end{tabular}

\caption{A typical example of the a Redwood Chair scene: the 1st, 3rd, 5th, 7th, $\ldots$, 29th of the selected scans are rendered from the same camera view point. Each scan is about 40 frames away from the next one.}
\label{fig:example:redwood}
\end{figure*}

To understand the difficulty of the datasets used in our experiments, we pick a typical scene from each of the Redwood and ScanNet datasets and render 15 out of 30 ground truth point clouds from the same camera view point. From Figure~\ref{fig:example:scannet} and Figure ~\ref{fig:example:redwood}, we can see that ScanNet is generally harder than Redwood, as there is less information that can be extracted by looking at pairs of scans.

\begin{figure*}[h!]

\begin{tabular}{ccccc}
    \includegraphics[width=0.32\textwidth]{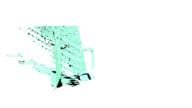}&
    \includegraphics[width=0.32\textwidth]{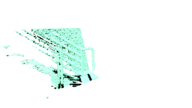}&
    \includegraphics[width=0.32\textwidth]{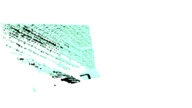}\\
    \includegraphics[width=0.32\textwidth]{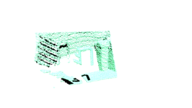}&
    \includegraphics[width=0.32\textwidth]{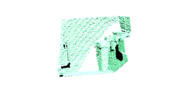}&
    \includegraphics[width=0.32\textwidth]{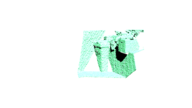}\\
    \includegraphics[width=0.32\textwidth]{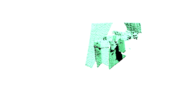}&
    \includegraphics[width=0.32\textwidth]{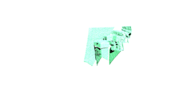}&
    \includegraphics[width=0.32\textwidth]{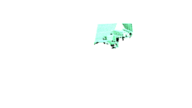}\\
    \includegraphics[width=0.32\textwidth]{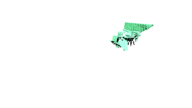}&
    \includegraphics[width=0.32\textwidth]{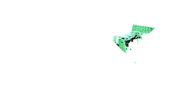}&
    \includegraphics[width=0.32\textwidth]{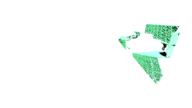}\\
    \includegraphics[width=0.32\textwidth]{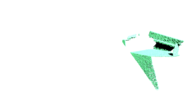}&
    \includegraphics[width=0.32\textwidth]{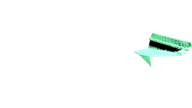}&
    \includegraphics[width=0.32\textwidth]{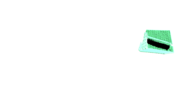}\\
\end{tabular}

\caption{A typical example of the a ScanNet scene: the 1st, 3rd, 5th, 7th, $\ldots$, 29th of the selected scans are rendered from the same camera view point. Each scan is about 40 frames away from the next one.}
\label{fig:example:scannet}
\end{figure*}

\section{Proofs of Propositions}
\label{Section:Proofs:Propositions}

We organize this section as follows. In Section~\ref{Section:Connection:Laplacian:Stability}, we provide key lemmas regarding the eigen-decomposition of a connection Laplacian, including stability of eigenvalues/eigenvectors and derivatives of eigenvectors with respect to elements of the connection Laplacian. In Section~\ref{Section:SO:Projection:Operator}, we provide key lemmas regarding the projection operator that maps the space of square matrices to the space of rotations.  Section~\ref{Section:Prop:4:1} to Section~\ref{Section:Prop:4:4} describe the proofs of all the propositions stated in the main paper. Section~\ref{Section:Exact:Recovery:Rot:Sync} provides an exact recovery condition of a rotation synchronization scheme via reweighted least squares.  Finally, Section~\ref{Section:Proof:Lemma} provides proofs for new key lemmas introduced in this section.

\subsection{Eigen-Stability of Connection Laplacian}
\label{Section:Connection:Laplacian:Stability}

We begin with introducing the problem setting and notations in Section~\ref{Section:Key:Lemma:Notation}. We then present the key lemmas in Section~\ref{Section:Key:Lemmas}.

\subsubsection{Problem Setting and Notations}
\label{Section:Key:Lemma:Notation}

Consider a weighted graph $\set{G} = (\set{V}, \set{E})$ with $n$ vertices, i.e., $|\set{V}|=n$. We assume that $\set{G}$ is connected. With $w_{ij}>0$ we denote an edge weight associated with edge $(i,j)\in \set{E}$. Let $\overline{L}$ be the weighted adjacency matrix (Note that we drop $\bs{w}$ from the expression of $\overline{L}$ to make the notations uncluttered). It is clear that the leading eigenvector of $\overline{L}$ is $\frac{1}{\sqrt{n}}\bs{1}\in \R^n$, and its corresponding eigenvalue is zero. In the following, we shall denote the eigen-decomposition of $\overline{L}$ as
$$
\overline{L} = \overline{U}\overline{\Lambda}\overline{U}^{T},
$$
where
$$
\overline{U} = (\overline{\bs{u}}_2,\cdots, \overline{\bs{u}}_n) \ \textup{and} \ \overline{\Lambda} = \diag(\overline{\lambda}_2,\cdots, \overline{\lambda}_n)
$$
collect the remaining eigenvectors and their corresponding eigenvalues of $L(\bs{w})$, respectively. Our analysis will also use a notation that is closely related to the pseudo-inverse of $\overline{L}$:
\begin{equation}
\overline{L}_{t}^{+}:= \overline{U}(\overline{\Lambda} + t I_{n-1})^{-1} \overline{U}^{T}, \quad \forall |t| < \overline{\lambda}_2.
\end{equation}

Our goal is to understand the behavior of the leading eigenvectors of $\overline{L}\otimes I_k + E$\footnote{Note that when applying the stability results to the problem studied in this paper, we always use $k=3$. However, when assume a general $k$ when describing the stability results.} for a symmetric perturbation matrix $E\in \R^{nk\times nk}$, which is a $n\times n$ block matrix whose blocks are given by
$$
E_{ij} = \left\{
\begin{array}{cc}
0 & i = j \\
-w_{ij}N_{ij} & (i,j) \in \set{E}
\end{array}
\right.\
$$
where $N_{ij}$ is the perturbation imposed on $R_{ij}$.

We are interested in $U\in \R^{nk\times k}$, which collects the leading $k$ eigenvectors of $\overline{L}\otimes I_k + E$ in its columns. With $\lambda_1\leq \lambda_2 \cdots \leq \lambda_k$ we denote the corresponding eigenvalues. Note that due to the property of connection Laplacian, $\lambda_i \geq 0,1\leq i \leq k$. Our goal is to 1) bound the eigenvalues $\lambda_i, 1\leq i \leq k$, and 2) to provide block-wise bounds between $U$ and $\frac{1}{\sqrt{n}}\bs{1}\otimes Q$, for some rotation matrix $Q\in SO(k)$. 


Besides the notations introduced above that are related to Laplacian matrices, we shall also use a few matrix norms. With $\|\cdot\|$ and $\|\cdot\|_{\set{F}}$ we denote the spectral norm and Frobenius norm, respectively. Given a vector $\bs{v}\in \R^n$, we denote $\|\bs{v}\|_{\infty} = \max\limits_{1\leq i \leq n}|v_i|$ as the element-wise infinity norm. We will also introduce a norm $\|\cdot\|_{1,\infty}$ for square matrices, which is defined as 
$$
\|A\|_{1,\infty} = \max\limits_{1\leq i \leq n} \sum\limits_{j=1}^n |a_{ij}|, \quad \forall A = (a_{ij})_{1\leq i,j \leq n} \in \R^{n \times n}.
$$
We will also use a similar norm defined for $n\times n$ block matrices $E\in \R^{nk\times nk}$ (i.e., each block is a $k\times k$ matrix):
$$
\|E\|_{1,\infty} = \max\limits_{1\leq i \leq n} \sum\limits_{j=1}^n \|E_{ij}\|, \quad \forall E = (E_{ij})_{1\leq i,j \leq n} \in \R^{nk \times nk}.
$$

\subsubsection{Key Lemmas}
\label{Section:Key:Lemmas}

This section presents a few key lemmas that will be used to establish main stability results regarding matrix eigenvectors and matrix eigenvalues. We begin with the classical result of the Weyl's inequality:
\begin{lem}
(\textbf{Eigenvalue stability}) For $1 \leq i \leq k$, we have
\begin{equation}
\lambda_i \leq \|E\|.
\end{equation}
\label{Lemma:Eigenvalue:Stability}
\end{lem}
We proceed to describe tools for controlling the eigenvector stability. To this end, we shall rewrite $U$ as follows:
$$
U = \frac{1}{\sqrt{n}}\bs{1}\otimes X + Y.
$$
Our goal is to bound the deviation between $X$ and a rotation matrix and blocks of $Y$.

We begin with controlling $X$, which we adopt a result described in \cite{DBLP:conf/icml/BajajGHHL18}:
\begin{lem}
(\textbf{Controlling $X$}\cite{DBLP:conf/icml/BajajGHHL18}) If
$$
\|E\|<\frac{\overline{\lambda}_2}{2},
$$
then there exists $Q \in SO(k)$\footnote{If not, we can always negate the last column of U.} such that
$$
\|X - Q\| \leq 1 - \sqrt{1 - \left(\frac{\|E\|}{\overline{\lambda}_2- \|E\|}\right)^2}.
$$
In particular,
$$
\|X - Q\| \leq \left(\frac{\|E\|}{\overline{\lambda}_2 - \|E\|}\right)^2
$$
\label{Lemma:Eigenvector:L2:Stability}
\end{lem}

It remains to control the blocks of $Y$. We state a formulation that expresses the column of $Y$ using a series:
\begin{lem}
Suppose $\|E\| < \frac{\overline{\lambda}_2}{2}$, then $\forall 1\leq j \leq k$,
\begin{equation}
Y \bs{e}_j^{(k)} = -\frac{1}{\sqrt{n}}\sum\limits_{l=1}^{\infty}\Big((\overline{L}_{-\lambda_j}^{+}\otimes I_k)E\Big)^l (\bs{1}\otimes X)\bs{e}_j^{(k)}.
\label{Y:expression}
\end{equation}
\label{Lemma:Y:Expression}
\end{lem}

We conclude this section by providing an explicit expression for computing the derivative of the leading eigenvectors of a connection Laplacian with its elements:
\begin{lem}
Let $L$ be an $N\times N$ non-negative definite matrix and its eigen-decomposition is
\begin{equation}
L=\sum_{i=1}^N \lambda_i \bs{u}_i\bs{u}_i^T
\label{eq:derivative:decomposition}
\end{equation}
where $0\leq \lambda_1\leq \lambda_2\leq \dots\lambda_N$.

Suppose $\lambda_{k}<\lambda_{k+1}$. Collect the eigenvectors corresponding to the smallest $k$ eigenvalues of $L$ as the columns of matrix $U_k$. Namely, $U_k=[\bs{u}_1,\dots,\bs{u}_k]$ where $\sigma_1,\dots,\sigma_k$ are the smallest $k$ eigenvelues of $L$.

Notice that $L$ can have different decompositions in (\ref{eq:derivative:decomposition}) when there are repetitive eigenvalues. But in our case where $\lambda_k<\lambda_{k+1}$, we claim that $U_k U_k^T$ is unique under different possible decomposition of $L$ so that $\dd(U_kU_k^T)$ is well-defined and has an explicit expression:
\begin{equation}
    \dd (U_kU_k^T)=\sum_{i=1}^k\sum_{j=k+1}^{N}\frac{\bs{u}_j^T\dd L\bs{u}_i}{\sigma_i-\sigma_j}(\bs{u}_i\bs{u}_j^T+\bs{u}_j\bs{u}_i^T)
\end{equation}
Moreover, the differentials of eigenvalues are
\begin{equation}
    \dd \sigma_i=\bs{u}_i^T\dd L\bs{u}_i.
    \label{eq:derivative:eigenvalue}
\end{equation}
\label{Lemma:Laplacian:Derivative}
\end{lem}

\subsection{Key Lemma Regarding the Projection Operator}
\label{Section:SO:Projection:Operator}

This section studies the projection operator which maps the space of square matrices to the space of rotation matrices. We begin with formally defining the projection operator as follows:

\begin{definition}
Suppose $det(M) > 0$. Let $M=\sum_{i=1}^n\sigma_i\bs{u}_i\bs{v}_i^T$ be the singular value decomposition of square matrix $M$ where $U=[\bs{u}_1,\dots,\bs{u}_1]$ and $V=[\bs{v}_1,\dots,\bs{v}_n]$ are both orthogonal matrices, and all coefficients $\sigma_i$ are non-negative. Then we define the rotation approximation of $M$ as
$$
R(M):=\sum_{i=1}^n\bs{u}_i\bs{v}_i^T=UV^T.
$$
It is clear that $R(M)$ is a rotation matrix, since 1) both $U$ and $V^T$ are rotations, and 2) $\det(UV^{T}) > 0$.
\label{def:rotation}
\end{definition}

\begin{lem}
Let $A\in\mathbb{R}^{nk\times k}$ be a block matrix of form
$$
A=\begin{bmatrix}
A_1\\
\vdots\\
A_n
\end{bmatrix}
$$
where $A_i\in\mathbb{R}^{k\times k}$. Use $a_{ij}$ to denote the element on position $i,j$ in $A$. Then we have
$$
\sum_{i=1}^n\|A_i\|^2\leq k\|A\|^2
$$
\label{Lemma:Block:Norm}
\end{lem}

We then present the following key lemma regarding the stability of the projection operator:
\begin{lem}
Let $M$ be a square matrix and $\epsilon=\|M-I\|$. Suppose $\epsilon< \frac{1}{3}$, then
$$
\|R(M)-I\|\leq \epsilon+\epsilon^2.
$$
\label{Lemma:Projection:Stability}
\end{lem}

\begin{lem}
Regarding $R(M)$ as a function about $M$, then the differential of $R(M)$ would be
$$
\dd R(M)=\sum_{i\neq j}\frac{\bs{u}_i^T\dd M\bs{v}_j-\bs{u}_j^T\dd M\bs{v}_i}{\sigma_i+\sigma_j}\bs{u}_i\bs{v}_j^T
$$
where all notations follow Definition (\ref{def:rotation}).
\label{prop:derivative:single_rotation}
\end{lem}

\subsection{Robust Recovery of Rotations}
\label{Section:Prop:4:1}

We state the following result regarding robust recovery of rotations using the connection:
\begin{proposition}
Suppose the underlying rotations are given by $R_i^{\star}, 1\leq i \leq n$. Modify the definition of $E$ such that 
$$
E_{ij} = \left\{
\begin{array}{cc}
-w_{ij}R_j^{\star}({R_j^{\star}}^T R_{ij}R_i^{\star}-I_k){R_i^{\star}}^T & (i,j)\in \set{E} \\
0 & \textup{otherwise}
\end{array}
\right.\
$$
Define 
\begin{equation}
\epsilon_1: = \frac{2\|E\|_{1,\infty}}{\overline{\lambda}_2},\quad \epsilon_2: = \|\overline{L}^{+}\|_{1,\infty}\|E\|_{1,\infty}.
\end{equation}
Suppose $\epsilon_1 < 1$, $\epsilon_2 < 1$, and
$$
\delta := \Big(\frac{\epsilon_1}{2-\epsilon_1}\Big)^2 + \sqrt{k}\cdot \Big(1+\big(\frac{\epsilon_1}{2-\epsilon_1}\big)^2\Big)\cdot \frac{\epsilon_2(1+\epsilon_2)}{1-\epsilon_2(1+\epsilon_2)}< \frac{1}{3}.
$$
Then the optimal solution $R_i, 1\leq i \leq n$ to the rotation synchronization step satisfies that there exists $Q\in SO(k)$,
\begin{equation}
\max\limits_{1\leq i \leq n}\|R_i-R_i^{\star}Q\|\leq \delta + \delta^2.
\end{equation}
\label{Prop:4:1:Formal}
\end{proposition}

\noindent\textbf{Proof of Prop.~\ref{Prop:4:1:Formal}:} Without losing generality, we assume $R_i^{\star} = I_k, 1\leq i \leq k$ when proving Prop.~\ref{Prop:4:1:Formal}. In fact, we can always apply an unitary transform to obtain $\diag({R_1^{\star}},\cdots,{R_n^{\star}})^TL\diag({R_1^{\star}},\cdots,{R_n^{\star}})$, which does not impact the structure of the eigen-decomposition, and which satisfies the assumption. 

Before proving Prop.\ref{Prop:4:1:Formal}, we shall utilize two Lemmas, whose proofs are deferred to Section~\ref{Section:Proof:Lemma}.
\begin{lem}
Under the assumptions described above, we have
\begin{equation}
\|\overline{L}_{-\lambda_j}^{+}\|_{1,\infty}\leq \|\overline{L}^{+}\|_{1,\infty} (1 + \|\overline{L}^{+}\|_{1,\infty}\|E\|_{1,\infty}).
\label{Eq:recovery}
\end{equation}
\label{Lemma:Axx:1}
\end{lem}
\begin{lem}
Given a $k\times k$ matrix $A$, we have
\begin{equation}
\|A\| \leq \sqrt{k}\max\limits_{1\leq j \leq k} \|A\bs{e}_j^{(k)}\|.
\end{equation}
\label{Lemma:Lose:Bound}
\end{lem}

Now we proceed to complete the proof of Prop.\ref{Prop:4:1:Formal}. First of all, applying Lemma~\ref{Lemma:Eigenvector:L2:Stability}, we obtain that there exists $Q\in SO(k)$ such that
\begin{equation}
\|X-Q\|\leq \big(\frac{\epsilon_1}{2-\epsilon_1}\big)^2.
\label{Eq:Prop:1:Bound:1}
\end{equation}
Applying Lemma~\ref{Lemma:Y:Expression}, we have $\forall 1\leq j \leq k$,
\begin{align}
&\ \sqrt{n}\|(\bs{e}_i^{(n)}\otimes I_k)Y\bs{e}_j^{(k)}\| \\
\leq & \sum\limits_{l=1}^{\infty}\|(\overline{L}_{-\lambda_j}^{+})E\|_{1,\infty}^l \|X\| \nonumber \\
\leq &\ \sum\limits_{l=1}^{\infty}\big(\|\overline{L}_{-\lambda_j}^{+}\|_{1,\infty}\|E\|_{1,\infty}\big)^l \|X\|\nonumber  \\
= &\ \frac{\|\overline{L}_{-\lambda_j}^{+}\|_{1,\infty}\|E\|_{1,\infty}}{1-\|\overline{L}_{-\lambda_j}^{+}\|_{1,\infty}\|E\|_{1,\infty}} \|X\|\nonumber  \\
\underset{\leq}{(\ref{Eq:recovery})} &\ \frac{\|\overline{L}^{+}\|_{1,\infty}\|E\|_{1,\infty}(1+\|\overline{L}^{+}\|_{1,\infty}\|E\|_{1,\infty})}{1-\|\overline{L}^{+}\|_{1,\infty}\|E\|_{1,\infty}(1+\|\overline{L}^{+}\|_{1,\infty}\|E\|_{1,\infty})} \cdot \|X\| \nonumber \\
\leq  &\ \frac{\epsilon_2(1+\epsilon_2)}{1-\epsilon_2(1+\epsilon_2)}\cdot \Big(1+\big(\frac{\epsilon_1}{2-\epsilon_1}\big)^2\Big).
\label{Eq:Prop:1:Bound:2}
\end{align}
We can now conclude the proof by combining (\ref{Eq:Prop:1:Bound:1}), (\ref{Eq:Prop:1:Bound:2}), Lemma~\ref{Lemma:Lose:Bound}, and Lemma~\ref{Lemma:Projection:Stability}.
\qed

\subsection{Robust Recovery of Translations}
\label{Section:Prop:4:2}

In the same spirit as the preceding section, we assume the underlying ground-truth satisfies
\begin{equation}
R_i^{\star} = I_k, \bs{t}_i^{\star} = \bs{0}, \quad 1\leq i \leq n.
\label{Eq:gt:trans}
\end{equation}
In other words, a correct measurement along edge $(i,j)\in \set{E}$ should satisfy $R_{ij} = I_k, \bs{t}_{ij} = \bs{0}$. As we will see later, this assumption makes the error bound easier to parse. It is easy to see that the more general setting can always be converted into this simple setup through factoring out the rigid transformations among the coordinate systems associated with the input objects. 

We present a formal statement of Prop. 4.2 of the main paper as follows:
\begin{proposition}
Consider the assumption of (\ref{Eq:gt:trans}). Define
\begin{equation}
\epsilon_3:= \max\limits_{1\leq i \leq n} \sum\limits_{j \in \set{N}(i)}w_{ij}\|t_{ij}\|
\end{equation}
Intuitively, $\epsilon_3$ measures the cumulative transformation error associated with each object. Under the same assumption as Prop.~\ref{Prop:4:1:Formal} for the connection Laplacian $L = \overline{L}\otimes I_k +E$, we can bound the error of the translation synchronization step as 
\begin{equation}
\max\limits_{1\leq i \leq n}\|\bs{t}_i\|\leq \frac{\|\overline{L}^{+}\|_{1,\infty}\epsilon_3}{1-4\epsilon_2}.
\label{eq:translation:error}
\end{equation}
\label{Prop:4:2:Formal}
\end{proposition}

\noindent\textbf{Proof of Lemma~\ref{Prop:4:2:Formal}:}
First of all, note that $(\bs{1}\otimes I_k)^{T}\bs{b} = 0$. Thus we can factor out the component in $E$ that corresponds the the subspace spanned by $\bs{1}\otimes I_k$. Specifically, define
$$
E' = \big((I-\frac{1}{n}\bs{1}\bs{1}^{T})\otimes I_k\big)E\big((I-\frac{1}{n}\bs{1}\bs{1}^{T})\otimes I_k\big).
$$
It is easy to check that 
$$
\|E'\|_{1,\infty}\leq 4\|E\|_{1,\infty}.
$$
Moreover, 
$$
\bs{t} = (\overline{L}\otimes I_k + E')^{+}\bs{b}.
$$
This means
$$
\max\limits_{1\leq i \leq n}\|\bs{t}_i\|\leq \|(\overline{L}\otimes I_k + E')^{+}\|_{1,\infty}\epsilon_3.
$$
Note that
$$
(\overline{L}\otimes I_k + E')^{+} = \sum\limits_{l=0}^{\infty}\big((\overline{L}^{+}\otimes I_k)E'\big)^l(\overline{L}^{+}\otimes I_k).
$$
It follows that
\begin{align*}
&\ \|(\overline{L}\otimes I_k + E')^{+}\|_{1,\infty} \\
\leq&\  \sum\limits_{l=0}^{\infty} \|\big((\overline{L}^{+}\otimes I_k)E'\big)\|_{1,\infty}^l\|(\overline{L}^{+}\otimes I_k)\|_{1, \infty} \\
=&\ \sum\limits_{l=0}^{\infty} (\|\overline{L}^{+}\|_{1,\infty}\|E'\|_{1,\infty})^l\|\overline{L}^{+}\|_{1,\infty} \\
\leq & \ \frac{\|\overline{L}^{+}\|_{1,\infty}}{1-4\|\overline{L}^{+}\|_{1,\infty}\|E\|_{1,\infty}}.
\end{align*}
\qed

\subsection{Proof of Proposition 1 in the Main Paper}
\label{Section:Prop:4:3}

Applying Lemma~\ref{prop:derivative:single_rotation}, we have
\begin{align}
\dd R_i=\sum_{1\leq s,t\leq k}\frac{{\bs{v}^{(i)}_s}^T\dd U_i\bs{w}_t^{(i)}-{\bs{v}_t^{(i)}}^T\dd U_i\bs{w}_s^{(i)}}{\sigma_s^{(i)}+\sigma_t^{(i)}}\bs{v}_s^{(i)}{\bs{w}_t^{(i)}}^T.
\end{align}

We further divide the computation of $\dd U_i$ into two parts. Consider the $j$-th column of $\dd U_i$:
\begin{align}
\dd U_i\bs{e}_j^{(k)}&=({\bs{e}_i^{(n)}}^T\otimes I_k)\dd U\bs{e}_j^{(k)}\nonumber\\
&=({\bs{e}_i^{(n)}}^T\otimes I_k)\dd \bs{u}_j\nonumber\\
&=({\bs{e}_i^{(n)}}^T\otimes I_k)\sum_{l\neq j}\bs{u}_l\frac{\bs{u}_j^T\dd L\bs{u}_l}{\lambda_j-\lambda_l}\nonumber\\
&=({\bs{e}_i^{(n)}}^T\otimes I_k)\Big(\sum_{\substack{l=1\\l\neq j}}^k\frac{\bs{u}_j^T\dd L\bs{u}_l}{\lambda_j-\lambda_l}\bs{u}_l+\nonumber\\
&\qquad\sum_{l=k+1}^{kn}\frac{\bs{u}_j^T\dd L\bs{u}_l}{\lambda_j-\lambda_l}\bs{u}_l\Big)\nonumber\\
 &=\dd U_i^{(inner)}\bs{e}_j^{(k)}+\dd U_i^{(outer)}\bs{e}_j^{(k)}
\end{align}
where
\begin{align}
\dd U_i^{(inner)}\bs{e}_j^{(k)}&=({\bs{e}_i^{(n)}}^T\otimes I_k)\sum_{\substack{l=1\\l\neq j}}^k\frac{\bs{u}_l\bs{u}_l^{T}}{\lambda_j-\lambda_l}\dd L\bs{u}_j
\label{eq:derivative:20}
\\
&=\sum_{\substack{l=1\\l\neq j}}^k (U_i\bs{e}_l^{(k)})\frac{\bs{u}_l\bs{u}_l^T}{\lambda_j-\lambda_l}\dd L\bs{u}_j\\
\dd U_i^{(outer)}\bs{e}_j^{(k)} &= ({\bs{e}_i^{(n)}}^T\otimes I_k)\sum_{l=k+1}^{kn}\frac{\bs{u}_l\bs{u}_l^T}{\lambda_j-\lambda_l}\dd L\bs{u}_j
\label{eq:derivative:17}
\end{align}
In (\ref{eq:derivative:20}), we used the fact that $({\bs{e}_i^{(n)}}^T\otimes I_k)\bs{u}_l$ is just $U_i\bs{e}_l^{(k)}$ by definition of $U_i$.

Since $\dd R_i$ is linear with respect to $\dd U_i$, we can divide $\dd R_i$ similarly:
\begin{align}
&\dd R_i = \dd R_{i}^{(inner)}+\dd R_{i}^{(outer)}\nonumber\\
&\dd R_i^{(inner)}:=\nonumber\\
&\sum_{1\leq s,t\leq k}\frac{{\bs{v}^{(i)}_s}^T\dd U_i^{(inner)}\bs{w}_t^{(i)}-{\bs{v}_t^{(i)}}^T\dd U_i^{(inner)}\bs{w}_s^{(i)}}{\sigma_s^{(i)}+\sigma_t^{(i)}}\bs{v}_s^{(i)}{\bs{w}_t^{(i)}}^T
\label{eq:derivative:15}
\\
&\dd R_i^{(outer)}:=\nonumber\\
&\sum_{1\leq s,t\leq k}\frac{{\bs{v}^{(i)}_s}^T\dd U_i^{(outer)}\bs{w}_t^{(i)}-{\bs{v}_t^{(i)}}^T\dd U_i^{(outer)}\bs{w}_s^{(i)}}{\sigma_s^{(i)}+\sigma_t^{(i)}}\bs{v}_s^{(i)}{\bs{w}_t^{(i)}}^T
\label{eq:derivative:16}
\end{align}

Then the derivative we would like to compute can be written as
\begin{align}
\dd (R_i R_j^T)&=\dd R_iR_j^T+R_i{\dd R_j}^T\nonumber\\
&=\dd R_i^{(inner)} R_j^T+R_i{\dd R_j^{(inner)}}^T\nonumber\\
&\quad +\dd R_i^{(outer)}R_j^T+R_i\dd {R_j^{(outer)}}^T.
\label{eq:derivative:18}
\end{align}
From (\ref{eq:derivative:16}) and (\ref{eq:derivative:17}) it can be easily checked that the formula in Proposition 1 of the main aper that we want to prove is just (\ref{eq:derivative:18}) except the extra terms $\dd R_i^{(inner)}R_j^T+R_i\dd {R_j^{(inner)}}^T$. Hence in the remaining proof it suffices to show that
$$
\dd R_i^{(inner)}R_j^T+R_i\dd {R_j^{(inner)}}^T=0.
$$

To this end, we define a $k$-by-$k$ auxiliary matrix $C$ as
$$
C_{lj}=\frac{\bs{u}_j^T\dd L\bs{u}_l}{\lambda_j-\lambda_l}
$$
for all $l\neq j$ and $C_{jj}=0$. Since $L$ is symmetric, $C$ would be skew-symmetric that means $C+C^T=0$. Fist of all, notice that
$$
\dd U_i^{(inner)}=\sum_{\substack{l=1\\l\neq k}}^k (U_i\bs{e}_l^{(k)})\frac{\bs{u}_j^T\dd L\bs{u}_l}{\lambda_j-\lambda_l}=U_i C.
$$
Also it is clear that
$$
{\bs{v}_s^{(i)}}^T U_i=\sigma_s\bs{w}_s^T
$$
by using simple properties of SVD. It follows that 
\begin{align}
\dd R_i^{(inner)}&=\sum_{1\leq s,t\leq k}\frac{\sigma_s{\bs{w}_s^{(i)}}^TC\bs{w}_t^{(i)}-\sigma_t{\bs{w}_t^{(i)}}^TC\bs{w}_s^{(i)}}{\sigma_s^{(i)}+\sigma_t^{(i)}}\bs{v}_s^{(i)}{\bs{w}_t^{(i)}}^T\label{eq:19:1}\\
&=\sum_{1\leq s,t\leq k}{\bs{w}_s^{(i)}}^TC\bs{w}_t^{(i)}\bs{v}_s^{(i)}{\bs{w}_t^{(i)}}^T\label{eq:19:2}\\
&=\sum_{1\leq s,t\leq k}{\bs{v}_s^{(i)}}{\bs{w}_s^{(i)}}^TC\bs{w}_t^{(i)}{\bs{w}_t^{(i)}}^T\label{eq:19:3}\\
&={V^{(i)}}{W^{(i)}}^TC\label{eq:19:4}\\
&=R_iC.
\label{eq:derivative:19}
\end{align}
In the derivations above, we used the fact that $C$ is a skew-symmetric matrix for deriving the first equality (\ref{eq:19:1}). In addition, we used the fact that ${\bs{v}_s^{(i)}}^TC\bs{v}_t^{(i)}$ is a scalar for deriving the second equality (\ref{eq:19:2}). When deriving (\ref{eq:19:3}), we used the expansion of $U^{(i)}{V^{(i)}}^T$ and the fact that $\{\bs{v}_1^{(i)},\dots,\bs{v}_k^{(i)}\}$ form an orthonormal basis:
$$
U^{(i)}{V^{(i)}}^T=\sum_{s=1}^k\bs{u}_s^{(i)}{\bs{v}_s^{(i)}}^T,\ I=\sum_{t=1}^k \bs{v}_s^{(i)}{\bs{v}_s^{(i)}}^T.
$$
(\ref{eq:19:4}) uses the definition of $R_i$. Finally, plugging (\ref{eq:derivative:19}) into (\ref{eq:derivative:18}) gives
$$
\dd R_i^{(inner)} R_j^T+R_i{\dd R_j^{(inner)}}^T=R_iC R_j^T+R_iC^TR_j^T=0,
$$
which completes our proof.
\qed

\subsection{Proof of Proposition 2 in the Main Paper}
\label{Section:Prop:4:4}

The proof is straightforward, since 
$$
0 = d(L\cdot L^{-1}) = dL\cdot L^{-1} + L\cdot d(L^{-1}),
$$
meaning 
$$
dL = -L^{-1} dL L^{-1}.
$$
In the degenerate case, we replace $L^{-1}$ by $L^{+}$. This is proper since the only null space of $L$ is $\bs{1}\otimes I_k$, which does not affect the solution $\bs{t}$. 
\qed

\subsection{Exact Recovery Condition of Rotation Synchronization}
\label{Section:Exact:Recovery:Rot:Sync}

Similar to~\cite{DBLP:conf/nips/HuangLBH17}, we can derive a truncated rotation synchronization scheme (the generalization to transformation synchronization is straight-forward). Specifically, consider an observation graph $\set{G} = (\set{V}, \set{E})$. Let $\set{E}_{bad}\subset \set{E}$ be the edge set associated with incorrect rotation measurements. Starting from $\set{G}$, at each iteration, we use the solution $R_i^{(k)},1\leq i \leq n$ at the kth iteration to prune input rotations whenever $\|R_{ij} - {R_j^{(k)}}{R_i^{(k)}}^T\\2\gamma^k$, where $\gamma <1$ is a constant. Using Prop.~\ref{Prop:4:1:Formal}, we can easily derive the following exact recovery condition:
\begin{proposition}
The truncated rotation synchronization scheme recovers the underlying ground-truth if
\begin{equation}
\|L_{\set{G}}^+\|_{1,\infty}d_{\max}(\set{E}_{bad})\leq \frac{1}{16}, \quad \gamma > 0.95,
\label{Eq:Rot:Sync:Exact:Recovery}
\end{equation}
where $L_{\set{G}}$ is the graph Laplacian of $\set{G}$, and $d_{\max}(\set{E}_{bad})$ is the maximum number of bad edges per vertex. Note that the constants in (\ref{Eq:Rot:Sync:Exact:Recovery}) are not optimized. 
\end{proposition}

\noindent\textsl{Proof:} Denote $\epsilon_4:=\|L_{\set{G}}^+\|_{1,\infty}d_{\max}(\set{E}_{bad})$. Consider an arbitrary set $\set{E}_r\subseteq \set{E}_{bad}$. Introduce the graph that collects the corresponding remaining observations $\set{G}_{cur} = (\set{V}, E_{cur})$, where $E_{cur} =  \set{E}\setminus (\set{E}_{bad}\setminus \set{E}_r)$. Suppose we apply rotation synchronization step to $\set{G}_{cur}$ and the associated observations, it is easy to show that (c.f.\cite{DBLP:conf/nips/HuangLBH17})
$$
\epsilon_2 \leq 2\frac{\epsilon_4}{1-\epsilon_4}\cdot \max_{(i,j)\in \set{E}_{cur}}\|N_{ij}\|, \quad \epsilon_1 \leq 2 \epsilon_2.
$$
Using Prop.~\ref{Prop:4:1:Formal} and after simple calculations, we can derive that the truncated scheme described above will never remove good measurements, which end the proof. 
\qed

\begin{remark}
This exact recovery condition suggests that if we simply let the weighting function to be small when the residual is big, then if the ratio of the incorrect measurements is small. It is guaranteed to remove all the incorrect measurement. Yet to maximize the effectiveness of the weighting scheme, it is suggested to learn the optimal weighting scheme from data. The approach presented in the main paper is one attempt in this direction.  
\end{remark}

\subsection{Proofs of Key Lemmas}
\label{Section:Proof:Lemma}

\subsubsection{Proof of Lemma~\ref{Lemma:Y:Expression}}

We first introduce the following notations, which essentially express $E$ in the coordinate system spanned by $\frac{1}{\sqrt{n}}(\bs{1}\otimes I_k)$ and $\overline{U}\otimes I_k$:
\begin{align*}
E_{11}: & = \frac{1}{n}(\bs{1}\otimes I_k)^{T}E(\bs{1}\otimes I_k), \\
E_{12}: & = \frac{1}{\sqrt{n}}(\bs{1}\otimes I_k)^{T}E(\overline{U}\otimes I_k),\\
E_{21}: & = \frac{1}{\sqrt{n}}(\overline{U}\otimes I_k)^{T}E(\bs{1}\otimes I_k), \\
E_{22}: & = (\overline{U}\otimes I_k)^{T}E(\overline{U}\otimes I_k).
\end{align*}
Let $Y: = (U\otimes I_k)\overline{Y}$. Substituting $U = \frac{1}{\sqrt{n}}(\bs{1}\otimes I_k) + (\overline{U}\otimes I_k)\overline{Y}$ into 
$$
(\overline{L}\otimes I_k+E)U = U\Lambda,
$$
we obtain
\begin{align*}
&(\overline{L}\otimes I_k+E)(\frac{1}{\sqrt{n}}(\bs{1}\otimes I_k) + (\overline{U}\otimes I_k)\overline{Y})\\
= & (\frac{1}{\sqrt{n}}(\bs{1}\otimes I_k) + (\overline{U}\otimes I_k)\overline{Y})\Lambda.
\end{align*}
Multiply both sides by $(U\otimes I_k)^{T}$, it follows that
$$
E_{21}X + (\overline{\Lambda}\otimes I_k)\overline{Y} + E_{22}\overline{Y} = \overline{Y}\Lambda.
$$
Since $\|E\| < \frac{\overline{\lambda}_2}{2}$, we have
\begin{align*}
Y\bs{e}_{j}^{(k)} &:= (\overline{U}\otimes I_k)\overline{Y}\bs{e}_j^{(k)} \\
& = -(\overline{U}\otimes I_k)\big((\overline{\Lambda}-\lambda_j)\otimes I_k-E_{22}\big)^{-1}E_{21}X\bs{e}_j^{(k)} \\
& = -\sum\limits_{l=0}^{\infty}(\overline{U}\otimes I_k)\big(((\overline{\Lambda}-\lambda_j)^{-1}\otimes I_k)E_{22}\big)^l\\
&\qquad \cdot \big((\overline{\Lambda}-\lambda_j)^{-1}\otimes I_k\big)E_{21}X\bs{e}_j^{(k)} \\
& = -\sum\limits_{l=1}^{\infty}\big((\overline{U}(\overline{\Lambda}-\lambda_j)^{-1}\overline{U}^T)\otimes I_k E\big)^l\frac{1}{\sqrt{n}}\bs{1}\otimes X\bs{e}_j^{(k)} \\
& = -\frac{1}{\sqrt{n}}\sum\limits_{l=1}^{\infty} \big(L_{-\lambda_j}^{+}E\big)^l(\bs{1}\otimes X\bs{e}_{j}^{(k)}). 
\end{align*}
\qed
\subsubsection{Proof of Lemma~\ref{Lemma:Laplacian:Derivative}} 
Let
\begin{align*}
    L=\sum_{i=1}^N\sigma_i\bs{u}_i\bs{u}_i^T=\sum_{i=1}^N\sigma_i\bs{u}'_i\bs{u}'^{T}_i
\end{align*}
be two different decompositions of $L$. It can be written in matrix form
$$
L=U\Lambda U^T=U'\Lambda U'^T
$$
where $U=[\bs{u}_1,\dots,\bs{u}_N]$, $\Lambda=\diag(\sigma_1,\dots,\sigma_N)$, $U'=[\bs{u}'_1,\dots,\bs{u}_N]$. Then we have
$$
(U'^TU)\Lambda=\Lambda (U'^TU)
$$

Let $A=U'^TU$ and the element of position $(i,j)$ on $A$ be $a_{ij}$, then we have
$$
a_{ij}\sigma_j=\sigma_i a_{ij},
$$
which means $a_{ij}=0$ for all $\sigma_i\neq \sigma_j$.

Since we have assumed $\sigma_1\leq \dots\leq \sigma_k<\sigma_{k+1}\leq\dots\leq \sigma_N$, the matrix $A$ would have form
$$
\begin{bmatrix}
A_{k,k} & O_{k,N-k}\\
O_{N-k,k} & A_{N-k,N-k}
\end{bmatrix}.
$$

But we have known that $A$ is an orthogonal matrix, thus $A_{k,k}$ is also an orthogonal matrix. In this way $A_{k,k}=U'^T_k U_k$ can be rewritten as
$$
U_k=U'_k A_{k,k}
$$
and furthermore we have
$$
U_k U_k^T=U_k'(A_{k,k} A_{k,k}^T) U'^T_k=U_k' U'^T_k.
$$

Since eigen-decomposition is a special case of SVD when dealing with symmetric matrix, (\ref{eq:derivative:9}) gives
\begin{align*}
\dd \bs{u}_i&=\sum_{j\neq i}\frac{\sigma_i \bs{u}_j^T\dd L\bs{u}_i+\sigma_j\bs{u}_i^T\dd L\bs{u}_j}{\sigma_i^2-\sigma_j^2}\bs{u}_j\\
&=\sum_{j\neq i}\frac{\bs{u}_i^T\dd M\bs{u}_j}{\sigma_i-\sigma_j}\bs{u}_j
\end{align*}
in which we used the fact that $\dd L$ is also symmetric in the last step.

Finally the differential of $U_kU_k^T$ can be written as
\begin{align*}
    &\dd(U_k U_k^T)\\
    =&\dd\left(\sum_{i=1}^k\bs{u}_i\bs{u}_i^T\right)\\
    =&\sum_{i=1}^k(\dd \bs{u}_i\bs{u}_i^T+\bs{u}_i\dd\bs{u}_i^T)\\
    =&\sum_{i=1}^k\sum_{\substack{j=1\\j\neq i}}^N\frac{\bs{u}_i^T\dd M\bs{u}_j}{\sigma_i-\sigma_j}(\bs{u}_j\bs{u}_i^T+\bs{u}_i \bs{u}_j^T)\\
    =&\sum_{i=1}^k\sum_{j=k+1}^N\frac{\bs{u}_i^T\dd M\bs{u}_j}{\sigma_i-\sigma_j}(\bs{u}_j\bs{u}_i^T+\bs{u}_i \bs{u}_j^T)+\\
    &\sum_{i=1}^k\sum_{j=i+1}^k\left(\frac{\bs{u}_i^T\dd M\bs{u}_j}{\sigma_i-\sigma_j}+\frac{\bs{u}_i^T\dd M\bs{u}_j}{\sigma_j-\sigma_i}\right)(\bs{u}_j\bs{u}_i^T+\bs{u}_i \bs{u}_j^T)\\
    =&\sum_{i=1}^k\sum_{j=k+1}^N\frac{\bs{u}_i^T\dd M\bs{u}_j}{\sigma_i-\sigma_j}(\bs{u}_j\bs{u}_i^T+\bs{u}_i \bs{u}_j^T)
\end{align*}

As for formula (\ref{eq:derivative:eigenvalue}), taking differential of equation $L\bs{u}_i=\sigma_i\bs{u}_i$, we obtain
$$
\dd L\bs{u}_i + L\dd\bs{u}_i=\dd\sigma_i\bs{u}_i+\sigma_i\dd\bs{u}_i
$$

Let us multiply both sides by $\bs{u}_i^T$ and notice that $\bs{u}_i^T\bs{u}_i=1$, $L\bs{u}_i=\sigma_i\bs{u}_i$, and $\bs{u}_i^T\dd\bs{u}_i=0$, we conclude that the equation above can be simplified to
$$
\dd\sigma_i=\bs{u}_i^T\dd L\bs{u}_i.
$$
\qed

\subsubsection{Proof of Lemma~\ref{Lemma:Block:Norm}} 

It is well-known that $\|X\|\leq \|X\|_F$ for any matrix $X$ where $\|\cdot\|_F$ represents the Frobenius norm. Thus
\begin{align*}
k\|A\|^2&= \sum_{j=1}^k\|A\|^2\geq\sum_{j=1}^k\sum_{i=1}^{kn}a_{ij}^2\\
    &=\sum_{i=1}^{kn}\sum_{j=1}^ka_{ij}^2=\sum_{i=1}^n \|A_{i}\|_F^2\\
    &\geq \sum_{i=1}^n \|A_i\|
\end{align*}
completes our proof.
\qed

\subsubsection{Proof of Lemma~\ref{Lemma:Projection:Stability}} 

Suppose $M=\sum_{i=1}^n\sigma_i\bs{u}_i\bs{v}_i^T$ is the SVD decomposition of $M$.  By definition of $R(\cdot)$,
$$R(M)=\sum_{i=1}^n\bs{u}_i\bs{v}_i^T.$$
First we have a simple lower bound on $\epsilon$:
\begin{equation}
\epsilon=\|M-I\|\geq |(M-I)\bs{v_i}|=|\sigma_i-1|.
\label{eq:derivative:12}
\end{equation}
It is enough to show that for any unit vector $\bs{p}\in\mathbb{R}^{n}$ we have
\begin{equation}
\|R(M)\bs{p}-\bs{p}\|\leq (1+\epsilon)\|M\bs{p}-\bs{p}\|.
\label{eq:derivative:11}
\end{equation}
In fact, if (\ref{eq:derivative:11}) is true, then
\begin{align*}
\|R(M)-I\|&=\max_{|\bs{p}|=1}\|R(M)\bs{p}-\bs{p}\|\\
&\leq \max_{|\bs{p}=1}(1+\epsilon)\|M\bs{p}-\bs{p}\|\\
&\leq (1+\epsilon)\epsilon\textup{\quad (by definition of $\epsilon$)}.
\end{align*}

By noting $\{\bs{v}_1,\dots,\bs{v}_n\}$ are a set of basis on $\mathbb{R}^n$, we can decompose $R(M)$ and $M$ into
\begin{align*}
R(M)\bs{p}-\bs{p}&=\sum_{i=1}^n\bs{u}_i\bs{v}_i^T\bs{p}-\sum_{i=1}^n\bs{v}_i(\bs{v}_i^T\bs{p})\\
&=\sum_{i=1}^n(\bs{u}_i-\bs{v}_i)(\bs{v}_i^T\bs{p})\\
M\bs{p}-\bs{p}&=\sum_{i=1}^n\sigma_i\bs{u}_i\bs{v}_i\bs{p}-\sum_{i=1}^n\bs{v}_i(\bs{v}_i^T\bs{p})\\
&=\sum_{i=1}^n(\sigma_i\bs{u}_i-\bs{v}_i)(\bs{v}_i^T\bs{p})
\end{align*}
To prove (\ref{eq:derivative:11}), it suffices to show that
$$
|\bs{u}_i-\bs{v}_i|\leq (1+\epsilon)|\sigma_i\bs{u}_i-\bs{v}_i|.
$$
Let $\delta=\bs{u}_i^T\bs{v}_i$. The case that $\bs{u}_i=\bs{v}_i$ is trivial. Also, if $\sigma_i=0$, then $\epsilon\geq 1$ and the resulting inequality
$$
|\bs{u}_i-\bs{v}_i|\leq 2|\bs{v}_i|
$$
is trivial. Thus we can always assume $|\sigma \bs{u}_i-\bs{v}_i|\neq 0$ and $\sigma_i>0$. Then
by the laws of cosines we have
\begin{align}
\frac{|\bs{u}_i-\bs{v}_i|}{|\sigma_i\bs{u}_i-\bs{v}_i|}&=\sqrt{\frac{2-2\delta}{1+\sigma_i^2-2\sigma_i\delta}}\nonumber\\
&=\sqrt{\sigma_i^{-1}+\frac{2-(\sigma_i^{-1}+\sigma_i)}{1+\sigma_i^2-2\sigma_i\delta}}
\label{eq:derivative:13}
\end{align}
In (\ref{eq:derivative:13}) it is clear that $2-(\sigma_i^{-1}+\sigma_i)\leq 0$ and $1+\sigma_i^2\geq 2\sigma_i\delta$. Hence by monotonicity (\ref{eq:derivative:13}) reaches its maximum when $\delta=-1$ and then
$$
\frac{|\bs{u}_i-\bs{v}_i|}{|\sigma_i\bs{u}_i-\bs{v}_i|}\leq \frac{2}{1+\sigma_i}=1+\frac{1-\sigma_i}{1+\sigma_i}\leq 1+|\sigma_i-1|\leq 1+\epsilon
$$

\subsubsection{Proof of Lemma~\ref{prop:derivative:single_rotation}} 

For the sake of brevity we simply write $R$ instead of $R(M)$ in the following proof. It is easy to see that
$$
M\bs{v}_i=\sigma_i\bs{u}_i\quad \bs{u}_i^TM=\sigma_i\bs{v}_i^T
$$
for $i=1,\dots,n$. Taking the differential on both sides we obtain
\begin{equation}
\dd M\bs{v}_i+M\dd\bs{v}_i=\dd\sigma_i\bs{u}_i+\sigma_i\dd\bs{u}_i
\label{eq:derivative:1}
\end{equation}
\begin{equation}
\dd\bs{u}_i^TM+\bs{u}_i^T\dd M=\dd\sigma_i\bs{v}_i^T+\sigma_i\dd\bs{v}_i^T
\label{eq:derivative:2}
\end{equation}

Left multiplying both sides of (\ref{eq:derivative:1}) by $\bs{u}_j$ with $j\neq i$ and observing that $\bs{u}_j^T\bs{u}_i=0$, we obtain

\begin{equation}
\bs{u}_j^T\dd M\bs{v}_i+\bs{u}_j^TM\dd\bs{v}_i=\sigma_i\bs{u}_j^T\dd \bs{u}_i
\label{eq:derivative:3}
\end{equation}
Similarly right multiplying both sides of (\ref{eq:derivative:2}) by $\bs{v}_j^T$ with $j\neq i$ gives

\begin{equation}
\dd\bs{u}_i^TM\bs{v}_j+\bs{u}_i^T\dd M\bs{v}_j=\sigma_i\dd \bs{v}_{i}^T \bs{v}_j
\label{eq:derivative:4}
\end{equation}

Since $\bs{u}_j^TM=\sigma_j v_j^T,\ Mv_j=\sigma_j\bs{u}_j^T$, we have
\begin{equation}
\bs{u}_j^T\dd M\bs{v}_i+\sigma_j \bs{v}_j^T\dd\bs{v}_i=\sigma_i\bs{u}_j^T\dd \bs{u}_i
\label{eq:derivative:5}
\end{equation}
\begin{equation}
\dd\bs{u}_i^T\sigma_j\bs{u}_j+\bs{u}_i^T\dd M\bs{v}_j=\sigma_i\dd \bs{v}_{i}^T \bs{v}_j
\label{eq:derivative:6}
\end{equation}
for all $i\neq j$.

Observe that $\bs{u}_j^T\dd \bs{u}_i=\dd\bs{u}_i^T\bs{u}_j,\ \bs{v}_j^T\dd \bs{v}_i=\dd \bs{v}_i^T\bs{v}_j$. Combining (\ref{eq:derivative:5}) and (\ref{eq:derivative:6}) and regarding them as a linear equation group about $\bs{u}_j^T\dd\bs{u}_i$ and $\bs{v}_j^T\dd\bs{v}_i$ they can be solved out as

\begin{equation}
\bs{u}_j^T\dd\bs{u}_i=\frac{\sigma_i\bs{u}_j^T\dd M\bs{v}_i+\sigma_j\bs{u}_i^T\dd M\bs{v}_j}{\sigma_i^2-\sigma_j^2}
\label{eq:derivative:7}
\end{equation}
\begin{equation}
\bs{v}_j^T\dd \bs{v}_i=\frac{\sigma_i\bs{u}_i^T\dd M\bs{v}_j+\sigma_j\bs{u}_j^T\dd M\bs{v}_i}{\sigma_i^2-\sigma_j^2}
\label{eq:derivative:8}
\end{equation}

Since $\bs{u}_i^T\bs{u}_i=\|\bs{u}_i\|=1$, we have $\bs{u}_i^T\dd \bs{u}_i=0$. As $\{\bs{u}_1,\dots,\bs{u}_n\}$ form a set of orthogonal basis of $\mathbb{R}^n$, we can write $\bs{u}_i$ as
\begin{equation}
\dd\bs{u}_i=\sum_{j\neq i}\frac{\sigma_i\bs{u}_j^T\dd M\bs{v}_i+\sigma_j\bs{u}_i^T\dd M\bs{v}_j}{\sigma_i^2-\sigma_j^2}\bs{u}_j
\label{eq:derivative:9}
\end{equation}
Similarly for $\dd \bs{v}_i$ we have
\begin{equation}
\dd\bs{v}_i=\sum_{j\neq i}\frac{\sigma_i\bs{u}_i^T\dd M\bs{v}_j+\sigma_j\bs{u}_j^T\dd M\bs{v}_i}{\sigma_i^2-\sigma_j^2}\bs{v}_j
\label{eq:derivative:10}
\end{equation}

Finally we can write $\dd R$ as
\begin{align*}
\dd R&=\sum \bs{u}_i\dd \bs{v}_i^T+\sum \dd\bs{u}_i\bs{v}_i^T\\
&=\sum_{i\neq j}\frac{\sigma_i\bs{u}_i^T\dd M\bs{v}_j+\sigma_j\bs{u}_j^T\dd M\bs{v}_i}{\sigma_i^2-\sigma_j^2}\bs{u}_i\bs{v}_j^T\\
&\quad+\sum_{i\neq j}\frac{\sigma_i\bs{u}_j^T\dd M\bs{v}_i+\sigma_j\bs{u}_i^T\dd M\bs{v}_j}{\sigma_i^2-\sigma_j^2}\bs{u}_j\bs{v}_i^T\\
&=\sum_{i\neq j}\frac{(\sigma_i-\sigma_j)\bs{u}_i^T\dd M\bs{v}_j-(\sigma_i-\sigma_j)\bs{u}_j^T\dd M\bs{v}_i}{\sigma_i^2-\sigma_j^2}\bs{u}_i\bs{v}_j^T\\
&=\sum_{i\neq j}\frac{\bs{u}_i^T\dd M\bs{v}_j-\bs{u}_j^T\dd M\bs{v}_i}{\sigma_i+\sigma_j}\bs{u}_i\bs{v}_j^T
\end{align*}
\qed 

\subsubsection{Proof of Lemma~\ref{Lemma:Axx:1}}

Since $\|E\|_{1,\infty}\leq \frac{\overline{\lambda}_2}{2}$, we have
\begin{align*}
\overline{L}_{-\lambda_j}^{+} & = \overline{U}(\overline{\Lambda}-\lambda_j)^{-1}\overline{U}^{T} \\
& = \overline{U}\sum\limits_{l=0}^{\infty}\overline{L}^{-(l+1)}\lambda_j^{l} \overline{U}^{T} \\
& = \sum\limits_{l=0}^{\infty} (\overline{L}^{+})^{l+1}\lambda_j^l, \quad 1\leq j \leq k.
\end{align*}
As $\|E\|_{1,\infty}\|\overline{L}^{+}\|_{1,\infty} < 1$, it follows that
\begin{align*}
\|\overline{L}_{-\lambda_j}^{+}\|_{1,\infty} & \leq \sum\limits_{l=0}^{\infty}\|\overline{L}^{+}\|_{1,\infty}^{l+1}\lambda_j^l \\
& \leq \sum\limits_{l=0}^{\infty}\|\overline{L}^{+}\|_{1,\infty}^{l+1}\|E\|_{1,\infty}^l \\
& = \|\overline{L}^{+}\|_{1,\infty}\big(1+\|\overline{L}^{+}\|_{1,\infty}\|E\|_{1,\infty}\big).
\end{align*}
\qed

\subsubsection{Proof of Lemma~\ref{Lemma:Lose:Bound}}

In fact, $\forall \bs{x}\in \R^k$, where $\|\bs{x}\| = 1$, we have
\begin{align*}
\|A\bs{x}\| & \leq \sum\limits_{j=1}^k \|A\bs{e}_j^{(k)}\| |x_j| \\
& \leq \max\limits_{1\leq j \leq k} \|A\bs{e}_j^{(k)}\| \sum\limits_{j=1}^k |x_j| \\
& \leq \max\limits_{1\leq j \leq k} \|A\bs{e}_j^{(k)}\| \sqrt{k}(\sum\limits_{j=1}^{k} x_j^2)^{\frac{1}{2}}.
\end{align*}
\qed

\section{Scenes used in this paper}
\label{Section:datasets}
For completeness, we show the scenes we used in this paper. Including 100 scenes from ScanNet \cite{dai2017scannet} dataset and 60 scenes from Redwood Chair dataset.
Fig.~\ref{fig:ScanNet:dataset}-Fig.~\ref{fig:ScanNet:dataset5} and Fig.~\ref{fig:Redwood:Chair:dataset}-Fig.~\ref{fig:Redwood:Chair:dataset3} show the scenes we used in the paper from ScanNet and Redwood chair dataset, respectively.

\begin{figure*}[!ht]
\begin{subfigure}{0.19\textwidth}
\includegraphics[width=\textwidth]{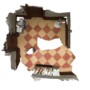}\\
\caption*{scene047300}
\end{subfigure}
\begin{subfigure}{0.19\textwidth}
\includegraphics[width=\textwidth]{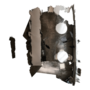}\\
\caption*{scene051300}
\end{subfigure}
\begin{subfigure}{0.19\textwidth}
\includegraphics[width=\textwidth]{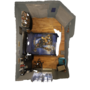}\\
\caption*{scene045700}
\end{subfigure}
\begin{subfigure}{0.19\textwidth}
\includegraphics[width=\textwidth]{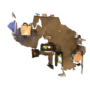}\\
\caption*{scene037400}
\end{subfigure}
\begin{subfigure}{0.19\textwidth}
\includegraphics[width=\textwidth]{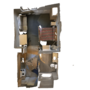}\\
\caption*{scene027601}
\end{subfigure}

\begin{subfigure}{0.19\textwidth}
\includegraphics[width=\textwidth]{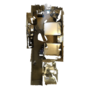}\\
\caption*{scene043500}
\end{subfigure}
\begin{subfigure}{0.19\textwidth}
\includegraphics[width=\textwidth]{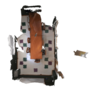}\\
\caption*{scene035802}
\end{subfigure}
\begin{subfigure}{0.19\textwidth}
\includegraphics[width=\textwidth]{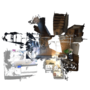}\\
\caption*{scene051600}
\end{subfigure}
\begin{subfigure}{0.19\textwidth}
\includegraphics[width=\textwidth]{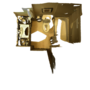}\\
\caption*{scene026001}
\end{subfigure}
\begin{subfigure}{0.19\textwidth}
\includegraphics[width=\textwidth]{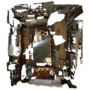}\\
\caption*{scene069601}
\end{subfigure}

\begin{subfigure}{0.19\textwidth}
\includegraphics[width=\textwidth]{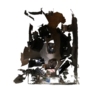}\\
\caption*{scene060800}
\end{subfigure}
\begin{subfigure}{0.19\textwidth}
\includegraphics[width=\textwidth]{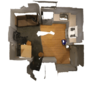}\\
\caption*{scene028801}
\end{subfigure}
\begin{subfigure}{0.19\textwidth}
\includegraphics[width=\textwidth]{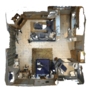}\\
\caption*{scene000001}
\end{subfigure}
\begin{subfigure}{0.19\textwidth}
\includegraphics[width=\textwidth]{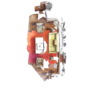}\\
\caption*{scene053600}
\end{subfigure}
\begin{subfigure}{0.19\textwidth}
\includegraphics[width=\textwidth]{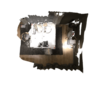}\\
\caption*{scene025601}
\end{subfigure}

\begin{subfigure}{0.19\textwidth}
\includegraphics[width=\textwidth]{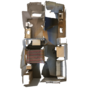}\\
\caption*{scene027600}
\end{subfigure}
\begin{subfigure}{0.19\textwidth}
\includegraphics[width=\textwidth]{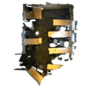}\\
\caption*{scene001500}
\end{subfigure}
\begin{subfigure}{0.19\textwidth}
\includegraphics[width=\textwidth]{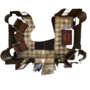}\\
\caption*{scene012900}
\end{subfigure}
\begin{subfigure}{0.19\textwidth}
\includegraphics[width=\textwidth]{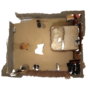}\\
\caption*{scene041800}
\end{subfigure}
\begin{subfigure}{0.19\textwidth}
\includegraphics[width=\textwidth]{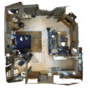}\\
\caption*{scene000002}
\end{subfigure}

\begin{subfigure}{0.19\textwidth}
\includegraphics[width=\textwidth]{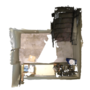}\\
\caption*{scene052400}
\end{subfigure}
\begin{subfigure}{0.19\textwidth}
\includegraphics[width=\textwidth]{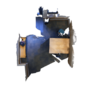}\\
\caption*{scene004301}
\end{subfigure}
\begin{subfigure}{0.19\textwidth}
\includegraphics[width=\textwidth]{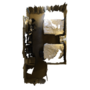}\\
\caption*{scene067700}
\end{subfigure}
\begin{subfigure}{0.19\textwidth}
\includegraphics[width=\textwidth]{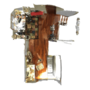}\\
\caption*{scene064600}
\end{subfigure}
\begin{subfigure}{0.19\textwidth}
\includegraphics[width=\textwidth]{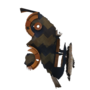}\\
\caption*{scene033400}
\end{subfigure}

\caption{ScanNet Train Dataset (1st to 25th)}
\label{fig:ScanNet:dataset}
\ContinuedFloat
\end{figure*}

\begin{figure*}[!ht]

\ContinuedFloat

\begin{subfigure}{0.19\textwidth}
\includegraphics[width=\textwidth]{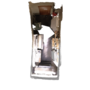}\\
\caption*{scene068501}
\end{subfigure}
\begin{subfigure}{0.19\textwidth}
\includegraphics[width=\textwidth]{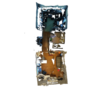}\\
\caption*{scene005400}
\end{subfigure}
\begin{subfigure}{0.19\textwidth}
\includegraphics[width=\textwidth]{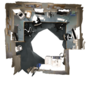}\\
\caption*{scene026401}
\end{subfigure}
\begin{subfigure}{0.19\textwidth}
\includegraphics[width=\textwidth]{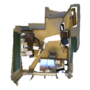}\\
\caption*{scene008900}
\end{subfigure}
\begin{subfigure}{0.19\textwidth}
\includegraphics[width=\textwidth]{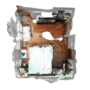}\\
\caption*{scene018400}
\end{subfigure}

\begin{subfigure}{0.19\textwidth}
\includegraphics[width=\textwidth]{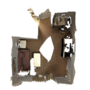}\\
\caption*{scene036203}
\end{subfigure}
\begin{subfigure}{0.19\textwidth}
\includegraphics[width=\textwidth]{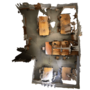}\\
\caption*{scene066700}
\end{subfigure}
\begin{subfigure}{0.19\textwidth}
\includegraphics[width=\textwidth]{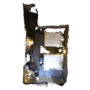}\\
\caption*{scene000602}
\end{subfigure}
\begin{subfigure}{0.19\textwidth}
\includegraphics[width=\textwidth]{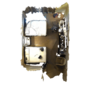}\\
\caption*{scene020901}
\end{subfigure}
\begin{subfigure}{0.19\textwidth}
\includegraphics[width=\textwidth]{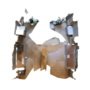}\\
\caption*{scene043100}
\end{subfigure}

\begin{subfigure}{0.19\textwidth}
\includegraphics[width=\textwidth]{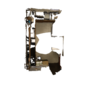}\\
\caption*{scene021001}
\end{subfigure}
\begin{subfigure}{0.19\textwidth}
\includegraphics[width=\textwidth]{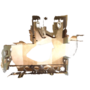}\\
\caption*{scene025400}
\end{subfigure}
\begin{subfigure}{0.19\textwidth}
\includegraphics[width=\textwidth]{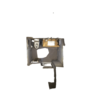}\\
\caption*{scene012400}
\end{subfigure}
\begin{subfigure}{0.19\textwidth}
\includegraphics[width=\textwidth]{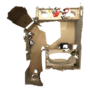}\\
\caption*{scene058102}
\end{subfigure}
\begin{subfigure}{0.19\textwidth}
\includegraphics[width=\textwidth]{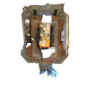}\\
\caption*{scene010200}
\end{subfigure}

\begin{subfigure}{0.19\textwidth}
\includegraphics[width=\textwidth]{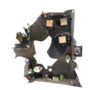}\\
\caption*{scene015201}
\end{subfigure}
\begin{subfigure}{0.19\textwidth}
\includegraphics[width=\textwidth]{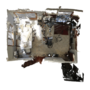}\\
\caption*{scene046501}
\end{subfigure}
\begin{subfigure}{0.19\textwidth}
\includegraphics[width=\textwidth]{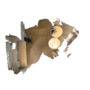}\\
\caption*{scene004800}
\end{subfigure}
\begin{subfigure}{0.19\textwidth}
\includegraphics[width=\textwidth]{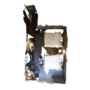}\\
\caption*{scene000600}
\end{subfigure}
\begin{subfigure}{0.19\textwidth}
\includegraphics[width=\textwidth]{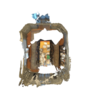}\\
\caption*{scene010201}
\end{subfigure}

\begin{subfigure}{0.19\textwidth}
\includegraphics[width=\textwidth]{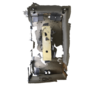}\\
\caption*{scene045201}
\end{subfigure}
\begin{subfigure}{0.19\textwidth}
\includegraphics[width=\textwidth]{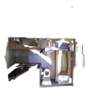}\\
\caption*{scene044702}
\end{subfigure}
\begin{subfigure}{0.19\textwidth}
\includegraphics[width=\textwidth]{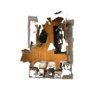}\\
\caption*{scene001601}
\end{subfigure}
\begin{subfigure}{0.19\textwidth}
\includegraphics[width=\textwidth]{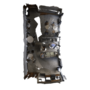}\\
\caption*{scene024701}
\end{subfigure}
\begin{subfigure}{0.19\textwidth}
\includegraphics[width=\textwidth]{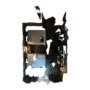}\\
\caption*{scene034000}
\end{subfigure}

\caption{ScanNet Train Dataset (26th to 50th)}
\ContinuedFloat
\label{fig:ScanNet:dataset2}
\end{figure*}

\begin{figure*}[!ht]

\ContinuedFloat
\begin{subfigure}{0.19\textwidth}
\includegraphics[width=\textwidth]{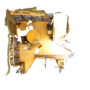}\\
\caption*{scene069201}
\end{subfigure}
\begin{subfigure}{0.19\textwidth}
\includegraphics[width=\textwidth]{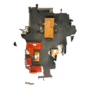}\\
\caption*{scene031701}
\end{subfigure}
\begin{subfigure}{0.19\textwidth}
\includegraphics[width=\textwidth]{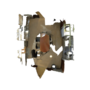}\\
\caption*{scene004700}
\end{subfigure}
\begin{subfigure}{0.19\textwidth}
\includegraphics[width=\textwidth]{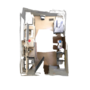}\\
\caption*{scene019702}
\end{subfigure}
\begin{subfigure}{0.19\textwidth}
\includegraphics[width=\textwidth]{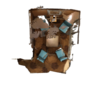}\\
\caption*{scene013401}
\end{subfigure}
\begin{subfigure}{0.19\textwidth}
\includegraphics[width=\textwidth]{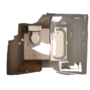}\\
\caption*{scene062500}
\end{subfigure}
\begin{subfigure}{0.19\textwidth}
\includegraphics[width=\textwidth]{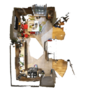}\\
\caption*{scene033501}
\end{subfigure}
\begin{subfigure}{0.19\textwidth}
\includegraphics[width=\textwidth]{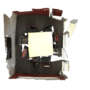}\\
\caption*{scene035400}
\end{subfigure}
\begin{subfigure}{0.19\textwidth}
\includegraphics[width=\textwidth]{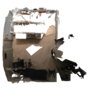}\\
\caption*{scene062900}
\end{subfigure}
\begin{subfigure}{0.19\textwidth}
\includegraphics[width=\textwidth]{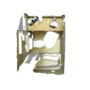}\\
\caption*{scene043402}
\end{subfigure}
\begin{subfigure}{0.19\textwidth}
\includegraphics[width=\textwidth]{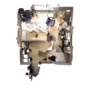}\\
\caption*{scene009200}
\end{subfigure}
\begin{subfigure}{0.19\textwidth}
\includegraphics[width=\textwidth]{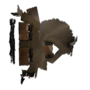}\\
\caption*{scene060901}
\end{subfigure}
\begin{subfigure}{0.19\textwidth}
\includegraphics[width=\textwidth]{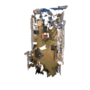}\\
\caption*{scene020600}
\end{subfigure}
\begin{subfigure}{0.19\textwidth}
\includegraphics[width=\textwidth]{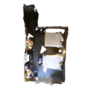}\\
\caption*{scene000601}
\end{subfigure}
\begin{subfigure}{0.19\textwidth}
\includegraphics[width=\textwidth]{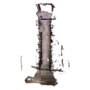}\\
\caption*{scene066900}
\end{subfigure}
\begin{subfigure}{0.19\textwidth}
\includegraphics[width=\textwidth]{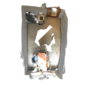}\\
\caption*{scene027401}
\end{subfigure}
\begin{subfigure}{0.19\textwidth}
\includegraphics[width=\textwidth]{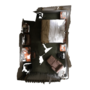}\\
\caption*{scene067901}
\end{subfigure}
\begin{subfigure}{0.19\textwidth}
\includegraphics[width=\textwidth]{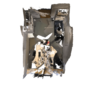}\\
\caption*{scene017702}
\end{subfigure}
\begin{subfigure}{0.19\textwidth}
\includegraphics[width=\textwidth]{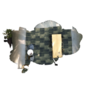}\\
\caption*{scene062200}
\end{subfigure}

\caption{ScanNet Train Dataset (51st to 69th)}
\ContinuedFloat
\label{fig:ScanNet:dataset3}
\end{figure*}

\begin{figure*}[!ht]

\ContinuedFloat

\begin{subfigure}{0.19\textwidth}
\includegraphics[width=\textwidth]{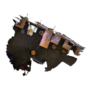}\\
\caption*{scene066100}
\end{subfigure}
\begin{subfigure}{0.19\textwidth}
\includegraphics[width=\textwidth]{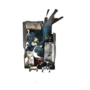}\\
\caption*{scene026201}
\end{subfigure}
\begin{subfigure}{0.19\textwidth}
\includegraphics[width=\textwidth]{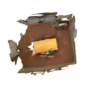}\\
\caption*{scene057801}
\end{subfigure}
\begin{subfigure}{0.19\textwidth}
\includegraphics[width=\textwidth]{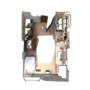}\\
\caption*{scene019701}
\end{subfigure}
\begin{subfigure}{0.19\textwidth}
\includegraphics[width=\textwidth]{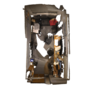}\\
\caption*{scene047401}
\end{subfigure}

\begin{subfigure}{0.19\textwidth}
\includegraphics[width=\textwidth]{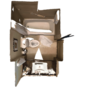}\\
\caption*{scene067601}
\end{subfigure}
\begin{subfigure}{0.19\textwidth}
\includegraphics[width=\textwidth]{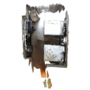}\\
\caption*{scene064202}
\end{subfigure}
\begin{subfigure}{0.19\textwidth}
\includegraphics[width=\textwidth]{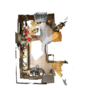}\\
\caption*{scene033502}
\end{subfigure}
\begin{subfigure}{0.19\textwidth}
\includegraphics[width=\textwidth]{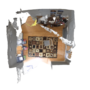}\\
\caption*{scene022901}
\end{subfigure}
\begin{subfigure}{0.19\textwidth}
\includegraphics[width=\textwidth]{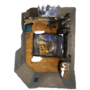}\\
\caption*{scene045701}
\end{subfigure}

\begin{subfigure}{0.19\textwidth}
\includegraphics[width=\textwidth]{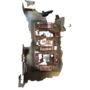}\\
\caption*{scene058802}
\end{subfigure}
\begin{subfigure}{0.19\textwidth}
\includegraphics[width=\textwidth]{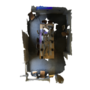}\\
\caption*{scene056900}
\end{subfigure}
\begin{subfigure}{0.19\textwidth}
\includegraphics[width=\textwidth]{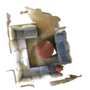}\\
\caption*{scene070101}
\end{subfigure}
\begin{subfigure}{0.19\textwidth}
\includegraphics[width=\textwidth]{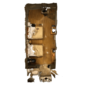}\\
\caption*{scene047701}
\end{subfigure}
\begin{subfigure}{0.19\textwidth}
\includegraphics[width=\textwidth]{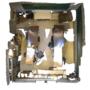}\\
\caption*{scene003002}
\end{subfigure}

\begin{subfigure}{0.19\textwidth}
\includegraphics[width=\textwidth]{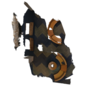}\\
\caption*{scene033402}
\end{subfigure}
\begin{subfigure}{0.19\textwidth}
\includegraphics[width=\textwidth]{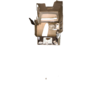}\\
\caption*{scene026502}
\end{subfigure}
\begin{subfigure}{0.19\textwidth}
\includegraphics[width=\textwidth]{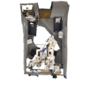}\\
\caption*{scene002501}
\end{subfigure}
\begin{subfigure}{0.19\textwidth}
\includegraphics[width=\textwidth]{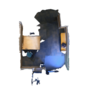}\\
\caption*{scene004300}
\end{subfigure}
\begin{subfigure}{0.19\textwidth}
\includegraphics[width=\textwidth]{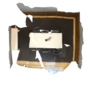}\\
\caption*{scene049301}
\end{subfigure}

\begin{subfigure}{0.19\textwidth}
\includegraphics[width=\textwidth]{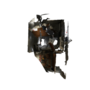}\\
\caption*{scene022400}
\end{subfigure}
\begin{subfigure}{0.19\textwidth}
\includegraphics[width=\textwidth]{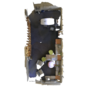}\\
\caption*{scene069400}
\end{subfigure}
\begin{subfigure}{0.19\textwidth}
\includegraphics[width=\textwidth]{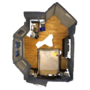}\\
\caption*{scene028602}
\end{subfigure}
\begin{subfigure}{0.19\textwidth}
\includegraphics[width=\textwidth]{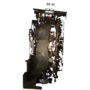}\\
\caption*{scene005701}
\end{subfigure}
\begin{subfigure}{0.19\textwidth}
\includegraphics[width=\textwidth]{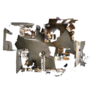}\\
\caption*{scene030900}
\end{subfigure}


\caption{ScanNet Test Dataset (1st to 25th)}
\ContinuedFloat
\label{fig:ScanNet:dataset4}
\end{figure*}

\begin{figure*}[ht!]
\ContinuedFloat

\begin{subfigure}{0.19\textwidth}
\includegraphics[width=\textwidth]{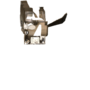}\\
\caption*{scene040602}
\end{subfigure}
\begin{subfigure}{0.19\textwidth}
\includegraphics[width=\textwidth]{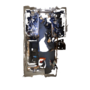}\\
\caption*{scene035300}
\end{subfigure}
\begin{subfigure}{0.19\textwidth}
\includegraphics[width=\textwidth]{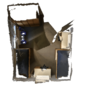}\\
\caption*{scene022300}
\end{subfigure}
\begin{subfigure}{0.19\textwidth}
\includegraphics[width=\textwidth]{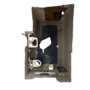}\\
\caption*{scene014602}
\end{subfigure}
\begin{subfigure}{0.19\textwidth}
\includegraphics[width=\textwidth]{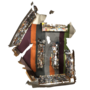}\\
\caption*{scene020800}
\end{subfigure}

\begin{subfigure}{0.19\textwidth}
\includegraphics[width=\textwidth]{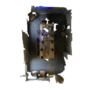}\\
\caption*{scene057502}
\end{subfigure}
\begin{subfigure}{0.19\textwidth}
\includegraphics[width=\textwidth]{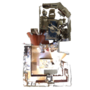}\\
\caption*{scene023101}
\end{subfigure}

\caption{ScanNet Test Dataset (26th to 32nd)}

\label{fig:ScanNet:dataset5}

\end{figure*}

\begin{figure*}[ht!]

\begin{subfigure}{0.19\textwidth}
\includegraphics[width=\textwidth]{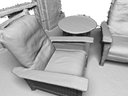}\\
\caption*{05150}
\end{subfigure}
\begin{subfigure}{0.19\textwidth}
\includegraphics[width=\textwidth]{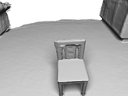}\\
\caption*{01022}
\end{subfigure}
\begin{subfigure}{0.19\textwidth}
\includegraphics[width=\textwidth]{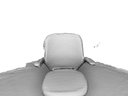}\\
\caption*{01203}
\end{subfigure}
\begin{subfigure}{0.19\textwidth}
\includegraphics[width=\textwidth]{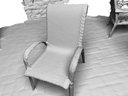}\\
\caption*{05452}
\end{subfigure}
\begin{subfigure}{0.19\textwidth}
\includegraphics[width=\textwidth]{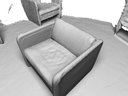}\\
\caption*{05120}
\end{subfigure}
\begin{subfigure}{0.19\textwidth}
\includegraphics[width=\textwidth]{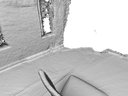}\\
\caption*{05330}
\end{subfigure}
\begin{subfigure}{0.19\textwidth}
\includegraphics[width=\textwidth]{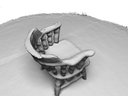}\\
\caption*{05119}
\end{subfigure}
\begin{subfigure}{0.19\textwidth}
\includegraphics[width=\textwidth]{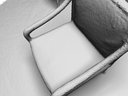}\\
\caption*{05845}
\end{subfigure}
\begin{subfigure}{0.19\textwidth}
\includegraphics[width=\textwidth]{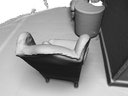}\\
\caption*{05698}
\end{subfigure}
\begin{subfigure}{0.19\textwidth}
\includegraphics[width=\textwidth]{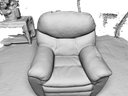}\\
\caption*{06020}
\end{subfigure}
\begin{subfigure}{0.19\textwidth}
\includegraphics[width=\textwidth]{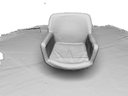}\\
\caption*{01205}
\end{subfigure}
\begin{subfigure}{0.19\textwidth}
\includegraphics[width=\textwidth]{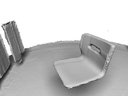}\\
\caption*{01201}
\end{subfigure}
\begin{subfigure}{0.19\textwidth}
\includegraphics[width=\textwidth]{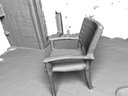}\\
\caption*{06042}
\end{subfigure}
\begin{subfigure}{0.19\textwidth}
\includegraphics[width=\textwidth]{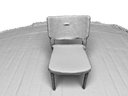}\\
\caption*{05970}
\end{subfigure}
\begin{subfigure}{0.19\textwidth}
\includegraphics[width=\textwidth]{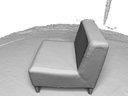}\\
\caption*{01193}
\end{subfigure}
\begin{subfigure}{0.19\textwidth}
\includegraphics[width=\textwidth]{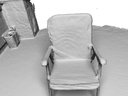}\\
\caption*{01030}
\end{subfigure}
\begin{subfigure}{0.19\textwidth}
\includegraphics[width=\textwidth]{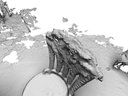}\\
\caption*{06207}
\end{subfigure}
\begin{subfigure}{0.19\textwidth}
\includegraphics[width=\textwidth]{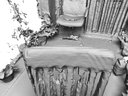}\\
\caption*{00288}
\end{subfigure}
\begin{subfigure}{0.19\textwidth}
\includegraphics[width=\textwidth]{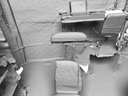}\\
\caption*{00578}
\end{subfigure}
\begin{subfigure}{0.19\textwidth}
\includegraphics[width=\textwidth]{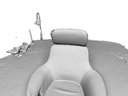}\\
\caption*{01191}
\end{subfigure}
\begin{subfigure}{0.19\textwidth}
\includegraphics[width=\textwidth]{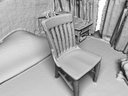}\\
\caption*{00279}
\end{subfigure}
\begin{subfigure}{0.19\textwidth}
\includegraphics[width=\textwidth]{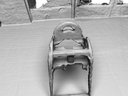}\\
\caption*{05657}
\end{subfigure}
\begin{subfigure}{0.19\textwidth}
\includegraphics[width=\textwidth]{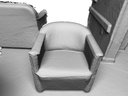}\\
\caption*{00033}
\end{subfigure}
\begin{subfigure}{0.19\textwidth}
\includegraphics[width=\textwidth]{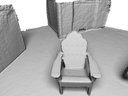}\\
\caption*{01038}
\end{subfigure}
\begin{subfigure}{0.19\textwidth}
\includegraphics[width=\textwidth]{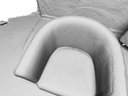}\\
\caption*{06130}
\end{subfigure}

\ContinuedFloat
\caption{Redwood Chair Train Dataset (1st to 25th)}

\label{fig:Redwood:Chair:dataset}
\end{figure*}

\begin{figure*}
\ContinuedFloat

\begin{subfigure}{0.19\textwidth}
\includegraphics[width=\textwidth]{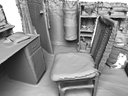}\\
\caption*{06160}
\end{subfigure}
\begin{subfigure}{0.19\textwidth}
\includegraphics[width=\textwidth]{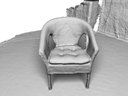}\\
\caption*{01053}
\end{subfigure}
\begin{subfigure}{0.19\textwidth}
\includegraphics[width=\textwidth]{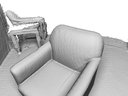}\\
\caption*{05703}
\end{subfigure}
\begin{subfigure}{0.19\textwidth}
\includegraphics[width=\textwidth]{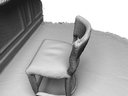}\\
\caption*{05702}
\end{subfigure}
\begin{subfigure}{0.19\textwidth}
\includegraphics[width=\textwidth]{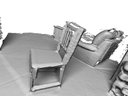}\\
\caption*{01668}
\end{subfigure}
\begin{subfigure}{0.19\textwidth}
\includegraphics[width=\textwidth]{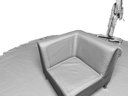}\\
\caption*{01194}
\end{subfigure}
\begin{subfigure}{0.19\textwidth}
\includegraphics[width=\textwidth]{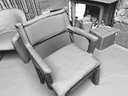}\\
\caption*{00037}
\end{subfigure}
\begin{subfigure}{0.19\textwidth}
\includegraphics[width=\textwidth]{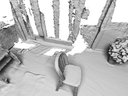}\\
\caption*{05324}
\end{subfigure}

\ContinuedFloat
\caption{Redwood Chair Train Dataset (26th to 33th)}
\label{fig:Redwood:Chair:dataset2}

\end{figure*}

\begin{figure*}

\ContinuedFloat
\begin{subfigure}{0.19\textwidth}
\includegraphics[width=\textwidth]{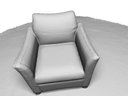}\\
\caption*{01317}
\end{subfigure}
\begin{subfigure}{0.19\textwidth}
\includegraphics[width=\textwidth]{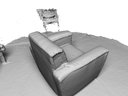}\\
\caption*{01606}
\end{subfigure}
\begin{subfigure}{0.19\textwidth}
\includegraphics[width=\textwidth]{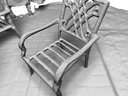}\\
\caption*{05454}
\end{subfigure}
\begin{subfigure}{0.19\textwidth}
\includegraphics[width=\textwidth]{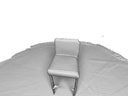}\\
\caption*{01185}
\end{subfigure}
\begin{subfigure}{0.19\textwidth}
\includegraphics[width=\textwidth]{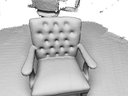}\\
\caption*{05989}
\end{subfigure}
\begin{subfigure}{0.19\textwidth}
\includegraphics[width=\textwidth]{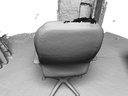}\\
\caption*{06031}
\end{subfigure}
\begin{subfigure}{0.19\textwidth}
\includegraphics[width=\textwidth]{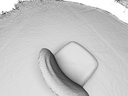}\\
\caption*{00503}
\end{subfigure}
\begin{subfigure}{0.19\textwidth}
\includegraphics[width=\textwidth]{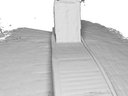}\\
\caption*{05987}
\end{subfigure}
\begin{subfigure}{0.19\textwidth}
\includegraphics[width=\textwidth]{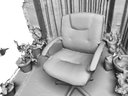}\\
\caption*{00286}
\end{subfigure}
\begin{subfigure}{0.19\textwidth}
\includegraphics[width=\textwidth]{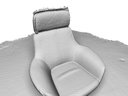}\\
\caption*{06266}
\end{subfigure}
\begin{subfigure}{0.19\textwidth}
\includegraphics[width=\textwidth]{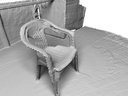}\\
\caption*{01672}
\end{subfigure}
\begin{subfigure}{0.19\textwidth}
\includegraphics[width=\textwidth]{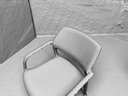}\\
\caption*{05472}
\end{subfigure}
\begin{subfigure}{0.19\textwidth}
\includegraphics[width=\textwidth]{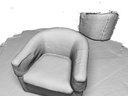}\\
\caption*{05988}
\end{subfigure}
\begin{subfigure}{0.19\textwidth}
\includegraphics[width=\textwidth]{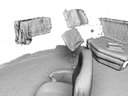}\\
\caption*{05333}
\end{subfigure}
\begin{subfigure}{0.19\textwidth}
\includegraphics[width=\textwidth]{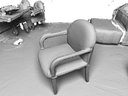}\\
\caption*{01214}
\end{subfigure}
\begin{subfigure}{0.19\textwidth}
\includegraphics[width=\textwidth]{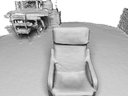}\\
\caption*{01034}
\end{subfigure}
\begin{subfigure}{0.19\textwidth}
\includegraphics[width=\textwidth]{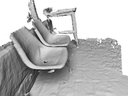}\\
\caption*{01682}
\end{subfigure}
\begin{subfigure}{0.19\textwidth}
\includegraphics[width=\textwidth]{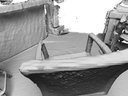}\\
\caption*{05624}
\end{subfigure}
\begin{subfigure}{0.19\textwidth}
\includegraphics[width=\textwidth]{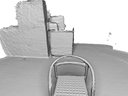}\\
\caption*{01207}
\end{subfigure}
\begin{subfigure}{0.19\textwidth}
\includegraphics[width=\textwidth]{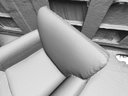}\\
\caption*{06286}
\end{subfigure}
\begin{subfigure}{0.19\textwidth}
\includegraphics[width=\textwidth]{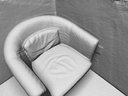}\\
\caption*{05195}
\end{subfigure}
\begin{subfigure}{0.19\textwidth}
\includegraphics[width=\textwidth]{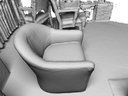}\\
\caption*{00036}
\end{subfigure}
\begin{subfigure}{0.19\textwidth}
\includegraphics[width=\textwidth]{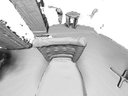}\\
\caption*{06283}
\end{subfigure}
\begin{subfigure}{0.19\textwidth}
\includegraphics[width=\textwidth]{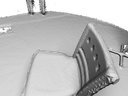}\\
\caption*{06282}
\end{subfigure}
\begin{subfigure}{0.19\textwidth}
\includegraphics[width=\textwidth]{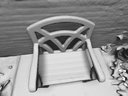}\\
\caption*{06198}
\end{subfigure}

\caption{Redwood Chair Test Dataset}
\label{fig:Redwood:Chair:dataset3}
\end{figure*}

\end{document}


\title{Supplemental Material of Learning Transformation Synchronization}

\author{
Xiangru Huang\\
The University of Texas at Austin\\
Austin, Texas, 78712\\
{\tt\small xrhuang@cs.utexas.edu}
}

\maketitle
\appendix
{\small
\bibliographystyle{ieee}
\bibliography{sync}
}


\title{Supplemental Material of Extreme Relative Pose Estimation for RGB-D Scans via Scene Completion}

\author{Zhenpei Yang\\
Department of Computer Science\\
The University of Texas at Austin\\
Austin, TX 78712, USA \\
\texttt{zhenpei12@cs.utexas.edu} \\
}

\maketitle

{\small
\bibliographystyle{ieee}
\bibliography{sync}
}
\newpage
\appendix